\documentclass[letterpaper,journal]{IEEEtran}
\usepackage{amsmath,amsfonts}
\usepackage{algorithm}
\usepackage{algpseudocode}
\usepackage{array}
\usepackage{textcomp}
\usepackage{stfloats}
\usepackage{url}
\usepackage{verbatim}
\usepackage{graphicx}
\usepackage{cite}
\usepackage{mathtools}
\usepackage[colorlinks,linkcolor=blue,anchorcolor=blue,citecolor=blue,urlcolor=blue,hyperfootnotes=true]{hyperref}
\usepackage[all]{hypcap} 
\usepackage{booktabs,multirow,xcolor}
\usepackage[caption=false,font=footnotesize]{subfig}
\usepackage{siunitx}
\DeclareSIUnit{\deg}{deg}
\newenvironment{ntp}{%
  \renewcommand{\abstractname}{Note to Practitioners}%
  \begin{abstract}%
}{%
  \end{abstract}%
}

\begin{document}

\title{Hierarchical Trajectory Planning of Floating-Base Multi-Link Robot for Maneuvering in Confined Environments}

\author{Yicheng Chen, Jinjie Li, Haokun Liu, Zicheng Luo, Kotaro Kaneko, and Moju Zhao$^{*}$
\thanks{All authors are with the Department of Mechanical Engineering, The University of Tokyo, Tokyo 113-8656, Japan. \textit{(Corresponding author: Moju Zhao)}}
\thanks{Email: \texttt{\{yicheng-chen, jinjie-li, haokun-liu, zicheng, kaneko, chou\}@dragon.t.u-tokyo.ac.jp}}
\thanks{A video demonstration is available at: \url{https://youtu.be/GOcorcrLGb4}}
\thanks{© 2026 IEEE. Personal use of this material is permitted.  Permission from IEEE must be obtained for all other uses, in any current or future media, including reprinting/republishing this material for advertising or promotional purposes, creating new collective works, for resale or redistribution to servers or lists, or reuse of any copyrighted component of this work in other works.}
}



\maketitle

\begin{abstract}
Floating-base multi-link robots can change their shape during flight, making them well-suited for applications in confined environments such as autonomous inspection and search and rescue. However, trajectory planning for such systems remains an open challenge because the problem lies in a high-dimensional, constraint-rich space where collision avoidance must be addressed together with kinematic limits and dynamic feasibility. This work introduces a hierarchical trajectory planning framework that integrates global guidance with configuration-aware local optimization. First, we exploit the dual nature of these robots---the root link as a rigid body for guidance and the articulated joints for flexibility---to generate global anchor states that decompose the planning problem into tractable segments. Second, we design a local trajectory planner that optimizes each segment in parallel with differentiable objectives and constraints, systematically enforcing kinematic feasibility and maintaining dynamic feasibility by avoiding control singularities. Third, we implement a complete system that directly processes point-cloud data, eliminating the need for handcrafted obstacle models. Extensive simulations and real-world experiments confirm that this framework enables an articulated aerial robot to exploit its morphology for maneuvering that rigid robots cannot achieve. To the best of our knowledge, this is the first planning framework for floating-base multi-link robots that has been demonstrated on a real robot to generate continuous, collision-free, and dynamically feasible trajectories directly from raw point-cloud inputs, without relying on handcrafted obstacle models.
\end{abstract}

\vspace{-0pt}

\begin{ntp}
Robots that can change their shape while moving have strong potential for tasks such as inspection, search and rescue, and exploration in spaces that are too tight for conventional drones. The key challenge is coordinating shape changes with safe and efficient navigation in cluttered environments. Our work addresses this by introducing a planning method that generates feasible trajectories considering both flight and configuration changes. For practitioners, the main benefit is improved reliability when operating in confined or unpredictable spaces: the robot can adapt its body to clear obstacles and complete maneuvering that would be impossible for rigid robots. The framework also reduces the need for manual modeling of environments by working directly with standard 3D sensor data. While these capabilities represent a meaningful advance, the current system still relies on accurate sensing and controlled test conditions. Adapting it to real-world uncertainty and longer, more complex missions remains an open direction. Future advances may enable wider deployment in industrial inspection, infrastructure maintenance, or hazardous-area exploration, as well as inspire similar solutions for other transformable robotic platforms.
\end{ntp}

\vspace{2pt}



\begin{figure}[t]
    \setlength{\abovecaptionskip}{-2pt}
    \centerline{\includegraphics[width=1\linewidth]{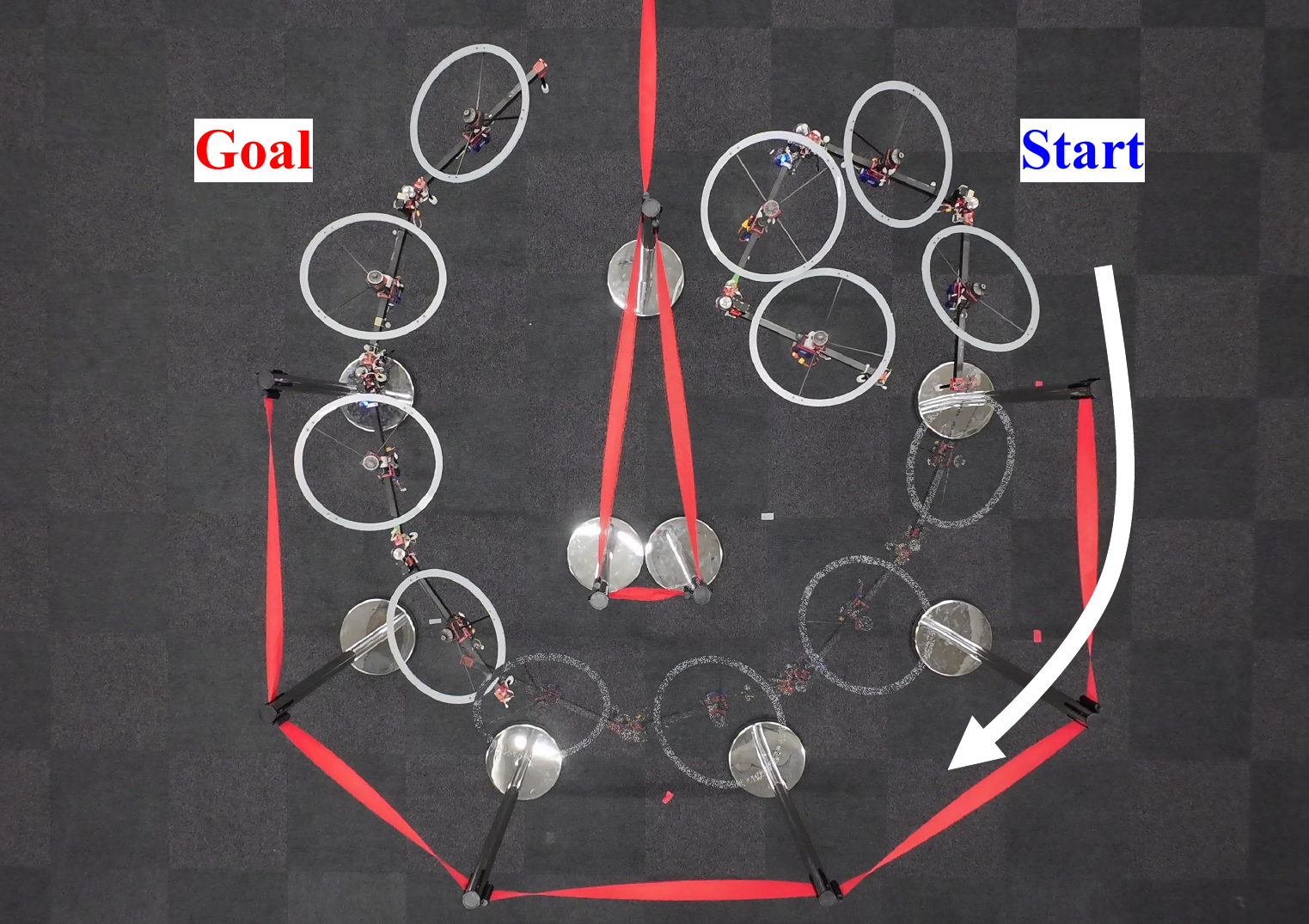}}
    \caption{Top-down view of a floating-base multi-link robot maneuvering through a U-shaped passage by deforming its structure using the hierarchical trajectory planning framework.}
    \label{fig: header_img}
    \vspace{-16pt}
\end{figure}

\section{Introduction}
\IEEEPARstart{T}{ransformable} floating-base multi-link robots can actively reshape their articulated structures during flight \cite{zhao_versatile_2023, zhao_transformable_2018, kulkarni_reconfigurable_2020, hameed_dragonfly_2025}, creating opportunities for applications such as inspection, search-and-rescue, and exploration in confined environments \cite{ollero_past_2022}. Nevertheless, coordinating navigation with configuration changes in such spaces remains an open problem. In this work, we aim to fill this gap by planning trajectories that ensure collision avoidance, respect kinematic limits, and maintain controllability of floating-base multi-link robots in constrained environments, as illustrated in Fig.~\ref{fig: header_img}.

Planning for floating-base multi-link robots poses unique difficulties. Unlike rigid aerial robots, the footprint of floating-base multi-link robots varies with joint motion \cite{lee_introspective_2024}, requiring collision avoidance to be coupled with configuration management. Classical methods such as swept-volume Signed Distance Fields (SDFs) \cite{wang_implicit_2024} can provide gradients for collision checking, but assume predefined shape evolution and are computationally expensive. Borrowing from surface locomotion planners, such as those developed for snakes \cite{thakker_eels_2023, richter_arcsnake_2022, liu_review_2021, vaquero_eels_2024}, introduces unnecessary nonholonomic assumptions, while head-body leader-follower strategies \cite{marshall_survey_2021, zhao_versatile_2021, zhao_flight_2018} tend to straighten the robot, pushing it toward control singularities where available torque is insufficient in certain directions \cite{zhao_versatile_2021, zhao_flight_2018}, thereby degrading controllability. Sampling-based planners in the full configuration space \cite{salzman_motion_2016, gammell_asymptotically_2021, zhao_transformable_2016, yang_rampage_2024, thakar_accelerating_2020} are more general, but suffer from narrow-passage difficulties and poor scalability.

These limitations highlight two open challenges. First, confined-space maneuvering requires large configuration changes, which expands the planning problem into a high-dimensional, constraint-rich space that is intractable without decomposition and guidance. Second, feasible motion demands trajectory-level optimization that can smoothly incorporate kinematic, dynamic, and collision constraints. To date, no framework addresses both challenges in a unified manner for floating-base multi-link aerial robots.

Our key idea is to exploit the dual nature of floating-base multi-link robots: the root link guides overall motion as a rigid body, while the joints provide articulated flexibility for confined-space maneuvering. This insight leads to the concept of \textit{global anchor states}. These anchor states decompose the planning problem into tractable segments and form the basis for our hierarchical framework.

The main contributions are as follows:

\begin{itemize}
\item \textbf{A hierarchical trajectory planning framework that coordinates global guidance with local optimization.} The novelty of the framework is that it introduces \textit{global anchor states} by exploiting the dual nature of floating-base multi-link robots, which combine traits of both rigid and articulated systems. These anchor states divide the global trajectory into independently solvable segments, making the global planning problem more tractable and enabling parallel computation.

\item \textbf{A local trajectory planner specifically designed for floating-base multi-link robots.} We guarantee continuity across global segments without explicit constraints and handle kinematic limits and dynamic feasibility. All objectives and constraints are formulated in a differentiable manner, and a unified constraint formulation keeps the motion free of collision and uncontrollable singularities.

\item \textbf{A complete system implementation that works directly with point-cloud data.} The framework requires no handcrafted semantic obstacle information and is validated through extensive simulations and real-world experiments.
\end{itemize}

The remainder of the paper is organized as follows. Section \ref{related_work} discusses the related work in the literature. Section \ref{hierarchical_trajectory_planning_framework} introduces the robot model and the hierarchical trajectory planning framework. Section \ref{Generation of Global Anchor States} describes the generation of global anchor states, and Section \ref{Local Trajectory Planning} presents the local trajectory planner, including trajectory parameterization and optimization. Section \ref{Simulations} reports implementation details and simulation results, followed by real-world experiments in Section \ref{Experiments}. Finally, Section \ref{Conclusions} concludes the paper and outlines directions for future research.

\section{Related Work}
\label{related_work}
Hierarchical trajectory planning for floating-base multi-link robots spans three threads: specialized methods for such robots, hierarchical planning strategies, and high-dimensional motion planning methods. Despite strong foundations, prior work does not provide a systemic solution for such robots to maneuver in confined spaces. We briefly review the most relevant efforts across these fronts.

\subsection{Motion Planning for Floating-Base Multi-Link Robots}
Early planning pipelines for articulated aerial systems typically combine sampling with post hoc smoothing \cite{zhao_transformable_2016, zhao_flight_2018}, which is computationally expensive and often requires simplified models \cite{salzman_motion_2016}. Differential-kinematics planning \cite{zhao_online_2020} improves responsiveness by optimizing small instantaneous motions, but its greedy horizon and local objectives make it sensitive to local optima. Decoupled methods further separate translation from shape selection. ARCplanner \cite{kulkarni_reconfigurable_2020} first finds a base path and then selects collision-free shapes from libraries or offline samples, with admissibility checked along edges in an occupancy map, but it lacks continuous trajectory optimization, dynamics awareness, and guarantees of cross-edge continuity. Learning-based approaches \cite{al_ali_path_2024} plan in joint space using reward shaping to manage coupling and noise, but they forego model-based guarantees. kinematic model predictive control \cite{liu_coordinated_2024} coordinates base-arm motion under constraints for compliant interaction, yet omits collision handling. Neural whole-body optimizers \cite{jin_whole-body_2024} encode inverse kinematics and trajectory generation with point-cloud losses and heuristic masks, but remain single-horizon and kinematic without segment continuity handling. Related control frameworks for morphing aerial vehicles focus on allocation and geometric control rather than trajectory planning, while surveys document broader morphing trends \cite{hameed_dragonfly_2025, xing_morphing_2024}.

Overall, planners tailored to specific platforms still rely on heuristics or restrictive assumptions, which limits their ability to produce smooth, dynamically feasible trajectories in diverse, confined environments. Our framework addresses these limitations with a more general trajectory planner that directly incorporates differentiable collision costs from raw environment representations, enforces kinematic limits, and maintains controllability.

\subsection{Hierarchical Planning Strategies}
Hierarchical strategies decompose planning into global guidance with local refinement to avoid directly solving a complex long-horizon problem \cite{ding_efficient_2019}. Early waypoint-polynomial pipelines for quadrotors \cite{mellinger_minimum_2011, richter_polynomial_2016} set sparse waypoints and solve a quadratic programming problem on spline coefficients, yielding smooth, dynamically consistent segments. However, they require careful waypoint placement and provide only limited support for collision avoidance. Corridor methods \cite{liu_planning_2017, liu_integrated_2023, wu_deep_2024} construct overlapped convex sets from a global path to convexify collision constraints and enable efficient refinement, yet often become conservative in narrow passages. Optimization-based hierarchical schemes \cite{elango_continuous-time_2025, kuindersma_optimization-based_2016} coordinate subproblems via structured surrogates or sequential convexification, but the induced coupling and continuity constraints can hinder scalability and parallel execution. Similar decompositions appear in manipulators \cite{liu_planning_2025} and legged systems \cite{meng_online_2024}, where task phases or gaits define subproblems. In autonomous navigation and racing \cite{wang_unlocking_2025, ren_online_2023, zhou_raptor_2021}, global references are likewise used to split horizons, but the underlying models typically assume near-rigid bodies. Collectively, these works motivate hierarchical decomposition, yet they offer limited mechanisms to exploit the configuration-dependent flexibility required by floating-base multi-link robots. We address this gap with an anchor-state formulation.

Specifically, we introduce global anchor states that exploit the dual nature of floating-base multi-link robots: the root link provides guidance while the joints enable flexible reconfiguration. Each anchor state defines a critical configuration the robot must traverse, and a sequence of such states divides the overall planning problem into tractable segments that can be solved in parallel.

\subsection{General High-Dimensional Motion Planning Methods}
High-dimensional motion planning faces combinatorial exploration and complex constraints that challenge scalability and reliability \cite{wensing_optimization-based_2023}. Sampling-based methods such as Probabilistic Roadmap (PRM) \cite{kavraki_probabilistic_1996} and Rapidly-exploring Random Tree (RRT) \cite{karaman_sampling-based_2011} provide probabilistic coverage and can be extended toward asymptotic optimality, but they tend to struggle in narrow passages and often require substantial post-processing for smoothness. Optimization-based planners—CHOMP \cite{zucker_chomp_2013}, TrajOpt \cite{schulman_motion_2014}, and continuous-time sequential convexification \cite{elango_continuous-time_2025}—use gradients or convexified surrogates to refine trajectories under collision and dynamics constraints. These ideas are similar to our local optimization component, but they typically do not explicitly address configuration-dependent loss of controllability near singularities. Recent efforts for mobile manipulators improve specific aspects via evolutionary search \cite{pi_omepp_2024}, repetitive structured motions \cite{sun_orthogonal_2024}, real-time safety under dynamics \cite{covic_real-time_2025}, and sampling on screw-constraint manifolds \cite{pettinger_efficient_2024}. While effective within their scopes, these methods largely target ground-actuated platforms. Overall, general-purpose planners offer valuable primitives yet lack mechanisms to co-design base and joint motion in confined environments.

Our approach complements these methods with a unified formulation that preserves both collision avoidance and controllability by enforcing clearance from generalized inadmissible regions, including obstacles and singular configurations. This differentiable construction enables principled coordination of base and joint motion within nonconvex optimization.

\section{Hierarchical Trajectory Planning Framework}
\label{hierarchical_trajectory_planning_framework}

\subsection{Notation}

We adopt the following notation conventions throughout this paper. Scalars are denoted in standard font, either lowercase or uppercase, e.g., $x \in \mathbb{R}$ or $X\in \mathbb{R}$. Vectors are represented by bold lowercase letters, e.g., $\boldsymbol{x} \in \mathbb{R}^n$, and are assumed to be column vectors by default. Matrices are denoted by bold uppercase letters, e.g., $\boldsymbol{X} \in \mathbb{R}^{n \times m}$. We use the operator $\operatorname{vec}(\cdot)$ to represent the 
column-wise vectorization of a matrix. We summarize the main symbols and their meanings in Table \ref{tab: notation}.

\begin{table}[!t]
\caption{Notation used in this paper}
\label{tab: notation}
\centering
\setlength{\tabcolsep}{3pt}
\renewcommand{\arraystretch}{1.00}
\begin{tabular}{l l p{5cm}}
\toprule
\textbf{Symbol} & \textbf{Domain} & \textbf{Description} \\
\midrule

\multicolumn{3}{c}{\textit{Robot Configuration}} \\
\cmidrule(lr){1-3}
$D$ & $\mathbb{Z}^+$ & Dimension of configuration space \\
$m$ & $\{2,3\}$ & Dimension of root link operational space \\
$\boldsymbol{q}$ & $\mathbb{R}^D$ & Full robot configuration vector \\
$\boldsymbol{p}_r$ & $\mathbb{R}^m$ & Root link position vector\\
$\boldsymbol{R}_r$ & $\mathrm{SO}(m)$ & Root link rotation matrix\\
$\theta_i$ & $\mathbb{R}$ & $i$-th joint angle \\
$\theta_{\min}, \theta_{\max}$ & $\mathbb{R}$ & Min. and max. allowable joint angle \\
$n_j$ & $\mathbb{Z}^+$ & Number of joints \\
$n_r$ & $\mathbb{Z}^+$ & Number of rotors \\
$L$ & $\mathbb{R}^+$ & Single link length \\
$R_p$ & $\mathbb{R}^+$ & Propeller radius \\
\midrule

\multicolumn{3}{c}{\textit{Robot Kinematics and Dynamics}} \\
\cmidrule(lr){1-3}
$\boldsymbol{p}^{\text{rot}}$ & $\mathbb{R}^3$ & Rotor postion \\
$\prescript{C}{}{\boldsymbol{e}}_i$ & $\mathbb{R}^3$ & Unit vector along the upforward direction of the $i$-th rotor in CoG frame \\
$\kappa$ & $\mathbb{R}$ & Rotor drag torque coefficient\\
$\boldsymbol{\tau}_i$ &  $\mathbb{R}^3$ & Torque contribution on the center of gravity from the $i$-th rotor \\
$d_{ij}^{\tau}(\boldsymbol{q})$ & $\mathbb{R}$ & Distance metric for controllability \\
$\delta_{\tau}$ & $\mathbb{R}^+$ &  Minimum admissible control torque \\
$\boldsymbol{J^{\boldsymbol{p}}}$ & $\mathbb{R}^{3 \times D}$ & Translational Jacobian matrix of the vector $\boldsymbol{p}$ w.r.t. robot's configuration $\boldsymbol{q}$ \\

\midrule

\multicolumn{3}{c}{\textit{Generation of global anchor states}} \\
\cmidrule(lr){1-3}

$d^{\text{ESDF}}(\boldsymbol{p})$ & $\mathbb{R}$ & Distance from $\boldsymbol{p}$ to nearest obstacle \\
$\delta_{\text{collision}}$ & $\mathbb{R}^+$ & Clearance margin to obstacles \\

$\boldsymbol{P}$ & $\mathbb{R}^{n_p \times m}$ & Global reference path for root with $n_p$ waypoints\\
$\mathcal{Q}$ & -- & Candidate set of local target states \\
$n_\theta$ & $\mathbb{Z}^+$ & Number of candidate local target states \\
$\mathcal{Q}_{\text{feas}}$ & -- & Feasible subset of $\mathcal{Q}$ \\
$\mathcal{J}(\boldsymbol{q})$ & $\mathbb{R}$ & Cost for candidate target selection \\
\midrule

\multicolumn{3}{c}{\textit{Trajectory Parameterizationn}} \\
\cmidrule(lr){1-3}
$T$ & $\mathbb{R}^+$ & Trajectory duration \\
$p$ & $\mathbb{R}^+$ & Degree of B-spline \\
$B_{i,p}(t)$ & $\mathbb{R}$ & The $i$-th B-spline basis function of degree $p$ \\
$\boldsymbol{c}_i$ & $\mathbb{R}^D$ & $i$-th control point of B-spline \\
$N$ & $\mathbb{Z}^+$ & Number of free control points \\
$\boldsymbol{u}$ & $\mathbb{R}^{N+p+5}$ & Knot vector of B-spline \\
$h$ & $\mathbb{R}^+$ & Knot interval of B-spline \\
$\boldsymbol{C}$ & $\mathbb{R}^{N \times D}$ & Matrix of free control points (opt. variables) \\
$\boldsymbol{C}_{\text{full}}$ & $\mathbb{R}^{(N+4)\times D}$ & Matrix of full control points\\
$\alpha_v$ & $\mathbb{R^+}$ & Generalized transition velocity \\
\midrule

\multicolumn{3}{c}{\textit{Trajectory Optimization}} \\
\cmidrule(lr){1-3}
$f(\boldsymbol{C})$ & -- & Objective/constraint function \\
$\boldsymbol{M}$ & $\mathbb{R}^{(N+4)\times(N+4)}$ & Precomputed energy matrix \\
$K$ & $\mathbb{Z}^+$ & Number of samples for approximating continuous-time constraints\\
$\alpha_K$ & $\mathbb{R^+}$ & Sample density parameter of continuous-time constraints\\
$\phi(t)$ & $\mathbb{R}$ & penalty function at time $t$ \\
$\nabla_{\boldsymbol{x}} y$ & -- & gradient of $y$ w.r.t. vector $\boldsymbol{x}$ \\
$v_{\text{max}}$ & $\mathbb{R}$ & Maximum translational velocity magnitude \\
$\omega_{\text{max}}$ & $\mathbb{R}$ & Maximum rotational velocity magnitude \\
\bottomrule
\end{tabular}
\vspace{-10pt}
\end{table}

\subsection{Robot Model and Problem Statement}

We describe the configuration of a floating-base multi-link robot as
\begin{equation}
    \boldsymbol{q} = 
    \begin{bmatrix} 
        \operatorname{vec}(\boldsymbol{q}_r)^\top, \; \boldsymbol{q}_j^\top 
    \end{bmatrix}^\top \in \mathbb{R}^D,
\end{equation}
where the root link pose is
\begin{equation}
    \boldsymbol{q}_r = \left( \boldsymbol{p}_r, \boldsymbol{R}_r \right) \in \mathrm{SE}(m),
\end{equation}
with $\boldsymbol{p}_r \in \mathbb{R}^m$ denoting the translational component and $\boldsymbol{R}_r \in \mathrm{SO}(m)$ the rotational component.  

The vectorization of the root pose is defined as
\begin{equation}
    \operatorname{vec}(\boldsymbol{q}_r) 
    = \begin{bmatrix} \boldsymbol{p}_r^\top, \; \rho(\boldsymbol{R}_r)^\top \end{bmatrix}^\top \in \mathbb{R}^{m+d_r},
\end{equation}
where $\rho: \mathrm{SO}(m) \to \mathbb{R}^{d_r}$ denotes a minimal parameterization of the rotation 
(e.g., $d_r=1$ with a yaw angle $\psi$ when $m=2$, or $d_r=3,4$ with Euler angles or a unit quaternion when $m=3$).  

\begin{figure}[t]
    \setlength{\abovecaptionskip}{-2pt} 
    \centerline{\includegraphics[width=0.8\linewidth]{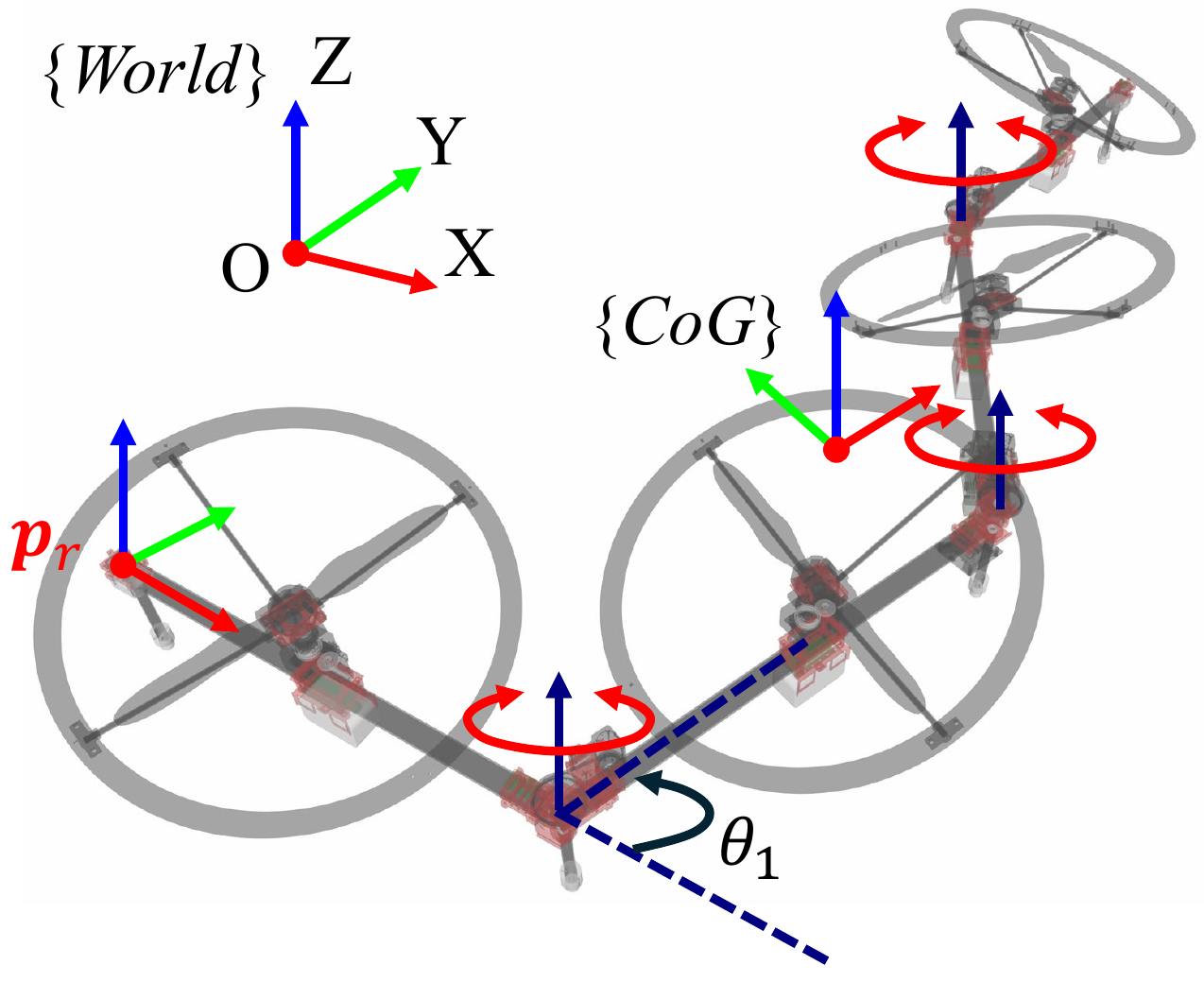}}
    \caption{Kinematic model of the floating-base multi-link robot. The root link position is denoted by $\boldsymbol{p}_r$. Consecutive links are connected by revolute joints, each allowing a single degree of rotational freedom. The joint angle $\theta_1$, for example, is defined as the relative rotation between two adjacent links. The origin of the frame \{CoG\} is the center of gravity of the robot.}
    \label{fig: model}
    \vspace{-10pt}
\end{figure}

The joint configuration is represented by
\begin{equation}
    \boldsymbol{q}_j = \begin{bmatrix}\theta_1, \theta_2, \dots, \theta_{n_j}\end{bmatrix}^\top \in \mathbb{R}^{n_j},
\end{equation}
which collects the $n_j$ joint angles. An illustrative example is provided in Fig.~\ref{fig: model}.

The planning problem can be described as: Finding a trajectory $\boldsymbol{q}^\star(t)$ to connect the given global initial state $\boldsymbol{q}_{\text{global}}^{\text{init}}$ and target state $\boldsymbol{q}_{\text{global}}^{\text{target}}$ with duration $T$, while remaining collision-free, energy-efficient, and satisfying user-defined constraints.

\begin{figure*}[b]
    \setlength{\abovecaptionskip}{-2pt} 
    \centerline{\includegraphics[width=0.95\linewidth]{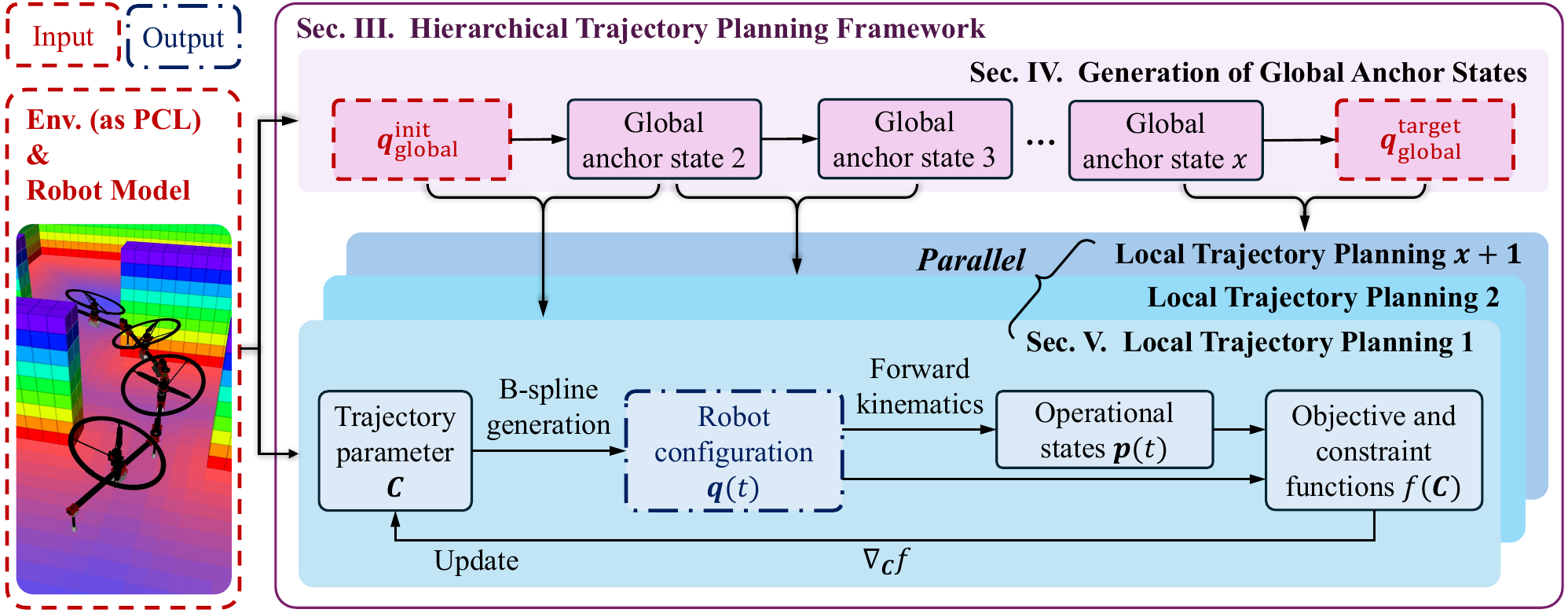}}
    \caption{System overview. The hierarchical trajectory planning framework decomposes the planning problem into independently solvable segments by introducing global anchor states, enabling parallel local planning. With clamped B-spline trajectory parameterization, continuity between adjacent segments is inherently maintained in the overall trajectory. Each segment is optimized in parallel with fully differential objective and constraint functions. The optimized local segments are directly concatenated in sequence to produce the final trajectory.}
    \label{fig: system_overview}
\end{figure*}

\subsection{Framework Overview}
The core motivation for adopting a hierarchical framework is to simplify the high-dimensional and complex trajectory optimization for floating-base multi-link robots by dividing it into smaller, independently solvable segments, thus enabling efficient parallel computation and improved convergence. As illustrated in Fig. \ref{fig: system_overview}, we first generate a series of global anchor states that exploit the robots' dual nature, combining rigid-body guidance with articulated flexibility (Section \ref{Generation of Global Anchor States}). These anchor states divide the global trajectory into independently solvable segments. Then, for each segment bounded by a pair of anchor states, we use clamped B-spline trajectory parameterization to ensure continuity across global segments and perform parallel local optimization that enforces collision avoidance, kinematic limits, and controllability (Section \ref{Local Trajectory Planning}). Finally, the trajectory segments are concatenated to form a whole trajectory. Algorithm~\ref{alg:hierarchical_framework} summarizes this process.

\begin{algorithm}[t]
\caption{Hierarchical Trajectory Planning Framework}
\label{alg:hierarchical_framework}
\begin{algorithmic}[1]
\Require $\boldsymbol{q}_{\text{global}}^{\text{init}}$, $\boldsymbol{q}_{\text{global}}^{\text{target}}$, environment map
\Ensure $\boldsymbol{q}^\star(t)$ (globally feasible trajectory)

\State Generate global anchor states $\mathcal{S}$ using Algorithm~\ref{alg:intermediate_states_list}
\State Initialize empty set $\mathcal{T}$ 
\Statex \Comment{\textit{Holds optimized local trajectory segments}}

\For{$s = 0$ to $|\mathcal{S}| - 2$ \textbf{in parallel}}
    \State $\boldsymbol{q}^{\text{init}}_s \gets \mathcal{S}[s]$ \Comment{\textit{Local initial state}}
    \State $\boldsymbol{q}^{\text{target}}_s \gets \mathcal{S}[s+1]$ \Comment{\textit{Local target state}}
    \State Obtain local trajectory $\boldsymbol{q}^\star_s(t)$ using Algorithm~\ref{alg:local_trajectory_optimization}
    \State Append $\boldsymbol{q}^\star_s(t)$ to $\mathcal{T}$
\EndFor

\State Concatenate all $\boldsymbol{q}^\star_s(t)$ in the order of $\mathcal{S}$ to obtain $\boldsymbol{q}^\star(t)$
\Statex \Comment{\textit{Total duration equals the sum of segment durations}}

\State \Return $\boldsymbol{q}^\star(t)$

\end{algorithmic}

\end{algorithm}

\section{Generation of Global Anchor States}
\label{Generation of Global Anchor States}

Each anchor state is defined as a full robot configuration in $\mathbb{R}^{D}$ and is consistent with the global initial and target configurations. It serves simultaneously as the target state of one local planning segment and the initial state of the next. Although sampling-based planners such as RRT and PRM are widely used in high-dimensional motion planning, directly sampling in the full configuration space is computationally expensive. We therefore decompose the generation of global anchor states into two stages.

First, to provide global guidance for the robot's motion, we compute a collision-free path in $\mathbb{R}^{m}$ from the root's global initial position $\boldsymbol{p}_{\text{global}}^{\text{init}}$ to the global target position $\boldsymbol{p}_{\text{global}}^{\text{target}}$. This path guides the motion of the root link, but the root is not required to follow it exactly. We use A* search \cite{hart_formal_1968} to obtain this guidance path because A* offers deterministic guarantees and runs efficiently in low-dimensional spaces. In our case, the path dimension $m$ is at most 3. The resulting path
\begin{equation}
    \boldsymbol{P} = \left[\boldsymbol{p}_1, \boldsymbol{p}_2, \dots, \boldsymbol{p}_{n_p-1}, \boldsymbol{p}_{n_p} \right]^\top \in \mathbb{R}^{n_p \times m}, \quad \boldsymbol{p}_i \in \mathbb{R}^m,
\end{equation}
serves purely as a geometric reference for the root and does not impose any constraints on the robot's joint configuration.

Second, we iteratively construct each global anchor state. This amounts to incrementally determining a valid local target configuration from the current local initial configuration, while ensuring that the robot progresses toward the global goal. At each step, the procedure consists of three stages:
\begin{enumerate}
    \item \textbf{Construction of the candidate local target set.} From the current local initial configuration, we generate a discrete set of candidate local targets directly in configuration space that respect kinematic structure and advance the robot forward.
    \item \textbf{Feasibility evaluation of independent candidate states.} Each candidate is tested for collision avoidance and controllability, yielding the feasible subset.
    \item \textbf{Selection of the best local target.} Among feasible candidates, the one closest to the reference path and advancing furthest along it is chosen as the best local target, i.e., the next global anchor state.
\end{enumerate}

By iterating this procedure from the global start to the goal, we obtain a sequence of anchor states that divides the planning problem into independently solvable segments. In this section, we denote a candidate local target state by $\boldsymbol{q}^{\text{target}}$ and the selected optimal target by ${\boldsymbol{q}^{\text{target}}}^\star$.

\subsection{Construction of the Candidate Local Target Set}
\label{Construction of the Candidate Local Target Set}
\begin{figure}[t]
    \setlength{\abovecaptionskip}{-2pt} 
    \centerline{\includegraphics[width=1\linewidth]{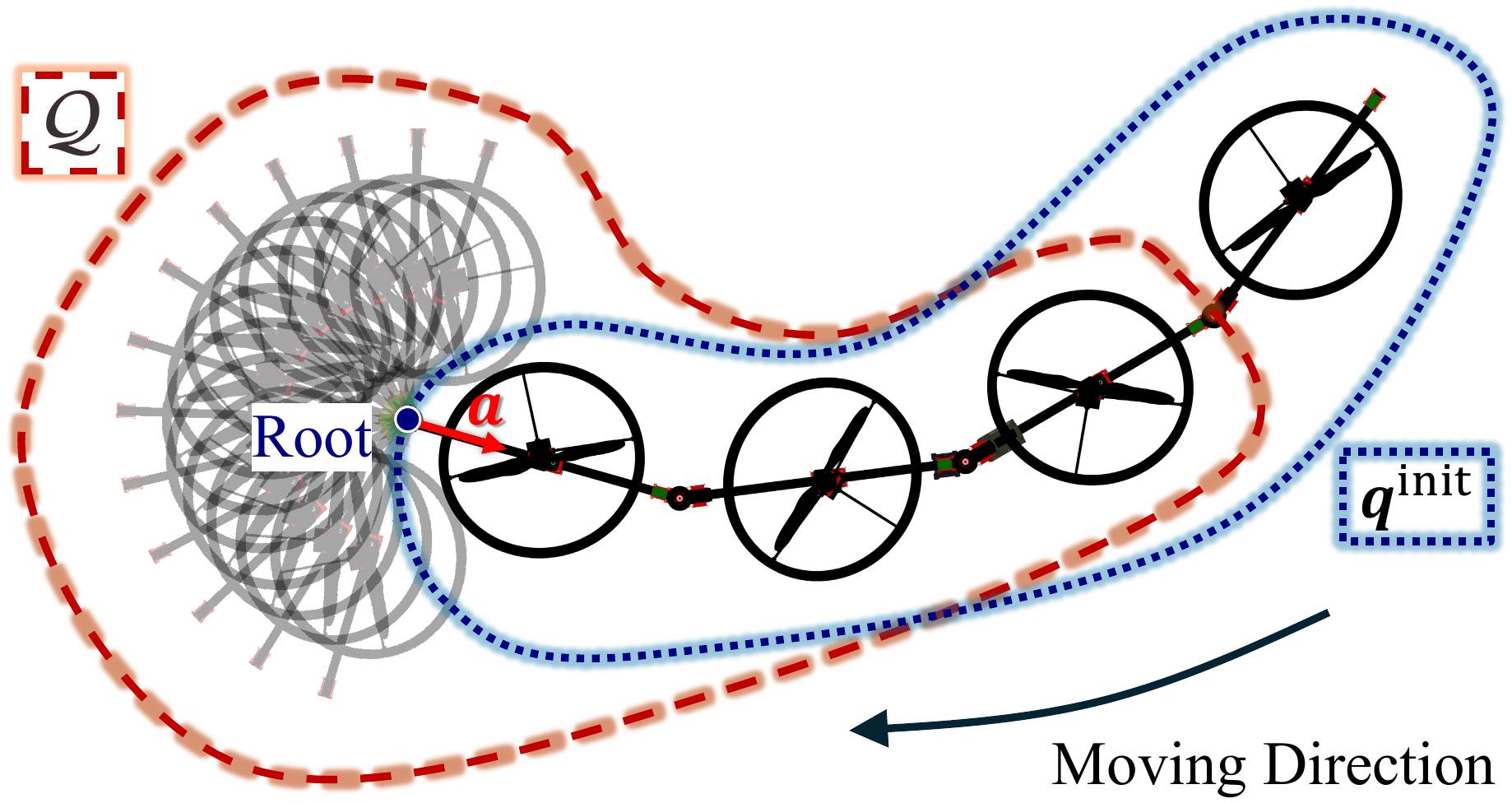}}
    \caption{Illustration of generating the candidate local target set $\mathcal{Q}$ from a given local initial state $\boldsymbol{q}^{\text{init}}$. The set $\mathcal{Q}$ is obtained by sampling the root link's new pose within the allowable joint angle range, exploring configurations located one link length ahead of the initial state.}
    \label{fig: candidate_set}
    \vspace{-10pt}
\end{figure}

Following the principle of state lattice planning, given a local initial state
\begin{equation}
    \boldsymbol{q}^{\text{init}} = \left [ {\boldsymbol{p}_r^{\text{init}}}^\top, \rho(\boldsymbol{R}_r^{\text{init}})^\top, \theta_1^{\text{init}}, \dots, \theta_{n_j}^{\text{init}} \right ] ^\top,
\end{equation}
we construct a set of candidate local targets
\begin{equation}
\label{eq: Q}
    \mathcal{Q} = 
    \left\{\, \boldsymbol{q}^{\text{target}}\left(\boldsymbol{q}^{\text{init}}, \Delta \theta \right) 
    \;\middle|\; \Delta \theta \in \Theta \right\},
\end{equation}
with
\begin{equation}
    \Theta = \left\{\, \theta_{\min} + k\frac{\theta_{\max}-\theta_{\min}}{n_{\theta}-1} 
    \;\middle|\; k=0,\dots,n_{\theta}-1 \right\},
\end{equation}
where $\theta_{\min}$ and $\theta_{\max}$ denote the allowable joint angle limits, and $\Delta \theta$ is chosen from $n_{\theta}$ uniformly spaced samples in the interval $[\theta_{\min}, \theta_{\max}]$, with the physical meaning of the relative rotation from the candidate root link to the initial root link. By construction, $\Delta \theta$ directly corresponds to the first joint angle of the candidate state. As illustrated in Fig.~\ref{fig: candidate_set}, each candidate is explicitly given by
\begin{subequations}
\begin{align}
    \boldsymbol{q}^{\text{target}} &= \left[ {\boldsymbol{p}_r^{\text{target}}}^\top, \rho(\boldsymbol{R}_r^{\text{target}})^\top, \theta_1^{\text{target}}, \dots, \theta_{n_j}^{\text{target}} \right] ^\top, \\
    \boldsymbol{p}_r^{\text{target}} &= \boldsymbol{p}_r^{\text{init}} - L \boldsymbol{R}_r^{\text{target}} \boldsymbol{a}, \\
    \boldsymbol{R}_r^{\text{target}} &= \mathrm{Rot}(-\Delta \theta) \boldsymbol{R}_r^{\text{init}}, \\
    \theta_1^{\text{target}} &= \Delta \theta, \\
    \theta_i^{\text{target}} &= \theta_{i-1}^{\text{init}}, \quad i = 2, 3, \dots, {n_j},
\end{align}
\end{subequations}
where $L$ is the link length, $\boldsymbol{a}$ is the unit vector along the $x$-axis of the root link in its body-fixed frame, and $\mathrm{Rot}(-\Delta \theta)$ denotes a rotation by $-\Delta \theta$ about the joint axis. Geometrically, the root tip advances to a point lying on the surface of a hypersphere of radius $L$ centered at the initial root position, so the candidate set $\mathcal{Q}$ spans the configuration space one link length ahead.

\subsection{Feasibility Evaluation of an Independent State}
A candidate state $\boldsymbol{q}$ (superscript omitted for brevity) is feasible only if it is simultaneously collision-free and controllable.

\subsubsection{\textbf{Collision Check}}
The collision between the robot and the obstacles in the environment can be evaluated with some kinds of map representation, e.g., the Euclidean Signed Distance Fields (ESDF). Querying a point $\boldsymbol{p}$ in ESDF gives the distance $d^{\text{ESDF}}(\boldsymbol{p})$ from $\boldsymbol{p}$ to its nearest obstacle. Therefore, we can perform a collision check by checking whether some critical points on the robot are far enough from the obstacles. 

For example, given a specific robot configuration $\boldsymbol{q}$, we can get the positions of all rotors using forward kinematics $f^{\text{fk}}(\cdot)$, i.e, 
\begin{equation}
\boldsymbol{p}_i^{\text{rot}} = f_{i}^{\text{fk}}(\boldsymbol{q}) \in \mathbb{R}^3, \quad i=1,2,\dots,n_r,
\label{eq: rotor_pos}
\end{equation}
where $n_r$ is the number of the rotors. In practice, we set an approximate condition to seek a balance between computation burden and safety: the robot is deemed collision-free if the distance from each rotor to its nearest obstacle exceeds a certain threshold:
\begin{equation}
\label{eq: collision_check}
    d_{i}^{\text{ESDF}}(\boldsymbol{p}_i^{\text{rot}}) > R_p + \delta_{\text{collision}}, \quad \forall i \in \{1,2,\dots, n_r\},
\end{equation}
where $R_p$ is the radius of the propeller and $\delta_{\text{collision}}$ is the clearance margin to obstacles.

\subsubsection{\textbf{Controllability Check}}
The configuration of a floating-base multi-link robot strongly affects its controllability, particularly in attitude when the system is underactuated. To quantitatively evaluate this, we adopt the metric introduced in \cite{zhao_versatile_2021}. Given a specific robot configuration $\boldsymbol{q}$, all available control torques $\boldsymbol{\tau}(\boldsymbol{q})$ about its center of gravity (CoG) form a convex polyhedron
\begin{equation}
\mathcal{T}(\boldsymbol{q}) = \left\{ \, \boldsymbol{\tau}(\boldsymbol{q}) \in \mathbb{R}^3 \,\middle|\,
\begin{aligned}
\boldsymbol{\tau}(\boldsymbol{q}) 
= \sum_{i=1}^{n_r} \boldsymbol{\tau}_i(\boldsymbol{q})
\end{aligned}
\right\},
\end{equation}
and the contribution from the $i$-th rotor $\boldsymbol{\tau}_i (\boldsymbol{q}) \in \mathbb{R}^3$ is given by
\begin{equation}
\boldsymbol{\tau}_i (\boldsymbol{q})= \lambda_i \left(\prescript{C}{}{\boldsymbol{p}}_i^{\text{rot}} \times \prescript{C}{}{\boldsymbol{e}}_i + \kappa s_i \prescript{C}{}{\boldsymbol{e}}_i \right), \quad 0 \leq \lambda_i \leq \lambda_{\max},
\end{equation}
where $\lambda_i$ is the thrust generated by the $i$-th rotor, $\prescript{C}{}{\boldsymbol{p}}_i^{\text{rot}}$ is its position in the CoG frame, $\prescript{C}{}{\boldsymbol{e}}_i$ is the unit vector along its thrust direction in the CoG frame, $\kappa$ is the rotor drag torque coefficient, and $s_i$ equals $+1$ for counterclockwise rotation and $-1$ for clockwise rotation.

The robot is controllable about all rotational axes if the origin of the torque space lies strictly inside the convex polyhedron $\mathcal{T}(\boldsymbol{q})$. For a convex polyhedron, this condition is equivalent to the signed distance from the origin to each supporting face plane being strictly positive. Since $\mathcal{T}(\boldsymbol{q})$ has $n_r(n_r-1)$ faces, we can evaluate this condition by checking the distances associated with all faces. In particular, the distance $d_{ij}^{\tau}(\boldsymbol{q})$ from the origin to the face plane whose normal is aligned with $\boldsymbol{\tau}_i \times \boldsymbol{\tau}_j$ (i.e., the plane spanned by $\boldsymbol{\tau}_i$ and $\boldsymbol{\tau}_j$) can be obtained by
\begin{subequations}
\label{eq: fc_t_dis}
\begin{align}
    d_{ij}^{\tau}(\boldsymbol{q}) &= 
\sum_{k=1}^{n_r} \max \left( 0, \tau_{ijk} \right), \\ 
\tau_{ijk} &= \frac{ \left( \boldsymbol{\tau}_i \times \boldsymbol{\tau}_j \right)^{\top} }{ \left\| \boldsymbol{\tau}_i \times \boldsymbol{\tau}_j \right\| } \boldsymbol{\tau}_k.
\end{align}
\end{subequations}
Therefore, robot of configuration $\boldsymbol{q}$ is controllable if the following condition holds:
\begin{equation}
\label{eq: control_check}
\tau_{\min}(\boldsymbol{q}) 
= \min_{i,j \in \mathcal{I}, i \ne j} d_{ij}^{\tau}(\boldsymbol{q}) > \delta_{\tau},\quad \mathcal{I}=\{1,2,\dots,n_r\},
\end{equation}
where $\delta_{\tau}$ denotes the platform-specific minimum admissible control torque.

Among the candidates in $\mathcal{Q}$, we define $\mathcal{Q}_{\text{feas}}$ as the subset that satisfies both the collision-free condition \eqref{eq: collision_check} and the controllability condition \eqref{eq: control_check}.

\subsection{Selection of the Best Local Target}
\begin{figure}[t]
    \centerline{\includegraphics[width=1\linewidth]{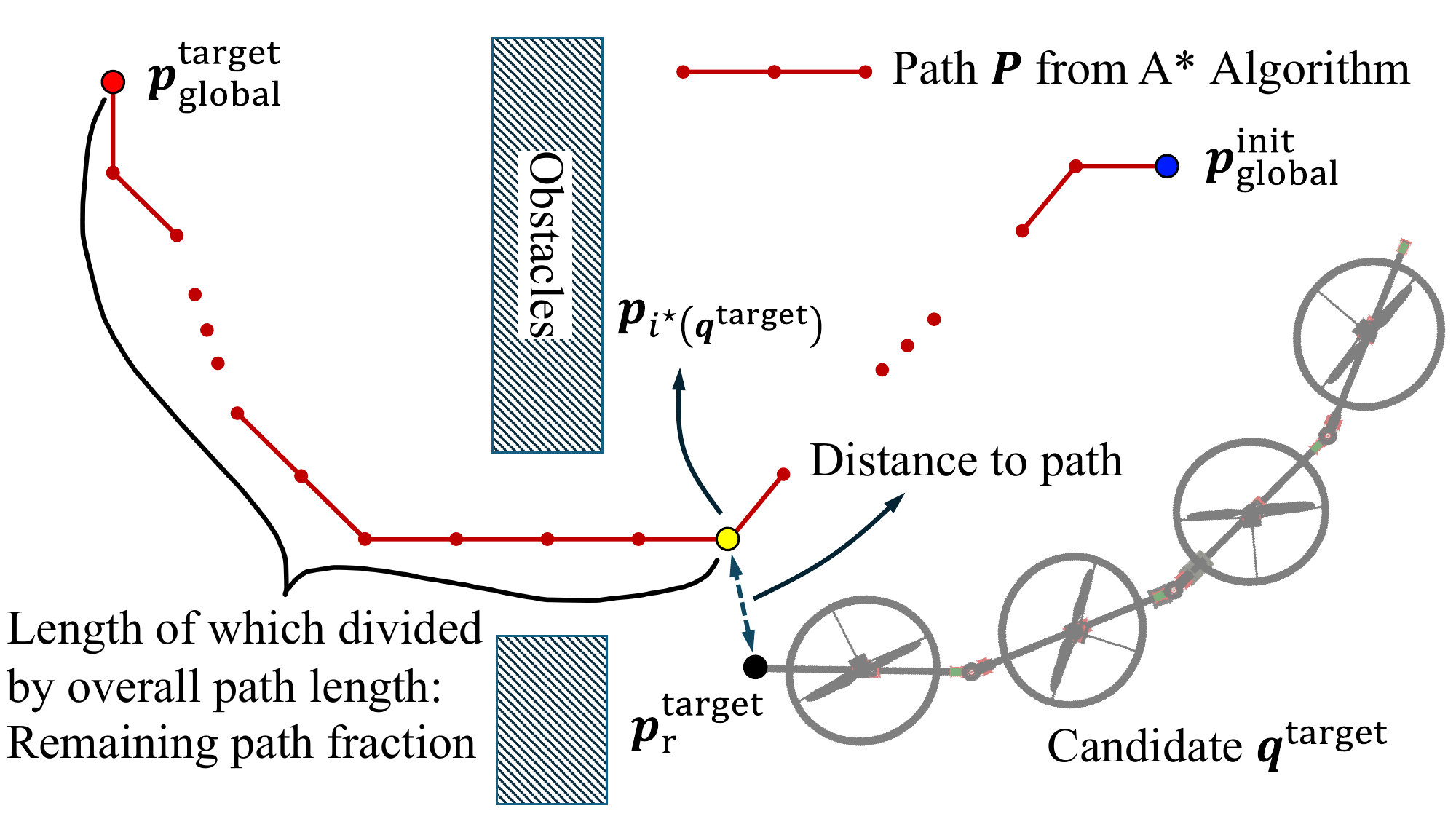}}
    \caption{Illustration of candidate target evaluation. Each candidate configuration is evaluated based on two factors: the distance between its root position and the global reference path, and the fraction of the path that remains ahead of this closest point. An ideal target should lie close to the reference path while advancing along it.}
    \label{fig: candidate_eva}
    \vspace{-10pt}
\end{figure}

An ideal target should lie close to the reference path while advancing along it. As visualized in Fig. \ref{fig: candidate_eva}, to evaluate the quality of each candidate, we assign every feasible $\boldsymbol{q}^{\text{target}} \in \mathcal{Q}_{\text{feas}}$ a cost
\begin{equation}
\mathcal{J}(\boldsymbol{q}^{\text{target}}) =
\underbrace{\left\| \boldsymbol{p}_{i^\star(\boldsymbol{q}^{\text{target}})} 
- \boldsymbol{p}_r^{\text{target}} \right\|_2}_{\mathclap{\text{Distance to path}}}
+
\underbrace{\left(1 - \tfrac{i^\star(\boldsymbol{q}^{\text{target}})}{n_p} \right)}_{\mathclap{\text{Remaining path fraction}}},
\end{equation}
where $i^\star$ is the index of the nearest path waypoint in $\boldsymbol{P}$, defined as
\begin{equation}
i^\star(\boldsymbol{q}^{\text{target}}) = \arg\min_{i \in \{1, \dots, n_p\}} \left\| \boldsymbol{p}_i - \boldsymbol{p}_r^{\text{target}} \right\|_2.
\end{equation}
The optimal local target is then selected by
\begin{equation}
\label{eq: best_local_target}
{\boldsymbol{q}^{\text{target}}}^\star = \arg\min_{\boldsymbol{q}^{\text{target}} \in \mathcal{Q}_{\text{feas}}} \mathcal{J}(\boldsymbol{q}^{\text{target}}).
\end{equation}

\begin{algorithm}[b]
\caption{Generation of Global Anchor States}
\label{alg:intermediate_states_list}
\begin{algorithmic}[1]
\Require $\boldsymbol{q}_{\text{global}}^{\text{init}}$, $\boldsymbol{q}_{\text{global}}^{\text{target}}$, ESDF map
\Ensure $\mathcal{S}$ (ordered list of global anchor states)

\State Compute $\boldsymbol{P}$ by A* search from $\boldsymbol{p}_{\text{global}}^{\text{init}}$ 
      to $\boldsymbol{p}^{\text{target}}_{\text{global}}$
\Statex \Comment{\textit{Global root-link reference path}}

\State $\mathcal{S} \gets [\boldsymbol{q}_{\text{global}}^{\text{init}}]$
\State $\boldsymbol{q}^{\text{init}} \gets \boldsymbol{q}_{\text{global}}^{\text{init}}$

\While{$\left\| \boldsymbol{p}_r^{\text{init}}(\boldsymbol{q}^{\text{init}}) - \boldsymbol{p}^{\text{target}}_{\text{global}} \right\|_2 > \varepsilon$}
    \State Construct candidate set $\mathcal{Q}$ from $\boldsymbol{q}^{\text{init}}$ using \eqref{eq: Q}
    \State Obtain feasible subset $\mathcal{Q}_{\text{feas}} \subseteq \mathcal{Q}$ satisfying \eqref{eq: collision_check} and \eqref{eq: control_check}
    \State Select ${\boldsymbol{q}^{\text{target}}}^\star$ from $\mathcal{Q}_{\text{feas}}$ using \eqref{eq: best_local_target}
    \State Append ${\boldsymbol{q}^{\text{target}}}^\star$ to $\mathcal{S}$
    \State $\boldsymbol{q}^{\text{init}} \gets {\boldsymbol{q}^{\text{target}}}^\star$
\EndWhile

\State Append $\boldsymbol{q}_{\text{global}}^{\text{target}}$ to $\mathcal{S}$
\State \Return $\mathcal{S}$

\end{algorithmic}
\end{algorithm}

Starting from the global initial state, we iteratively select the next global anchor state until reaching the global target, as summarized in Algorithm~\ref{alg:intermediate_states_list}. 

The proposed anchor-state generation procedure is superficially similar to PRM in that both use local feasibility checks to obtain a collision-free route, but the mechanisms differ fundamentally. PRM samples directly in the full configuration space, connects samples with local planners, and then performs graph search on the resulting roadmap. In contrast, our method separates global guidance and configuration generation: the A* search is carried out only in the low-dimensional workspace of the root link, and no roadmap is constructed in the full configuration space. Full robot configurations are generated deterministically and locally from the current anchor state via the lattice-based candidate set $\mathcal{Q}$, producing a single feasible chain of anchor states rather than a reusable graph. This structure-aware design is more efficient and task-focused for global anchor-state generation.

This process decomposes the planning problem into independent segments, each spanning from a local initial state to a local target state and solvable in parallel. Continuity of the trajectory between adjacent segments is required, and in the following section, we introduce a parameterization method that inherently ensures this property.

\section{Local Trajectory Planning}
\label{Local Trajectory Planning}

This section introduces the local trajectory planner for each segment defined by global anchor states. We begin with a clamped B-spline parameterization that preserves continuity across segments. We then formulate the objective and constraints in a fully differentiable manner, with a unified treatment of collision avoidance and controllability. Forward kinematics is invoked whenever evaluating these terms. Fig. \ref{fig: system_overview} illustrates the variable layers and dependencies, and Algorithm \ref{alg:local_trajectory_optimization} summarizes the overall computation procedure.

\subsection{Trajectory Parameterization}

For a trajectory of $\boldsymbol{q}$ that starts from $\boldsymbol{q}^\text{init}$ with velocity $ \boldsymbol{v}^\text{init}$ and ends at $\boldsymbol{q}^\text{target}$ with velocity $ \boldsymbol{v}^\text{target}$, we parameterize it using uniform clamped B-Splines of degree $p$, expressed as
\begin{equation}
\boldsymbol{q}(t) = \sum_{i=0}^{N+3} B_{i,p} \left(t \right) \boldsymbol{c}_i.
\label{eq: B-Splines}
\end{equation}
In this formulation, $\boldsymbol{c}_i$ $(i = 0, 1, \dots, N+3)$ denote the control points, where $N$ is the number of free control points. The basis functions $B_{i,p}(\cdot)$ $(i = 0, 1, \dots, N+3)$ are uniquely defined by a knot vector of length $N+p+5$, which we specify as
\begin{equation}
\label{eq: knot}
    \boldsymbol{u} = [\underbrace{0, \dots, 0}_{p+1}, h, 2h,\dots, (N+3-p)h, \underbrace{T, \dots, T}_{{p+1}}].
\end{equation}
Here $h = T/(N+4-p)$ is the uniform knot interval. The parameter $T$ denotes the trajectory duration and is chosen as
\begin{equation}
    T = \left \| \boldsymbol{q}^{\text{init}} - \boldsymbol{q}^{\text{target}}\right \| _2 / \alpha_v.
\end{equation}
The scalar $\alpha_v$ is a user-defined dimensionless scaling factor with the physical meaning of a generalized transition velocity: it sets the ratio between the geometric distance in configuration space and the trajectory duration.

This formulation clamps the trajectory at both ends: the first and last control points coincide with the start and end configurations, and the repeated knots ensure that boundary velocities also match. As a result, the local trajectory maintains velocity continuity across the global segments defined in Section \ref{Generation of Global Anchor States}. Concretely, the boundary conditions are imposed as
\begin{subequations}
\begin{align}
\boldsymbol{c}_{0} &= \boldsymbol{q}^\text{init}, \\
\boldsymbol{c}_{N+3} &= \boldsymbol{q}^\text{target}, \\
\boldsymbol{c}_{1} &= \boldsymbol{q}^\text{init} + \boldsymbol{v}^\text{init} h, \\
\boldsymbol{c}_{N+2} &= \boldsymbol{q}^\text{target} - \boldsymbol{v}^\text{target} h.
\end{align}
\label{eq: bound_conditions}
\end{subequations}

Given these boundary conditions and factor $\alpha_v$, the trajectory is entirely determined by the control points $\boldsymbol{c}_2$ to $\boldsymbol{c}_{N+1}$. We select these attributes as the optimization variables in trajectory optimization and denote
\begin{equation}
    \boldsymbol{C} = \left [ \boldsymbol{c}_2, \boldsymbol{c}_3, \dots, \boldsymbol{c}_{N+1} \right ] ^\top \in \mathbb{R}^{N\times D},
\end{equation}
\begin{equation}
    \boldsymbol{C}_\text{full} = \left[\boldsymbol{c}_0, \boldsymbol{c}_1, \boldsymbol{C}^\top, \boldsymbol{c}_{N+2}, \boldsymbol{c}_{N+3} \right] ^\top \in \mathbb{R}^{(N+4)\times D}.
\end{equation}

\subsection{Trajectory Optimization}

We formulate the trajectory planning problem as an optimization problem
\begin{equation}
\begin{aligned}
\min_{\operatorname{vec}(\boldsymbol{C})} \quad & f_{\text{energy}}(\boldsymbol{C}) \\
\text{s.t.} \quad & \boldsymbol{C}_{\text{min}} < \boldsymbol{C} < \boldsymbol{C}_{\text{max}}, \\
                 & \boldsymbol{f}_\text{vel}(\boldsymbol{C}) \le  \textbf{0}, \\
                 & f_\text{collision}(\boldsymbol{C}) \le 0, \\
                 & f_\text{controllability}(\boldsymbol{C}) \le 0,
\end{aligned}
\label{eq: opt_problem}
\end{equation}
where the operator $\operatorname{vec}(\cdot)$ represents the 
column-wise vectorization of a matrix. Then, we present the specific objective and constraint functions in detail.

\subsubsection{\textbf{Energy Cost}}
\label{sec: Energy Cost}
The energy cost is evaluated by integrating the velocity along the trajectory:
\begin{subequations}
\label{eq: energy cost}
\begin{align}
    f_{\text{energy}}(\boldsymbol{C}) &= \int_{0}^{T} \left \| \dot{\boldsymbol{q}}(t) \right \|^2 \text{dt} \\
    &= \int_{0}^{T}\sum_{i=0}^{N+3} \sum_{j=0}^{N+3}\boldsymbol{c}_i^\top \dot{B}_{i,p}(t) \dot{B}_{j,p}(t) \boldsymbol{c}_j \text{dt} \\
    &= \sum_{i=0}^{N+3} \sum_{j=0}^{N+3} \boldsymbol{c}_{i}^\top M_{ij} \boldsymbol{c}_j \\
    &= \text{Tr}\left(\boldsymbol{C}_{\text{full}}^\top \boldsymbol{M} \boldsymbol{C}_{\text{full}}\right), \\
M_{ij} &= \int_{0}^{T}\dot{B}_{i,p}(t) \dot{B}_{j,p}(t) \text{dt}.
\end{align}
\end{subequations}
In our framework, once $T$ is specified, $\boldsymbol{M} \in \mathbb{R}^{(N+4) \times (N+4)}$ is determined and can be precomputed in an analytical way. An example is provided in Appendix \ref{appendix_a}.

The gradient of $f_{\text{energy}}(\boldsymbol{C})$ with respect to (w.r.t.) $\operatorname{vec}(\boldsymbol{C})$ is
\begin{equation}
    \nabla_{\operatorname{vec}(\boldsymbol{C})} f_{\text{energy}} = \operatorname{vec}\left( \left(2 \boldsymbol{M} \boldsymbol{C}_{\text{full}} \right)_{2:N+1, :}\right) \in \mathbb{R}^{ND}.
\end{equation}

\subsubsection{\textbf{Joint Angle Constraints}} 
\label{sec: Joint Angle Constraints}
Since joint angles are part of the configuration space, their bounds can be imposed directly on the corresponding components of the control points by exploiting the convex hull property of B-splines. Specifically,\begin{subequations}
\begin{align}
\left(\boldsymbol{C}_{\min}\right)_{i,j} &= \theta_{\min}, \quad i = 1,\dots,N,\ j = m,\dots,D,\\
\left(\boldsymbol{C}_{\max}\right)_{i,j} &= \theta_{\max}, \quad i = 1,\dots,N,\ j = m,\dots,D.
\end{align}
\end{subequations}

\subsubsection{\textbf{Velocity Constraint}}
\label{sec: Velocity Constraint}
To prevent the robot from performing radical motion, we constrain the maximum velocity along the trajectory. The derivative of $\boldsymbol{q}(t)$ is a B-spline of degree $p-1$, controlled by $N+3$ control points
\begin{equation}
    \boldsymbol{\beta}_i = \frac{1}{h}\left(\boldsymbol{c}_{i+1} - \boldsymbol{c}_i \right) \in \mathbb{R}^{D},\quad i=0,1,\dots, N+2.
\end{equation}

We denote the collection of these control points as
\begin{subequations}
\begin{align}
\boldsymbol{G}(\boldsymbol{C}) 
&= \left [ \boldsymbol{\beta}_0, \boldsymbol{\beta}_1, \dots, \boldsymbol{\beta}_{N+2} \right ] ^\top \in \mathbb{R}^{(N+3)\times D} \\
&= \boldsymbol{A}\boldsymbol{C}_{\text{full}},
\end{align}
\end{subequations}
where 
\begin{equation}
\boldsymbol{A} = \frac{1}{h} \begin{bmatrix}-1  &1  &0  &\dots &0 \\0  &-1  &1  &\dots  &0 \\\vdots  &   &\ddots  &  &\vdots \\0  &\dots  &0  &-1  &1\end{bmatrix}_{(N+3) \times (N+4)}.
\end{equation}

We vectorize $\boldsymbol{G}(\boldsymbol{C})$ by column, thus getting
\begin{equation}
\operatorname{vec}(\boldsymbol{G}(\boldsymbol{C})) = 
    \left( \boldsymbol{I}_D \otimes \boldsymbol{A} \right) \operatorname{vec}(\boldsymbol{C}_{\text{full}}) \in \mathbb{R}^{(N+3)D},
\end{equation}
where $\boldsymbol{I}_D$ denotes the identity matrix of dimension $D$, and $\otimes$ denotes the operation of calculating Kronecker product. 

Using the convex hull property of B-Splines, to set constraints on the velocity of the trajectory, we constrain all control points of $\dot{\boldsymbol{q}}(t)$ to be within the velocity threshold. Therefore, we can formulate the velocity constraint $\boldsymbol{f}_\text{vel}(\boldsymbol{C}) \in \mathbb{R}^{2(N+3)D}$ as
\begin{equation}
    \boldsymbol{f}_\text{vel}(\boldsymbol{C}) =  \begin{bmatrix}
\operatorname{vec}(\boldsymbol{G}(\boldsymbol{C})) - \left[{\boldsymbol{\nu}_{1}^{\text{max}}}^\top, \dots, {\boldsymbol{\nu}_{D}^{\text{max}}}^\top \right]^\top \\
-\operatorname{vec}(\boldsymbol{G}(\boldsymbol{C})) - \left[{\boldsymbol{\nu}_{1}^{\text{max}}}^\top, \dots, {\boldsymbol{\nu}_{D}^{\text{max}}}^\top \right]^\top
\end{bmatrix},
\end{equation}
where $\boldsymbol{\nu}_{i}^{\text{max}} \in \mathbb{R}^{N+3}\ (i=1,\dots,D)$ specifies the maximum allowable velocity magnitude along the $i$-th dimension of the trajectory $\dot{\boldsymbol{q}}(t)$. For instance, let $v_{\max}$ denote the maximum translational velocity of the root link, and $\omega_{\max}$ the maximum rotational velocity associated with both the root link orientation and the joint angles, then
\begin{equation}
\boldsymbol{\nu}_{i}^{\text{max}} =
\begin{cases}
\displaystyle
v_{\text{max}} \cdot \mathbf{1}_{N+3},
& \text{if } i \in \left\{1,2,\dots,m  \right\}, \\
\omega_{\text{max}} \cdot \mathbf{1}_{N+3},
& \text{if } i \in \left\{m+1,m+2,\dots,D  \right\},
\end{cases}
\end{equation}
with $\mathbf{1}_{N+3} \in \mathbb{R}^{N+3}$ denoting the all-ones vector.

The gradient of $\boldsymbol{f}_\text{vel}(\boldsymbol{C})$ w.r.t. $\operatorname{vec}(\boldsymbol{C})$ can be derived as
\begin{equation}
\nabla_{\operatorname{vec}(\boldsymbol{C})} \boldsymbol{f}_{\text{vel}} = \begin{bmatrix}
\boldsymbol{I}_D \otimes \boldsymbol{A} \\
-\boldsymbol{I}_D \otimes \boldsymbol{A}
\end{bmatrix}_{:, 2D:(N+1)D} \in \mathbb{R}^{2(N+3)D \times ND}.
\end{equation}

\subsubsection{\textbf{Sampling-Based Approximation of Continuous-Time Constraints}}
\label{sec: Sampling-Based Constraints}
Collision avoidance and controllability are inherently continuous-time constraints, as they must hold throughout the entire trajectory rather than at isolated instants. To enforce them in optimization, we adopt an integral penalty formulation that accumulates violations over time, following the general principle of constraint transcription \cite{jennings_computational_1990}. 

Specifically, for a class of constraints (e.g., collision avoidance or controllability) at time $t$, let $d_i(\boldsymbol{q}(t))\ (i=1,\dots, \mu)$ denote the non-negative constraint function, where $\mu$ denotes the number of constraint functions within this class. Ideally, we require
\begin{equation}
    d_i(\boldsymbol{q}(t)) > \delta, \quad \forall t \in [0, T], \quad i=1, \dots, \mu.
\label{eq: g_func}
\end{equation}
where $\delta$ is the minimum allowed value of all $d_i(\boldsymbol{q}(t))$ of the class along the trajectory.
We define the violation of \eqref{eq: g_func} at time $t$ as
\begin{subequations}
\begin{align}
\Phi(t) &= \sum_{i=1}^{\mu} \phi \left(d_i(t) \right), \\
\phi(d_i(t)) &= 
\left\{
\begin{aligned}
  &\frac{1}{2\delta} \left(d_i(\boldsymbol{q}(t)) - \delta\right)^2, &&\text{if } d_i(\boldsymbol{q}(t)) < \delta, \\
  &0, \quad &&\text{else}.
\end{aligned}
\right.
\end{align}
\end{subequations}

Constraint \eqref{eq: g_func} is satisfied if the time integral of $\Phi(t)$ vanishes. Since evaluating this integral exactly is intractable, we approximate it using a uniform sampling scheme. Specifically, $K$ states are sampled along the trajectory, and the integral is replaced with a discrete sum:
\begin{subequations}
\label{eq: f_sample}
\begin{align}
f_{\text{sampled}}(\boldsymbol{C}) &= \sum_{n=1}^{K} \sum_{i=1}^{\mu} \phi(d_i(t_n)),\\
t_n &= \frac{n}{K}T,\\
K &= \left \lceil \alpha_K \left \| \boldsymbol{q}^{\text{init}} - \boldsymbol{q}^{\text{target}}\right \| _2 \right \rceil,
\end{align}
\end{subequations}
where $\alpha_K$ is a sampling density parameter that specifies how many samples are taken per unit distance in configuration space. This sampling-based integral approximation provides a practical and unified way to handle both collision avoidance and controllability constraints within the trajectory optimization framework. Its gradient w.r.t. $\operatorname{vec}(\boldsymbol{C})$ can be obtained using the chain rule
\begin{equation}
\begin{split}
\label{eq: grad_sample2Q}
\nabla_{\operatorname{vec}(\boldsymbol{C})} f_{\text{sampled}} 
&= \sum_{n=1}^{K} \sum_{i=1}^{\mu}
\frac{\partial \phi(d_i(t_n))}{\partial d_i(t_n)} \\
&\left(
    \nabla_{\operatorname{vec}(\boldsymbol{C})} \boldsymbol{q}(t_n)
\right)^\top 
\nabla_{\boldsymbol{q}(t_n)} d_i(t_n).
\end{split}
\end{equation}

For a given sampled time $t_n$, we can easily get
\begin{equation}
\frac{\partial \phi(d_i(t_n))}{\partial d_i(t_n)} = 
\left\{
\begin{aligned}
  &\frac{1}{\delta} \left(d_i(\boldsymbol{q}(t_n)) - \delta\right), &&\text{if } d_i(\boldsymbol{q}(t_n)) < \delta, \\
  &0, \quad &&\text{else}.
\end{aligned}
\right.
\end{equation}

To obtain $\nabla_{\operatorname{vec}(\boldsymbol{C})} \boldsymbol{q}(t_n)$, we rewrite \eqref{eq: B-Splines} as
\begin{subequations}
\begin{align}
\boldsymbol{q}^\top(t) &= \boldsymbol{b}(t) \boldsymbol{C}_{\text{full}} \\
& = \left(\boldsymbol{I}_D \otimes \boldsymbol{b}(t) \right) \operatorname{vec}(\boldsymbol{C}_{\text{full}}), \\
\boldsymbol{b}(t) &= \left [ B_{0,p}(t),B_{1,p}(t),\dots,B_{N+3,p}(t) \right ].
\end{align}
\end{subequations}
Therefore, this gradient can be derived as
\begin{equation}
\nabla_{\operatorname{vec}(\boldsymbol{C})} \boldsymbol{q}(t_n) = \left(\boldsymbol{I}_D \otimes \boldsymbol{b}(t_n) \right)_{:,2D:(N+1)D} \in \mathbb{R}^{D \times ND}.
\end{equation}

The last term $\nabla_{\boldsymbol{q}(t_n)}d_i(t_n) \in \mathbb{R}^D$ in \eqref{eq: grad_sample2Q} is determined by the specific implementation of the constraint function $d(\cdot)$, and will be presented below. 

\textit{4-1) Collision Avoidance Constraint}: For a state $\boldsymbol{q}(t)$ from the trajectory (time $t$ omitted below for brevity), we can calculate each rotor's position $\boldsymbol{p}_i^{\text{rot}}(\boldsymbol{q})$ using forward kinematics $f^{\text{fk}}(\cdot)$ as in \eqref{eq: rotor_pos}. The ESDF is then queried to obtain the distance $d_{i}^{\text{ESDF}}(\boldsymbol{p}_i^{\text{rot}})$ from $\boldsymbol{p}_i^{\text{rot}}$ to its nearest obstacle, along with its gradient $\nabla_{\boldsymbol{p}_i^{\text{rot}}}d_{i}^{\text{ESDF}} \in \mathbb{R}^3$.

Following the general definition in \eqref{eq: g_func}, the collision-avoidance constraint for rotor $i$ at configuration $\boldsymbol{q}$ is given by its distance to the nearest obstacle, $d_{i}^{\text{ESDF}}(\boldsymbol{q})$. In this case, the clearance parameter $\delta$ in \eqref{eq: g_func} is set to $R_p + \delta_{\text{collision}}$. 

Substituting $d_{i}^{\text{ESDF}}(\boldsymbol{q})$ into \eqref{eq: f_sample}, with the inner summation running over all rotors ($\mu=n_r$), yields the sampled penalty for collision avoidance. The resulting function $f_{\text{collision}}(\cdot)$ is therefore a direct instance of the generic sampled constraint $f_{\text{sampled}}(\cdot)$. 

By the chain rule, the gradient of $d_{i}^{\text{ESDF}}(\boldsymbol{q})$ w.r.t. $\boldsymbol{q}$ is
\begin{equation}
\label{eq: grad_esdf2q}
\nabla_{\boldsymbol{q}} d_i^{\text{ESDF}} = \boldsymbol{J}_{i}^{\text{rot}}(\boldsymbol{q})^\top 
\nabla_{\boldsymbol{p}_i^{\text{rot}}} d_{i}^{\text{ESDF}},
\end{equation}
where $\boldsymbol{J}_{i}^{\text{rot}}(\boldsymbol{q}) \in \mathbb{R}^{3\times D}$ is the Jacobian matrix of $\boldsymbol{p}_i^{\text{rot}}$ w.r.t. the robot's configuration $\boldsymbol{q}$. Substituting \eqref{eq: grad_esdf2q} into the last term in \eqref{eq: grad_sample2Q}, with the inner summation extended over all rotors ($\mu=n_r$), gives the gradient of the collision avoidance constraint $\nabla_{\operatorname{vec}(\boldsymbol{C})} f_{\text{collision}} \in \mathbb{R}^{ND}$.

\textit{4-2) Controllability Constraint}:
According to the general definition in \eqref{eq: g_func}, the constraint function for controllability maintenance at configuration $\boldsymbol{q}$ is given by the distance measure $d_{ij}^{\tau}(\boldsymbol{q})$ defined in \eqref{eq: fc_t_dis}. Here the indexing adapts to the pairwise structure of the measure, with $i,j \in \mathcal{I},\ i \neq j$, where $\mathcal{I}=\{1,2,\dots,n_r\}$. In this case, the clearance parameter $\delta$ in \eqref{eq: g_func} is set to $\delta_{\tau}$. 

Substituting \eqref{eq: fc_t_dis} into \eqref{eq: f_sample}, with the inner summation running over all rotor pairs $i,j \in \mathcal{I},\ i \neq j$, yields the sampled penalty for controllability maintenance. The resulting function $f_{\text{controllability}}(\cdot)$ is therefore another direct instance of the generic sampled constraint $f_{\text{sampled}}(\cdot)$.

The gradient of $d_{ij}^{\tau}(\boldsymbol{q})$ w.r.t. $\boldsymbol{q}$ is
\begin{equation}
\label{eq: grad_controllability2q}
\nabla_{\boldsymbol{q}} d_{ij}^{\tau} = \sum_{k=1}^{n_r} 1_{\{\tau_{ijk} > 0\}} \nabla_{\boldsymbol{q}}\tau_{ijk},
\end{equation}
where the term $1_{\{\tau_{ijk} > 0\}} \in \mathbb{R}$ is an indicator function that returns the scalar value $1$ if $\tau_{ijk} > 0$, and $0$ otherwise. The expression of $\nabla_{\boldsymbol{q}}\tau_{ijk}$ is given in Appendix \ref{appendix_b}. It is determined by $\left(\boldsymbol{\tau}_i, \boldsymbol{\tau}_j, \boldsymbol{\tau}_k \right)$ and their gradients w.r.t. $\boldsymbol{q}$, each of which can be calculated as
\begin{equation}
\label{eq: grad_tau2q}
\nabla_{\boldsymbol{q}}\boldsymbol{\tau}_{i} = \lambda_i \left(-\left[\prescript{C}{}{\boldsymbol{e}}_i \right]_{\times} \prescript{C}{}{\boldsymbol{J}}_{i}^{\text{rot}} + \left[\prescript{C}{}{\boldsymbol{p}}_i^{\text{rot}} \right]_{\times} \prescript{C}{}{\boldsymbol{J}}_{i}^{\boldsymbol{e}} + \kappa s_i \prescript{C}{}{\boldsymbol{J}}_{i}^{\boldsymbol{e}} \right),
\end{equation}
where the operation $[\cdot]_\times$ denotes calculating the skew-symmetric matrix. $\prescript{C}{}{\boldsymbol{J}}_{i}^{\text{rot}}, \prescript{C}{}{\boldsymbol{J}}_{i}^{\boldsymbol{e}} \in \mathbb{R}^{3 \times D}$ denote the Jacobian matrix of the $i$-th rotor's position and the vector $\boldsymbol{e}$ w.r.t. $\boldsymbol{q}$ in the CoG frame, respectively. Their expressions are given in Appendix \ref{appendix_c}.

Finally, substituting \eqref{eq: grad_controllability2q} into the last term of \eqref{eq: grad_sample2Q}, with the inner summation spanning all pairs $i,j \in \mathcal{I},\ i \neq j$, yields the gradient of the controllability constraint $\nabla_{\operatorname{vec}(\boldsymbol{C})} f_{\text{controllability}} \in \mathbb{R}^{ND}$.

In summary, collision avoidance and controllability are both cast into the same sampling-based penalty framework. As a feature of our formulation, this unified and differentiable treatment provides a consistent way to approximate continuous-time safety and feasibility conditions.

Algorithm \ref{alg:local_trajectory_optimization} summarizes the overall computation procedure of this section.

\begin{algorithm}[t]
\caption{Local Trajectory Optimization}
\label{alg:local_trajectory_optimization}
\begin{algorithmic}[1]
\Require $\boldsymbol{q}^{\text{init}}, \boldsymbol{v}^{\text{init}}, \boldsymbol{q}^{\text{target}}, \boldsymbol{v}^{\text{target}}$, ESDF map
\Ensure Local optimized trajectory $\boldsymbol{q}^\star(t)$

\State Compute the clamped knot vector $\boldsymbol{u}$ from \eqref{eq: knot}
\State Set boundary control points $\boldsymbol{c}_{0}, \boldsymbol{c}_{1}, \boldsymbol{c}_{N+2}, \boldsymbol{c}_{N+3}$ using \eqref{eq: bound_conditions}
\State Initialize free control points $\boldsymbol{C} = [\boldsymbol{c}_2,\dots,\boldsymbol{c}_{N+1}]^\top$
\Statex \Comment{\textit{Free control points $\boldsymbol{C}$ are the optimization variables}}

\State Establish the mapping $\boldsymbol{q}(t) \gets \texttt{B\text{-}spline}(\boldsymbol{C})$ using \eqref{eq: B-Splines}
\Statex \Comment{\textit{Evaluated whenever an objective or constraint function is computed}}

\State \textbf{Build the optimization problem \eqref{eq: opt_problem} over $\boldsymbol{C}$}, with components obtained as follows:
\Statex \hspace{\algorithmicindent}1) Compute the energy objective $f_{\text{energy}}(\boldsymbol{C})$ and its gradient (Sec.~\ref{sec: Energy Cost})
\Statex \hspace{\algorithmicindent}2) Apply joint angle limits (Sec.~\ref{sec: Joint Angle Constraints})
\Statex \hspace{\algorithmicindent}3) Construct the velocity constraint $\boldsymbol{f}_{\text{vel}}(\boldsymbol{C})$ and its gradient (Sec.~\ref{sec: Velocity Constraint})
\Statex \hspace{\algorithmicindent}4) Define sampled penalty functions $f_{\text{sampled}}(\boldsymbol{C})$ and their gradients (Sec.~\ref{sec: Sampling-Based Constraints})
\Statex \hspace{0.8cm}4-1) Obtain the collision-avoidance penalty $f_{\text{collision}}(\boldsymbol{C})$ and its gradient using $d_i^{\text{ESDF}}(\boldsymbol{q})$ and \eqref{eq: grad_esdf2q}
\Statex \hspace{0.8cm}4-2) Obtain the controllability penalty $f_{\text{controllability}}(\boldsymbol{C})$ and its gradient using $d_{ij}^{\tau}(\boldsymbol{q})$ and \eqref{eq: grad_controllability2q}

\State Solve the optimization problem \eqref{eq: opt_problem} to obtain $\boldsymbol{C}^\star$
\State Obtain the final trajectory $\boldsymbol{q}^\star(t) \gets \texttt{B\text{-}spline}(\boldsymbol{C}^\star)$
\State \Return $\boldsymbol{q}^\star(t)$

\end{algorithmic}
\end{algorithm}

\section{Implementation and Simulation}
\label{Simulations}

\subsection{Implementation and Reproducibility}
\label{sec: Implementation and Reproducibility}

\begin{figure}[t]
    \setlength{\abovecaptionskip}{-2pt} 
    \centerline{\includegraphics[width=1\linewidth]{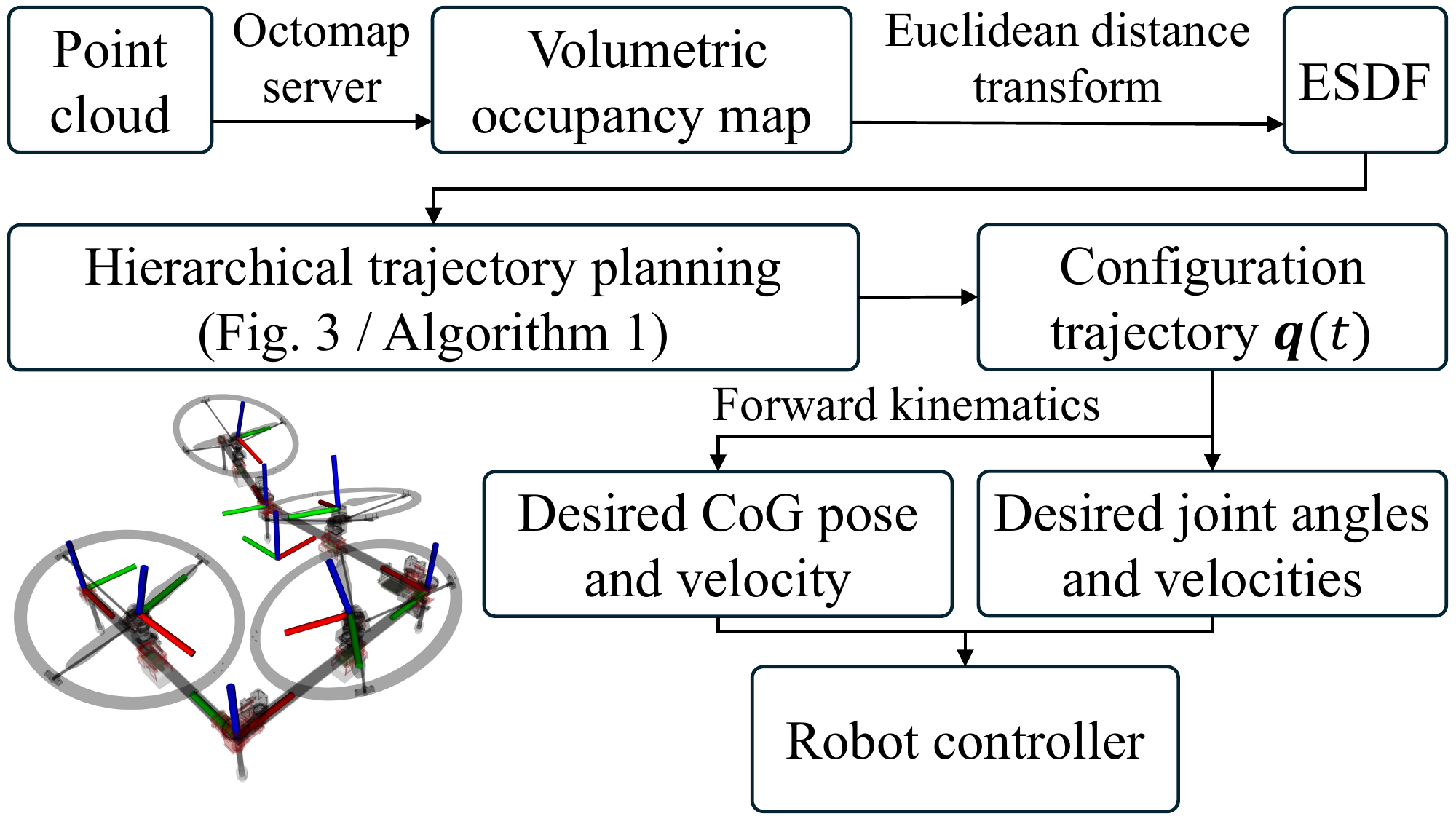}}
    \caption{Overview of the perception, planning, and control pipeline. Coordinate transformation is managed by the TF tools in ROS.}
    \label{fig: implementation_pipeline}
    \vspace{-10pt}
\end{figure}

We implemented the system in the \textit{Robot Operating System (ROS)} and used \textit{Pinocchio}\footnote{\url{https://stack-of-tasks.github.io/pinocchio}} for forward-kinematics evaluation. Simulations were conducted in Gazebo with the floating-base multi-link robot shown in Fig.~\ref{fig: model} on a personal computer equipped with an AMD Ryzen 9 9950X CPU (16 cores, 32 threads). In the square configuration (all three joints at \SI{90}{\degree}), the robot diameter was \SI{0.89}{\meter}. The environment was represented by offline-generated point clouds with \SI{0.1}{\meter} resolution. The whole perception, planning, and control pipeline is illustrated in Fig. \ref{fig: implementation_pipeline}.

When processing point-cloud data, we built and maintained a volumetric occupancy map online using the \textit{Octomap Server}\footnote{\url{https://wiki.ros.org/octomap_server}} as new point clouds arrived. From the current occupancy map, we computed a Euclidean Signed Distance Field (ESDF) via a Euclidean distance transform\footnote{\url{https://docs.scipy.org/doc/scipy/reference/generated/scipy.ndimage.distance_transform_edt.html}}, and obtained its spatial derivatives using NumPy's gradient function\footnote{\url{https://numpy.org/doc/stable/reference/generated/numpy.gradient.html}}. The gradient operator estimates partial derivatives along each axis using central differences in the interior and forward or backward finite differences near boundaries.

Trajectory optimization was formulated as \eqref{eq: opt_problem} and solved with the SLSQP solver in \textit{NLopt} library\footnote{\url{https://nlopt.readthedocs.io/en/latest}}. Since collision avoidance is our primary concern, we promoted convergence toward collision-free trajectories by incorporating the collision-avoidance function $f_\text{collision}(\boldsymbol{C})$ into the objective of \eqref{eq: opt_problem}, weighting the energy and collision terms at a ratio of 1:1000. Following common practice \cite{zucker_chomp_2013, marcucci_motion_2023}, we initialized the optimization variables $\boldsymbol{C}$ to the values that correspond to the minimum-energy trajectory without enforcing constraints. The solver terminated when the relative objective change fell below $f_{\text{tol}}$ or when runtime exceeded 10 seconds.

For reproducibility, the main robot and algorithm parameters are summarized in Table~\ref{tab: params}. 
For the generalized transition velocity $\alpha_v$, increasing its value produces a shorter transition duration, which tightens the velocity constraints and reduces the feasibility margin. If $\alpha_v$ is too large, planning can fail because satisfying the motion within the shortened duration would require violating the velocity limits. A smaller $\alpha_v$ yields slower and more conservative trajectories with a larger feasibility margin, at the cost of longer traversal time. Based on this trade-off, we used $\alpha_v=0.3$ in the simulation.

In each trial, once a global target was set, trajectory planning was performed to generate a global trajectory. The robot then executed this trajectory by tracking the corresponding control commands. These commands were extracted from the trajectory in real time and published to the controller for tracking. Since the trajectory is analytical, the command frequency remains configurable as long as it is supported by the controller. 

In our case, the desired commands comprise CoG-frame pose and velocity---obtained via forward kinematics from the configuration trajectory---and joint angle and velocity for each joint. All commands were published at \SI{40}{\hertz}.

\begin{figure*}[t]
    \setlength{\abovecaptionskip}{-2pt} 
    \centerline{\includegraphics[width=1\linewidth]{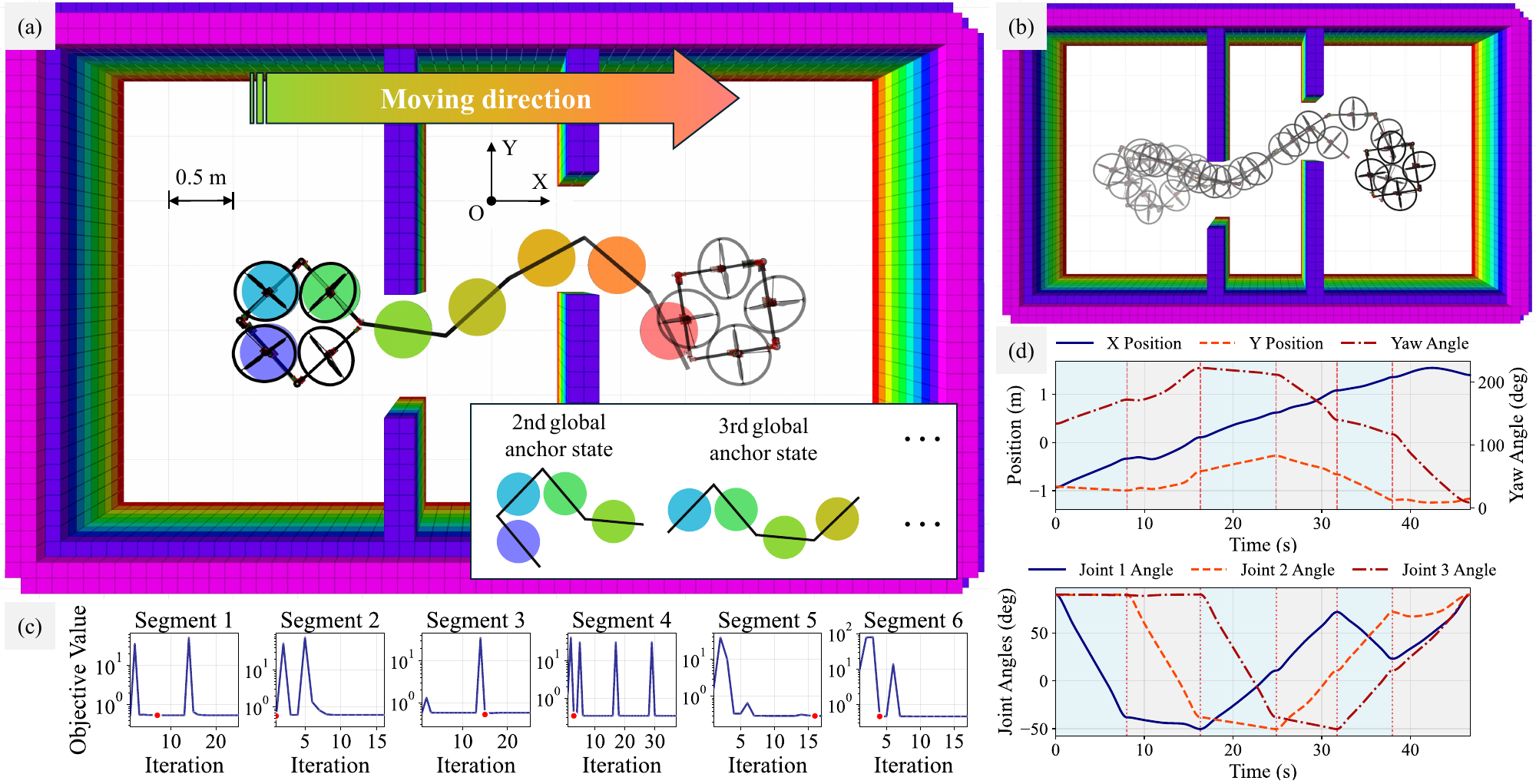}}
    \caption{Simulation of a floating-base multi-link robot navigating through a confined environment with narrow passages. (a) Illustration of the global anchor states, each specifying a complete robot configuration, with the global initial and target states serving as the first and last global anchor states. There were seven anchor states in total. (b) Snapshots of the robot along the flight trajectory. (c) Changing histories of the objective function for all six trajectory segments, where red dots mark the minimum values of the objective. (d) Desired trajectories of the root link position and yaw angle (top) and the joint angles (bottom), with background shading indicating different planning segments.}
    \label{fig: sim_example}
    \vspace{-10pt}
\end{figure*}

\subsection{Dual-Gap-Passing Example}
We first presented an example in which the floating-base multi-link robot passed through two consecutive gaps using the hierarchical trajectory framework, as illustrated in Fig. \ref{fig: sim_example}. Each gap was \SI{0.8}{\meter} in width, and the distance between the centers of the two gaps along the $x$-direction was \SI{1.2}{\meter}. In this example, we set $f_{\text{tol}} = 10^{-5}$. The results in Fig. \ref{fig: sim_example} demonstrate that the planner effectively coordinated the root motion and articulated joints, allowing the robot to traverse confined spaces that would be infeasible for a rigid counterpart.

As shown in Fig. \ref{fig: sim_example}, consecutive anchor states were constructed so that the corresponding local planning problems shared three overlapping links. As a result, each local optimization was solved between two nearby boundary configurations that already satisfied collision-avoidance and controllability constraints. This overlap limited large topological changes within each segment and was the main motivation for decomposing the global problem into multiple local segments. Section \ref{sec:Ablation Study and Benchmark Testing} demonstrated that this decomposition improved the tractability of the overall planning problem.

\begin{table}[t]
\caption{Robot and algorithm parameters}
\label{tab: params}
\centering
\setlength{\tabcolsep}{6pt} 
\renewcommand{\arraystretch}{1.15}
\begin{tabular}{l l | l l | l l}
\toprule
\textbf{Param.} & \textbf{Value} & \textbf{Param.} & \textbf{Value} & \textbf{Param.} & \textbf{Value} \\
\midrule

\multicolumn{6}{c}{\textit{Robot}} \\
\cmidrule(lr){1-6}
$D$ & $6$ & $R_p$ & \SI{0.2025}{\meter} & $\theta_{\min}$ & \SI{-90}{\degree} \\
$m$ & $2$ & $n_j$, $n_r$ & $3,\ 4$ & $\theta_{\max}$ & \SI{90}{\degree} \\
$L$ & \SI{0.6}{\meter} & $\delta_\tau$ & \SI{0.001}{\newton\meter} & $\kappa$ & \SI{-0.0182}{\meter} \\
\midrule

\multicolumn{6}{c}{\textit{Trajectory Planning}} \\
\cmidrule(lr){1-6}
$N$ & $5$ & $v_{\max}$ & \SI{1.0}{\meter\per\second} & $n_{\theta}$ & $60$ \\
$p$ & $3$ & $\omega_{\max}$ & \SI{0.5}{\radian\per\second} & $\alpha_K$ & $100$ \\
$\delta_{\text{collision}}$ & \SI{0.05}{\meter} & \multicolumn{4}{c}{\textit{Resolution of A* discretization:} \SI{0.1}{\meter}} \\

\bottomrule
\end{tabular}
\vspace{-10pt}
\end{table}

\subsection{Ablation Study and Benchmark Testing}
\label{sec:Ablation Study and Benchmark Testing}

\subsubsection{\textbf{Algorithm Settings}}
The proposed framework combines three key features: (1) global anchor states that decompose the overall planning problem into independently solvable segments, (2) local trajectory planning between each pair of local initial and target states, and (3) parallel computation across segments. To assess the contribution of each feature, we performed an ablation study with three degenerate variants, each disabling one feature. For this simulation, we used $f_{\text{tol}} = 10^{-3}$.

For broader evaluation, we also benchmarked against existing methods. As far as we are aware, no prior work has provided a general motion planning framework for floating-base multi-link robots to the same extent as our approach. Nonetheless, solutions exist for specific tasks, such as maneuvering through a single gap, which can be regarded as a subproblem of our setting. To evaluate performance in this case, we compared with a Differential-Kinematics-based (DK-based) planner \cite{zhao_online_2020}. Although a sampling-based method \cite{zhao_transformable_2016} also exists, it is reported in the original study to have prohibitively long computation times (on the order of hours); thus, we excluded it from our comparison.

The five algorithms included in the evaluation were summarized as follows:

a) \textbf{\textit{Ours}}: the full algorithm with all features.

b) \textbf{\textit{w/o AS}} --- without global anchor states: the planning problem was solved as a whole without segmentation.

c) \textbf{\textit{w/o LP}} --- without local planning: a linear transition was applied between each pair of local initial and target states.

d) \textbf{\textit{w/o PC}} --- without parallel computation: all segments were solved sequentially.

e) \textbf{\textit{DK}}: the Differential-Kinematics-based planner \cite{zhao_online_2020}.

\subsubsection{\textbf{Task Settings and Metrics}}
The task required the robot to traverse a gap of width \SI{0.7}{\meter}. The robot started in a square configuration with all three joint angles set to \SI{90}{\degree}. To pass through the gap, the robot must deform accordingly. We generated 200 random instances by sampling the initial distance to the gap: the $x$-position of the root link was uniformly drawn from \SI{0.5}{\meter} to \SI{1.32}{\meter}, while the $y$-position and yaw angle were fixed at \SI{0.25}{\meter} and \SI{5}{\degree}, respectively. Each method was tested on all instances.

We report three metrics:

a) \textit{Success rate}: a trial succeeds if the robot reaches the specified target state without collision. For the DK-based planner, which is designed for gap passing and does not support arbitrary target states, success is defined as clearing the gap without collision.

b) \textit{Computation time}: elapsed time from target specification to producing a feasible trajectory. It is indispensable for comparing algorithms with similarly high success rates. When success rates are low, reliability dominates, leaving computation time with little interpretive value.

c) \textit{Trajectory length}: we report both the root trajectory length of $\boldsymbol{p}_r \in \mathbb{R}^m$ and the generalized trajectory length of the full configuration $\boldsymbol{q} \in \mathbb{R}^D$. Length statistics are reported only when the produced solution is comparable to our continuous-time, target-reaching formulation.

\begin{figure}[t]
    \setlength{\abovecaptionskip}{-2pt}
    \centerline{\includegraphics[width=1\linewidth]{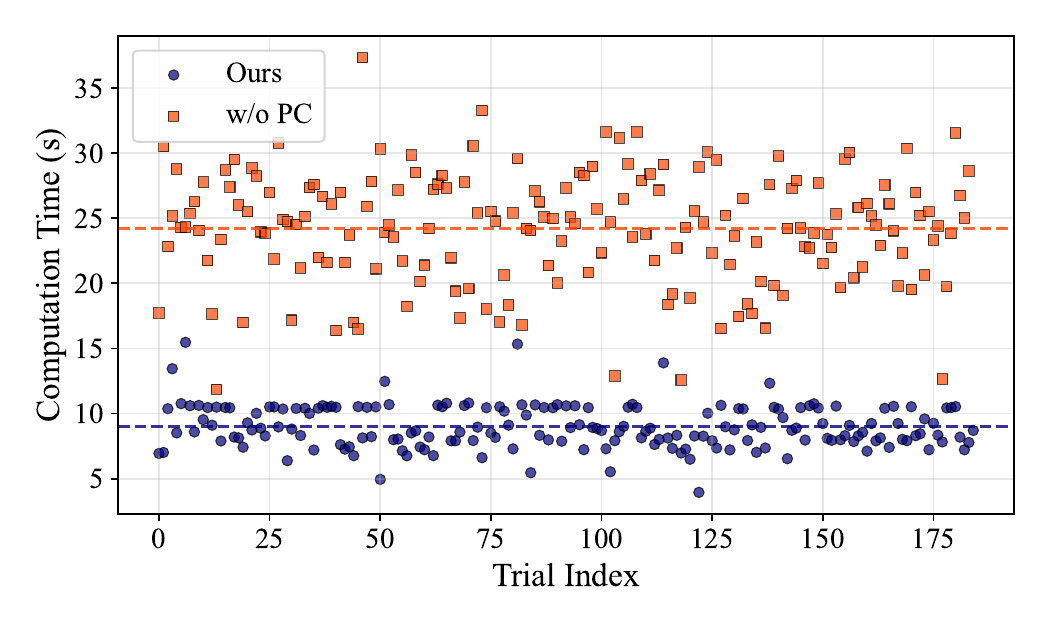}}
    \caption{Computation time comparison between our planner and its sequential counterpart (w/o PC). Parallel computation substantially reduces runtime while preserving the same planning formulation.}
    \label{fig: comp_time_vs}
    \vspace{-16pt}
\end{figure}

\begin{figure*}[t]
    \centering
    \subfloat[Mission setup]{
        \includegraphics[width=0.3\linewidth]{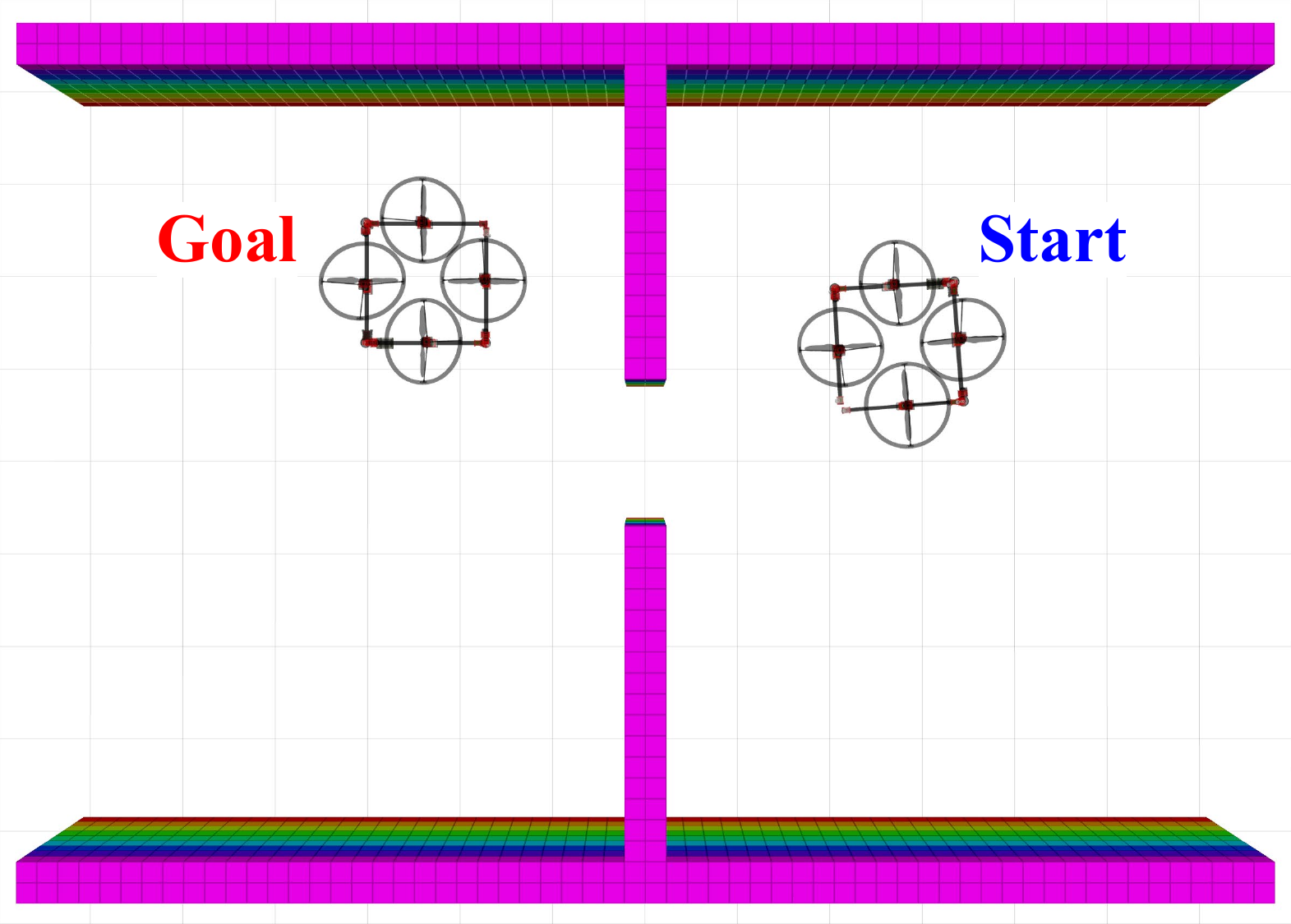}
        \label{fig: ablation_mission}}
    \hfil
    \subfloat[Ours]{
        \includegraphics[width=0.3\linewidth]{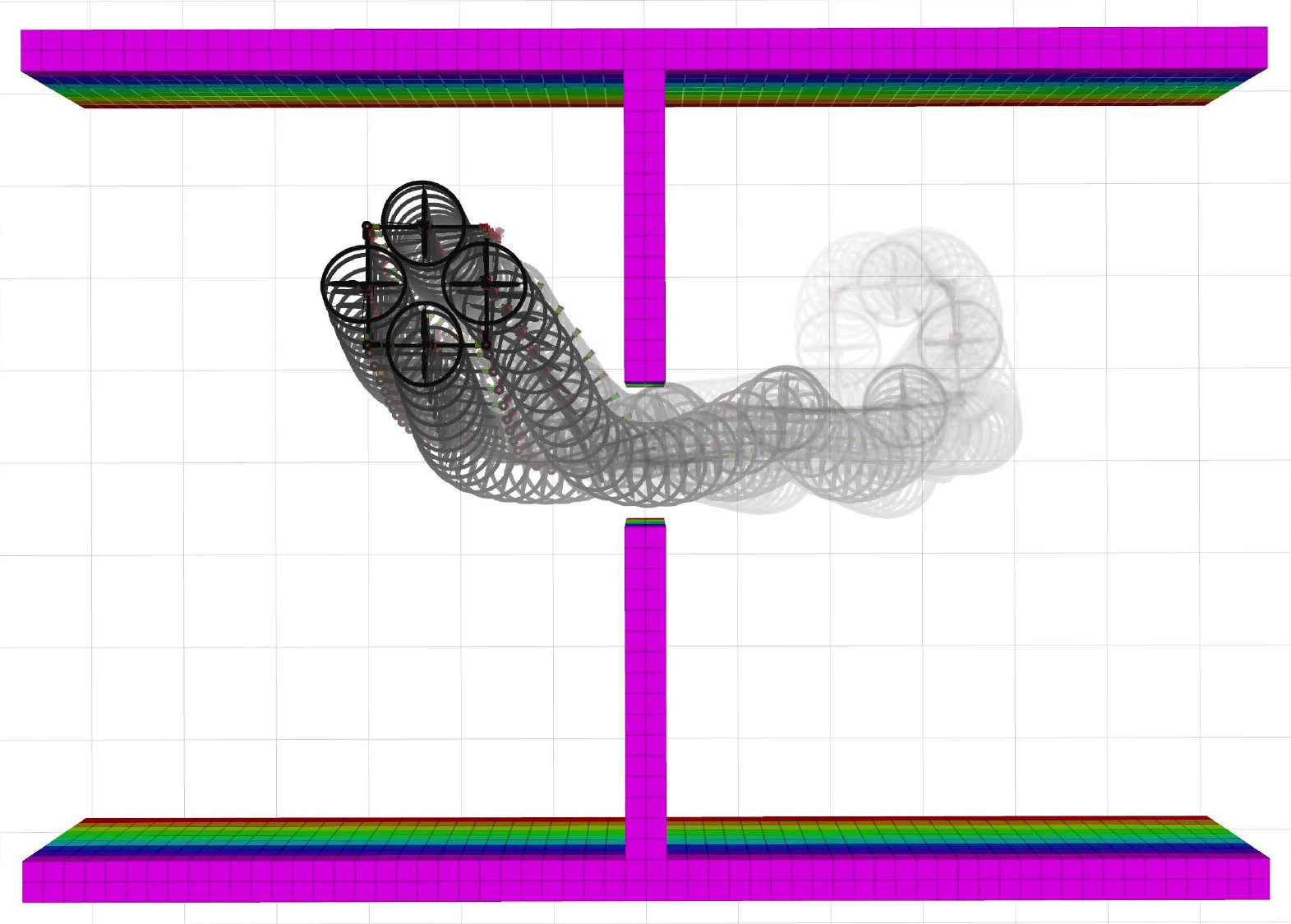}
        \label{fig: Ours}}
    \hfil
    \subfloat[DK-based planner]{
        \includegraphics[width=0.3\linewidth]{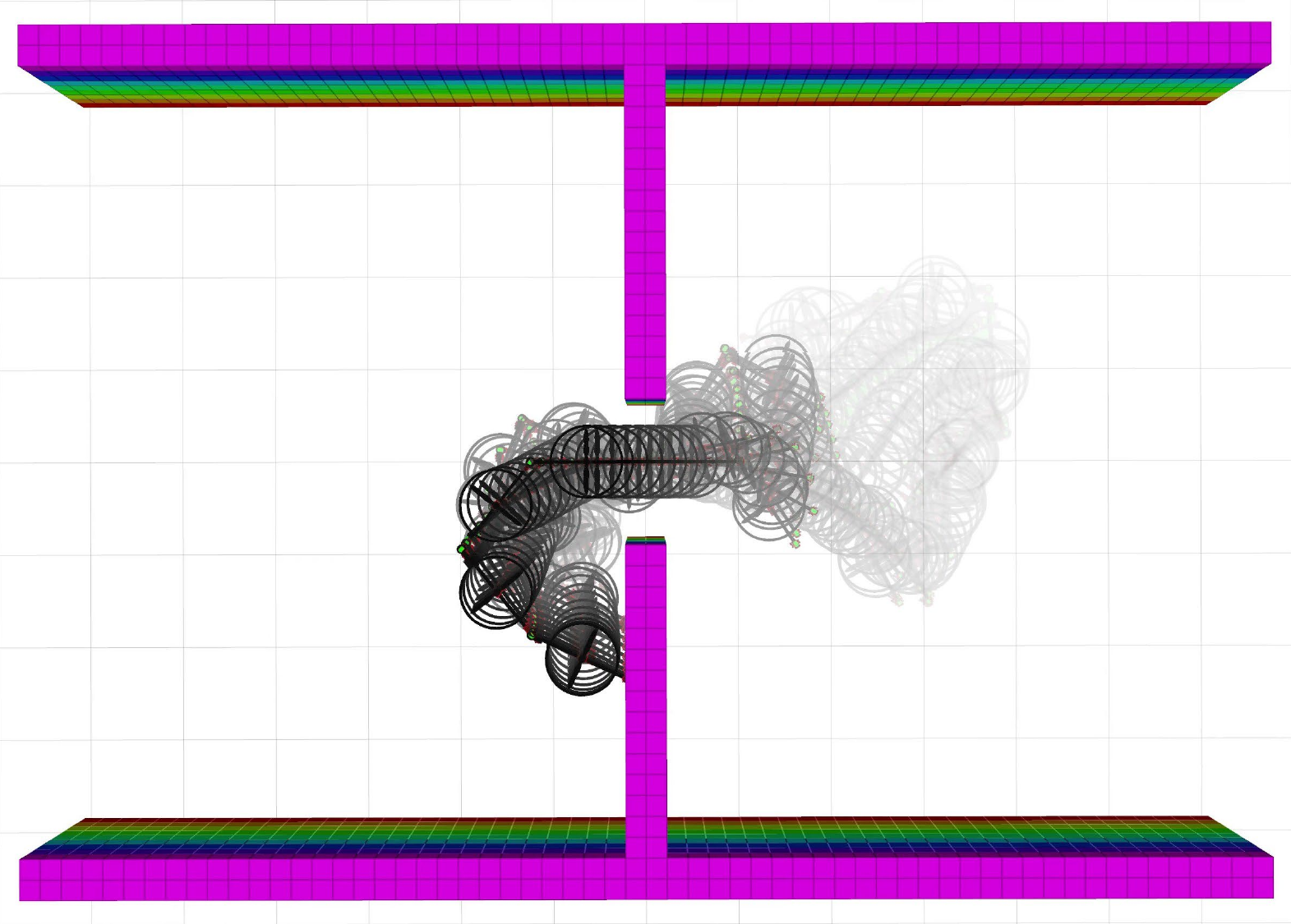}
        \label{fig: DK}}
    \hfil
    \subfloat[w/o AS]{
        \includegraphics[width=0.3\linewidth]{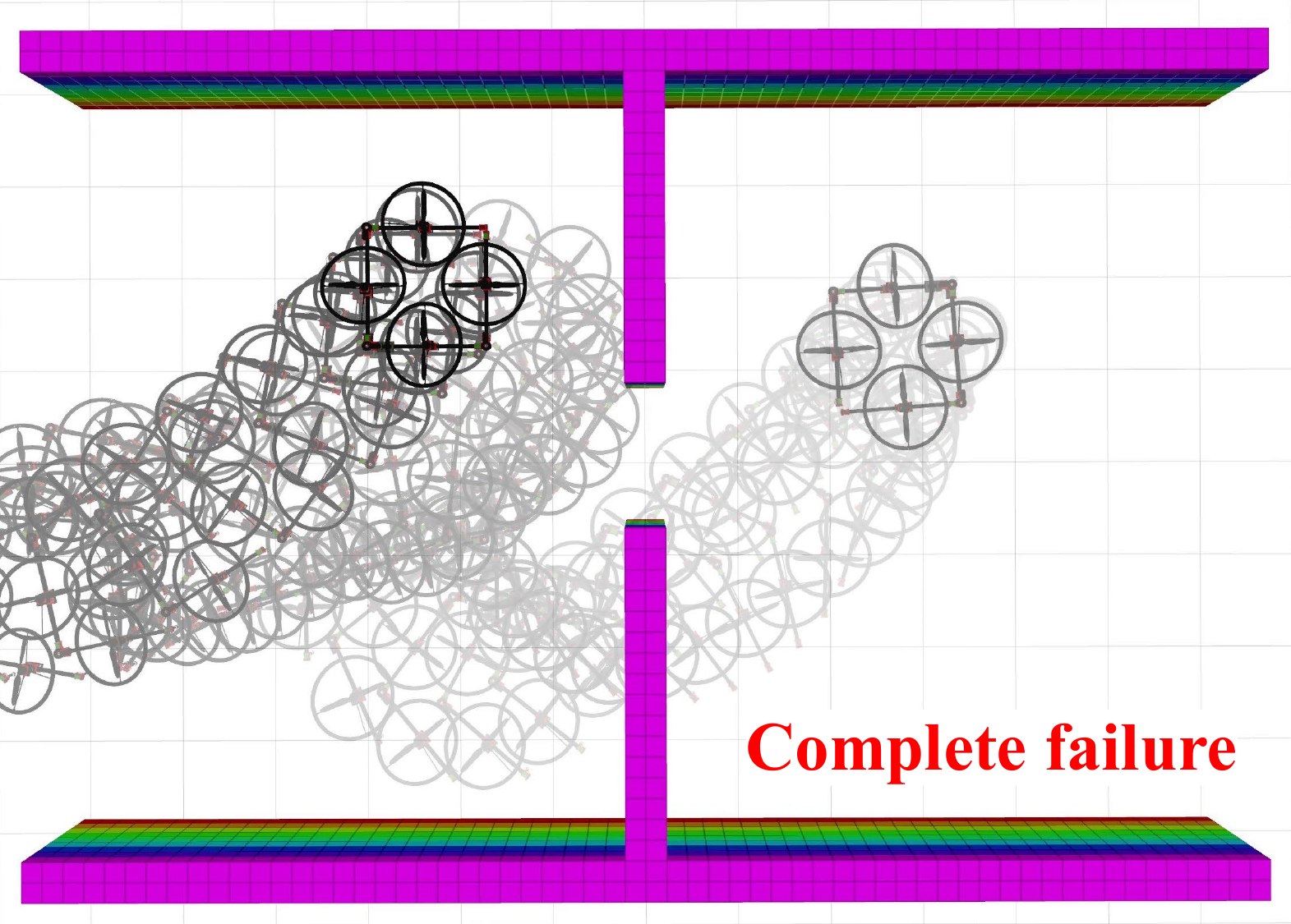}
        \label{fig: wo_AS}}
    \hfil
    \subfloat[w/o LP]{
        \includegraphics[width=0.3\linewidth]{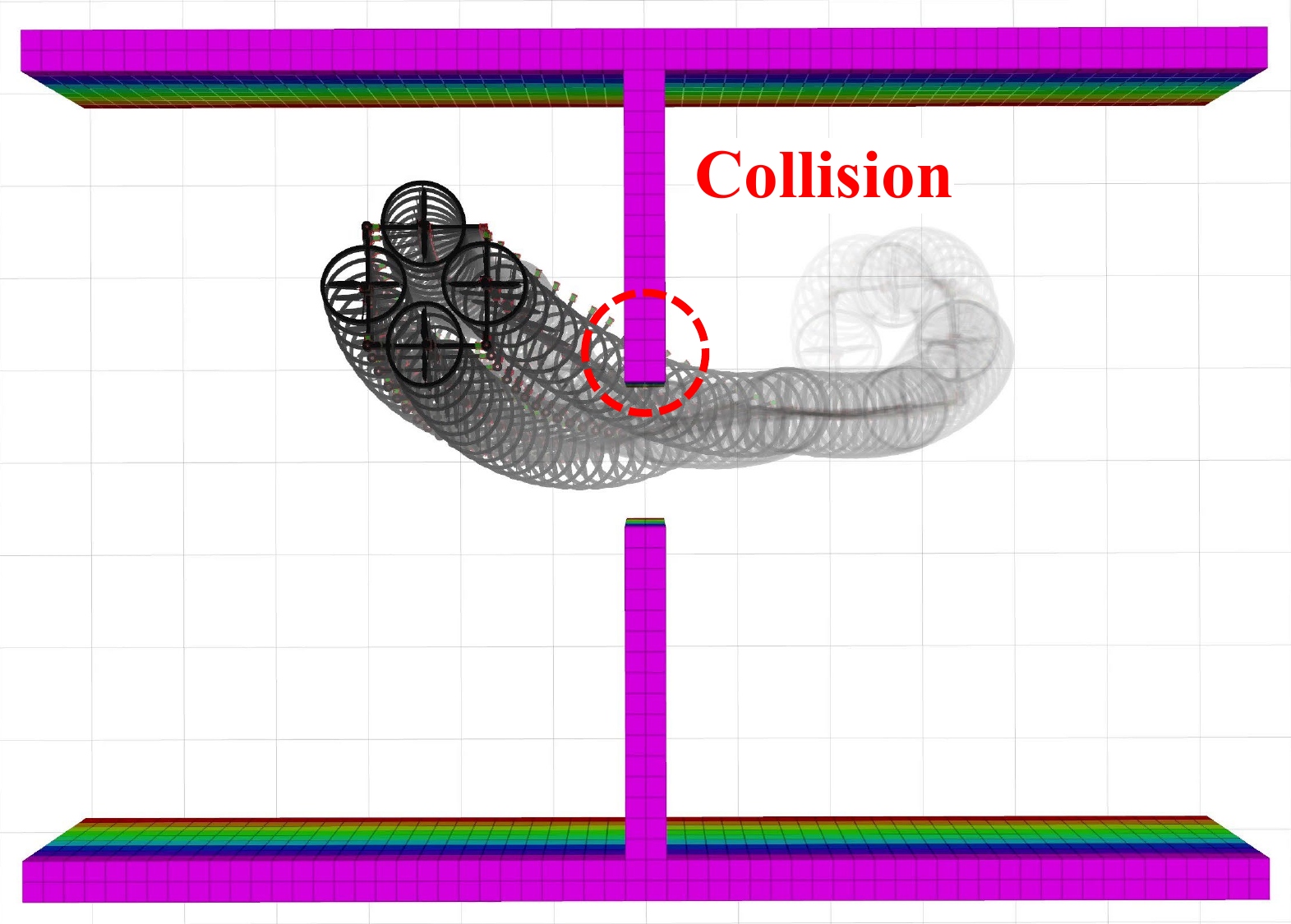}
        \label{fig: wo_LP}}
    \hfil
    \subfloat[The anchor states in the trial w/o LP]{
        \includegraphics[width=0.3\linewidth]{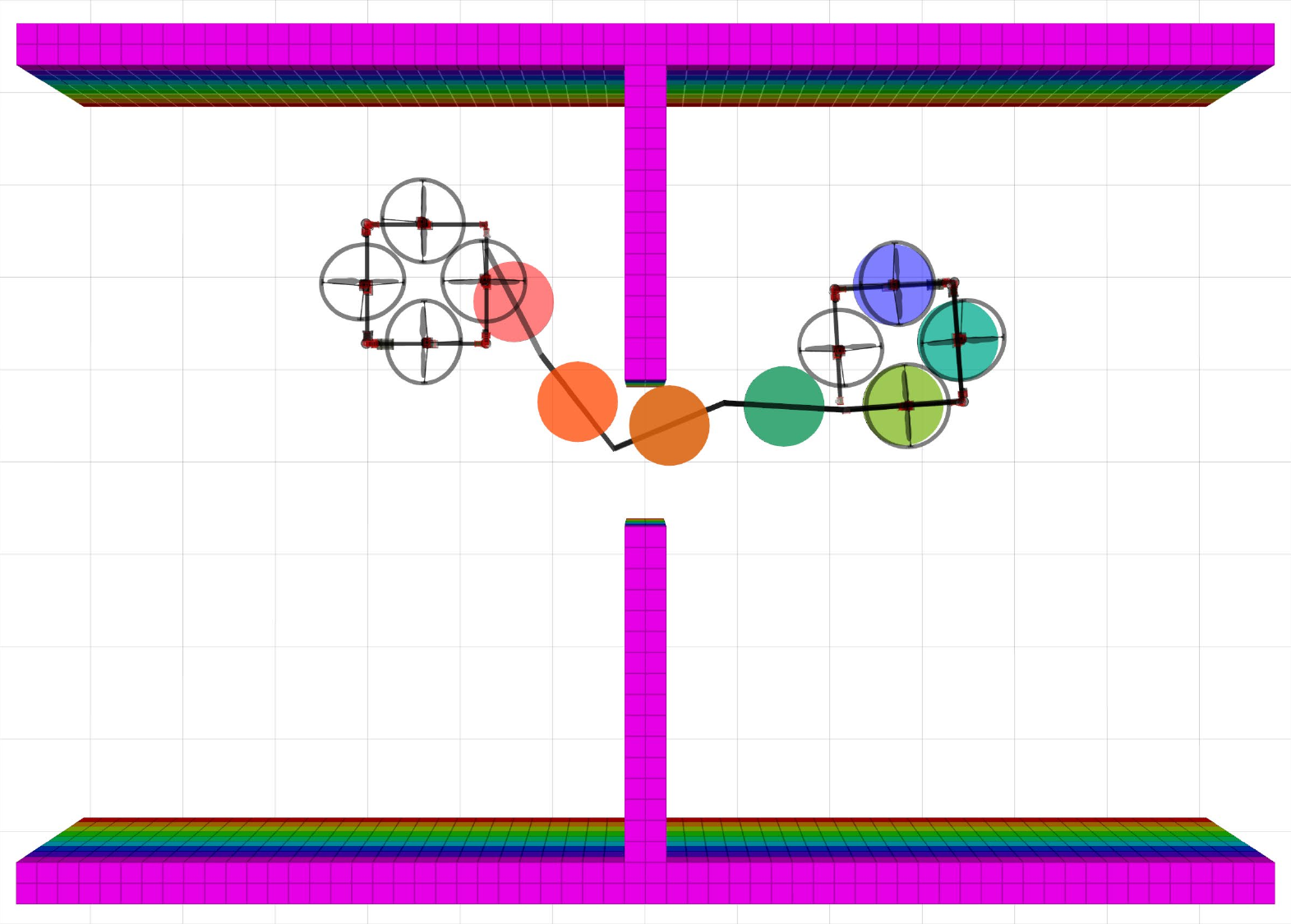}
        \label{fig: wo_LP_anchor_states}}
    \hfil
    \caption{Comparison of planned motions from different algorithms in one trial. (a) Mission setup with start and goal states. (b) The proposed planner generated a feasible trajectory through the narrow gap. (c) The DK-based planner produced a motion that passed the gap but lacked the flexibility to reach the designated target state. (d) The variant without global anchor states (w/o AS) completely failed to find a feasible trajectory. (e) The variant without local planning (w/o LP) produced a trajectory that collides with the obstacle. (f) The anchor states obtained in (e) were individually feasible, but simple linear interpolation between them led to a collision, underscoring the necessity of local planning within the framework.}
    \label{fig: comp_example}
\end{figure*}

\begin{table*}[!t]
\centering
\caption{Results of ablation and benchmark tests over 200 trials}
\renewcommand{\arraystretch}{1.10}
\setlength{\tabcolsep}{6pt}
\label{tab: sim_data}
\begin{tabular}{cccccccc}
\toprule
\multirow{2}{*}{\textbf{Algorithm}}
& \multirow{2}{*}{\textbf{Success Rate$^{\ast}$} (\si{\percent}) $\uparrow$}
& \multicolumn{2}{c}{\textbf{Computation Time} (\si{\second}) $\downarrow$}
& \multicolumn{2}{c}{\textbf{Root Trajectory Length} (\si{\meter}) $\downarrow$} 
& \multicolumn{2}{c}{\textbf{Generalized Trajectory Length} $\downarrow$} \\
\cmidrule(lr){3-4}\cmidrule(lr){5-6}\cmidrule(lr){7-8}
& & Average & Standard Deviation & Average & Standard Deviation & Average & Standard Deviation \\
\midrule
\textbf{Ours}
& \textbf{\textcolor{green!60!black}{\SI{92.5}{}}}
& \SI{9.01}{}  & \SI{1.66}{}
& \SI{2.84}{}  & \SI{0.302}{}
& \SI{10.2}{}  & \SI{0.601}{} \\
w/o AS
& \textbf{\textcolor{red!80!black}{\SI{0.0}{}}}
& -- & -- & -- & -- & -- & -- \\
w/o LP
& \textbf{\textcolor{red!80!black}{\SI{14.0}{}}}
& \SI{0.307}{} & \SI{0.0240}{}
& -- & -- & -- & -- \\
w/o PC
& \textbf{\textcolor{green!60!black}{\SI{92.0}{}}}
& \textbf{\textcolor{red!80!black}{\SI{24.2}{}}}
& \textbf{\textcolor{red!80!black}{\SI{4.30}{}}}
& \SI{2.87}{}  & \SI{0.417}{}
& \SI{10.3}{}  & \SI{0.667}{} \\
DK \cite{zhao_online_2020}
& \textbf{\textcolor{red!80!black}{\SI{13.0}{}}}
& \textbf{\textcolor{green!60!black}{\SI{0.0580}{}}}
& \textbf{\textcolor{green!60!black}{\SI{0.00353}{}}}
& -- & -- & -- & -- \\
\bottomrule
\end{tabular}
\\[0.6ex]
\footnotesize{$^{\ast}$Primary metric. Computation time and path-length metrics are supplementary.}
\vspace{-10pt}
\end{table*}

\subsubsection{\textbf{Results and Discussions}}
Fig.~\ref{fig: comp_time_vs} isolates the impact of parallel computation by comparing our method against its sequential counterpart under similarly high success rates. Fig.~\ref{fig: comp_example} visualizes representative motions, and Table~\ref{tab: sim_data} reports the quantitative results, where success rate is the primary metric and computation time and trajectory length are supplementary.

Overall, the full framework achieved a success rate of \SI{92.5}{\percent}. Among successful methods, it maintained moderate computation time (\SI{9.01}{\second} on average) and produced consistent trajectory lengths (root: \SI{2.84}{\meter} on average; generalized: \SI{10.2} on average), indicating reliable target-reaching motion generation.

\textbf{Effect of Global Anchor States:}
Removing global anchor states (w/o AS) resulted in failure across all $200$ trials, since the optimization could not converge when solved as a whole, as shown in Fig. \ref{fig: wo_AS}. We assume the reason was that direct global planning produced an extremely nonconvex problem with numerous local minima, where gradient-based solvers easily stalled. In contrast, introducing global anchor states decomposed the problem into smaller, well-posed segments, each centered on local feasibility around a guided anchor, enabling reliable convergence.

\textbf{Effect of Local Planning:}
Without local planning (w/o LP), the success rate dropped to \SI{14.0}{\percent}. Fig.~\ref{fig: wo_LP} and \ref{fig: wo_LP_anchor_states} show that anchor states could remain individually feasible, yet naive linear interpolation between them introduced collisions because it ignored continuous-time feasibility and obstacle clearance. This explained why w/o LP might appear computationally cheap but was unreliable: it skipped the key step that enforced feasibility between discrete states.

\textbf{Effect of Parallel Computation:}
Disabling parallel computation (w/o PC) preserved the success rate (\SI{92.0}{\percent}) and yielded similar trajectory lengths (root: \SI{2.87}{\meter}, generalized: \SI{10.3}), confirming that parallelization did not change solution quality. In contrast, the average computation time increased to \SI{24.2}{\second} with a larger variance. Fig.~\ref{fig: comp_time_vs} highlights this gap directly, confirming that parallelization improved computational efficiency.

\textbf{Comparison with DK Baseline:}
The DK-based planner was extremely fast (\SI{0.0580}{\second}) due to its quadratic-programming structure, but its success rate was only \SI{13.0}{\percent}. This behavior was consistent with its greedy, locally optimal updates, which were sensitive to local minima. Moreover, as shown in Fig.~\ref{fig: DK}, it was tailored to gap passing using a manually specified gap pose and did not support arbitrary target states, limiting its generality for broader trajectory planning. Consequently, its outputs were not directly comparable in terms of target-reaching trajectory length.

\textbf{Summary of Findings:}
These results confirm that each feature of the hierarchical framework is indispensable: global anchor states enable tractable decomposition, local planning maintains feasibility, and parallelization enhances efficiency. Compared with a specialized gap-passing baseline, the proposed framework provides a more general and reliable trajectory planning capability for floating-base multi-link robots in confined environments.

\textbf{Failure Cases of the Proposed Method:}
We assumed that the remaining failure cases of the proposed method arose from three factors. First, global anchor-state generation was a greedy iterative search and could become trapped in local minima. Second, in some trials, all anchor states were generated, but the local trajectory optimization did not recover a feasible trajectory, since it was initialized by a minimum-energy spline that ignored obstacles, and the resulting nonconvex penalty landscape could admit unfavorable local minima. Third, because the continuous-time controllability constraint was enforced via sampling, violations could still occur between sampled states.

\begin{figure}[t]
    \centering
    \subfloat[Success rate]{
        \includegraphics[width=0.47\linewidth]{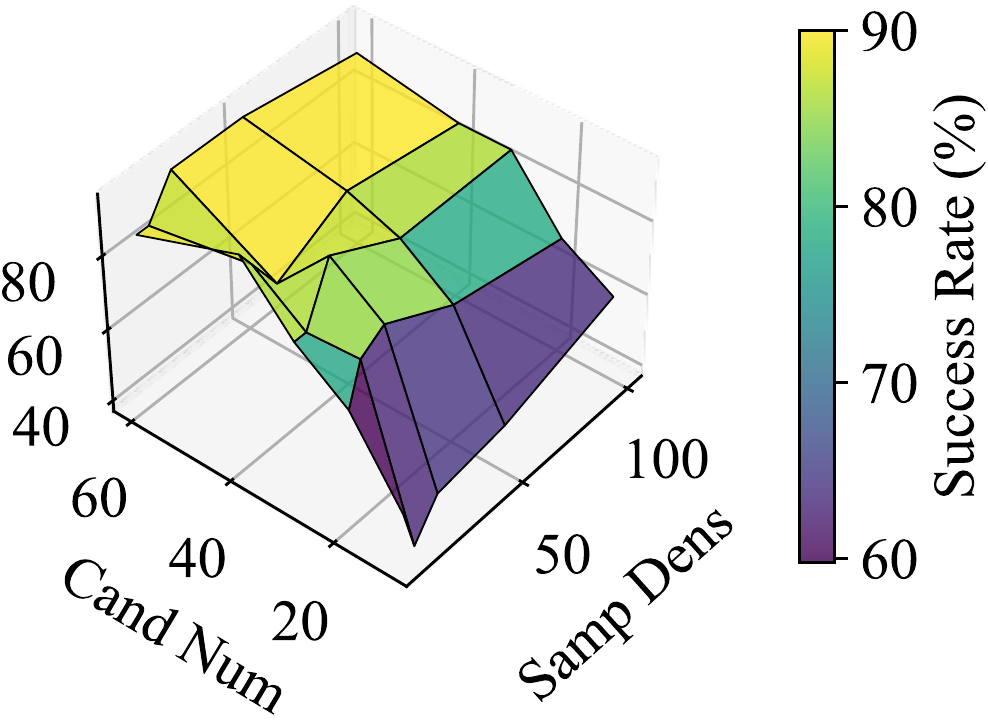}
        \label{fig: hyperparam_success_rate}}
    \hfil
    \subfloat[Average computation time]{
        \includegraphics[width=0.47\linewidth]{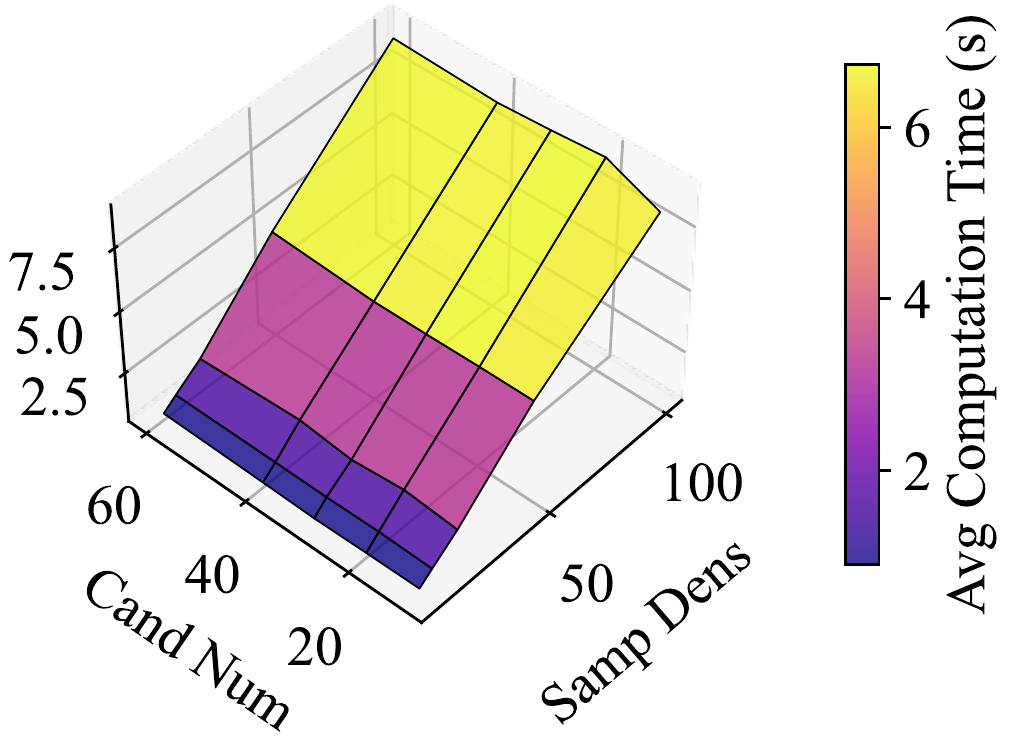}
        \label{fig: hyperparam_time}}
    \hfil
    \caption{Influence of hyperparameters on planning performance. The plots showed the effects of the number of candidate configurations $n_{\theta}$ generated per anchor-state iteration (Cand Num) and the sampling density $\alpha_K$ used for continuous-time constraints (Sample Dens) on (a) the success rate and (b) the average computation time. The results indicated that the success rate was primarily governed by the candidate number, whereas the average computation time was dominated by the sampling density.}
    \label{fig: hyperparam_analysis}
    \vspace{-10pt}
\end{figure}

\begin{figure*}[t]
    \setlength{\abovecaptionskip}{-2pt} 
    \centerline{\includegraphics[width=1\linewidth]{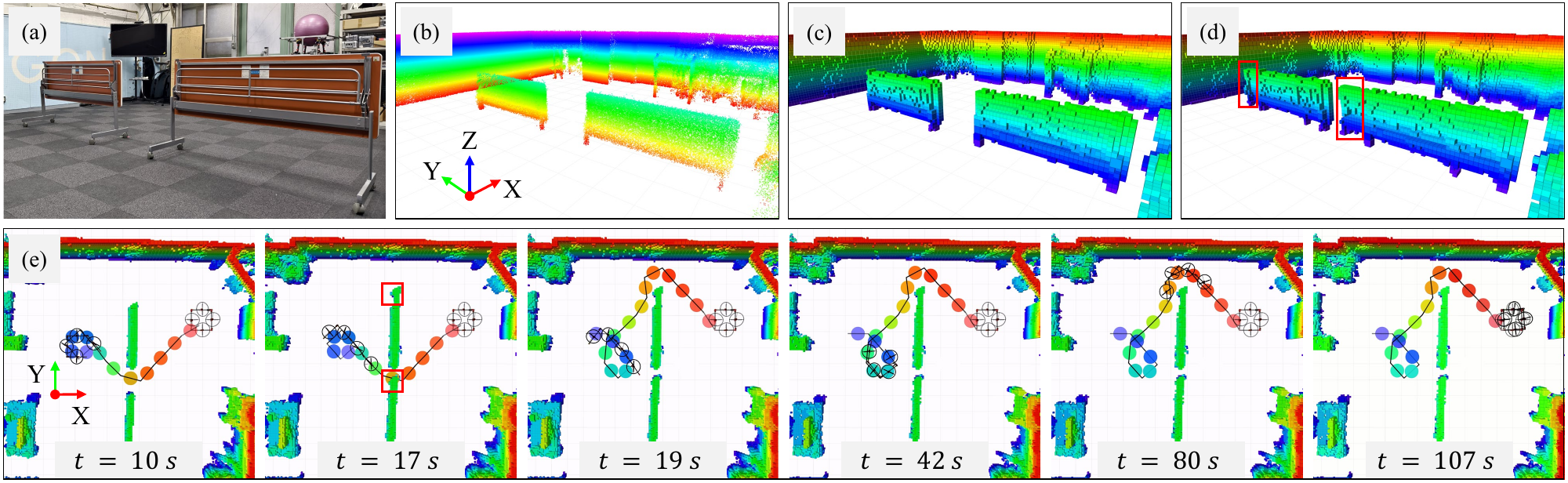}}
    \caption{Evaluation with real-world point clouds under environment changes. A handheld Livox Mid-360 LiDAR was used to capture the scene, and FAST-LIO was used to reconstruct two registered point clouds before and after the obstacles were rearranged. (a) Experimental scene. (b) Point cloud of the initial scene. (c) Online volumetric occupancy map reconstructed from (b) and used for the initial plan. (d) Updated occupancy map after switching the input to the second point cloud, where the original passage was blocked (red boxes). (e) Execution snapshots in Gazebo (time was measured from the onset of motion): the robot started with the initial plan, the map update blocked the original passage at $t=\SI{17}{\second}$, replanning produced a new feasible trajectory at $t=\SI{19}{\second}$, and the robot switched to it and continued to the goal.}
    \label{fig: real_world_pcl_test}
    \vspace{-10pt}
\end{figure*}

\begin{figure}[t]
    \setlength{\abovecaptionskip}{-2pt} 
    \centerline{\includegraphics[width=1\linewidth]{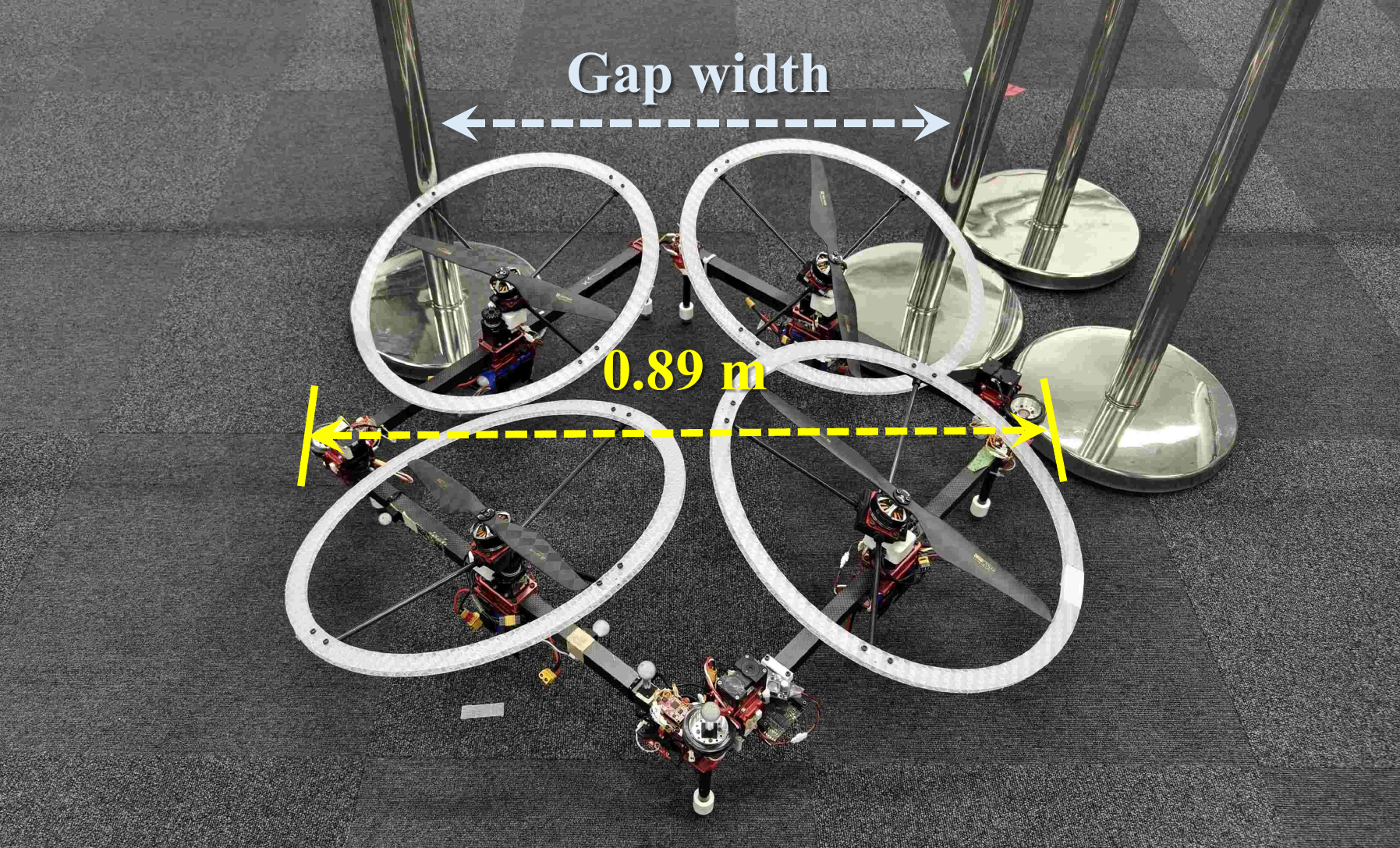}}
    \caption{The robot used in the experiments. In its square configuration, the robot's diameter was \SI{0.89}{\meter}, larger than the gap width, requiring it to transform its shape to maneuver through the gaps.}
    \label{fig: robot_picture}
    \vspace{-12pt}
\end{figure}

\subsection{Parameter Sensitivity Analysis}
We investigated two hyperparameters that govern the success--runtime trade-off of the proposed framework: (i) the number of candidate configurations $n_{\theta}$ generated at each anchor-state iteration, and (ii) the sampling density $\alpha_K$ used to enforce continuous-time constraints. We swept $n_{\theta} \in \{10,20,30,40,60\}$ and $\alpha_K \in \{5,10,20,50,100\}$, yielding 25 parameter combinations. We pre-sampled a fixed set of 20 random mission instances under the same task settings as in Section~\ref{sec:Ablation Study and Benchmark Testing}, and evaluated the same 20 instances for every parameter combination. We reported the success rate and average computation time aggregated over these instances.

Fig.~\ref{fig: hyperparam_analysis} summarizes the results and indicates a stable performance trend with a clear performance--time trade-off. The success rate increased monotonically with $n_{\theta}$ and exhibited a broad plateau at larger $n_{\theta}$, suggesting that the planner was not overly sensitive once a sufficient number of candidates was provided. In contrast, varying $\alpha_K$ had a comparatively minor influence on success within the tested range, but it strongly affected runtime: the average computation time grew monotonically with denser sampling, consistent with the increased number of constraint evaluations induced by larger $K$. Taken together, these results supported using a moderately large $n_{\theta}$ to ensure robustness, while tuning $\alpha_K$ to meet a desired computational budget. Here $n_{\theta}=60$ and $\alpha_K=20$ provided a favorable operating point.

\subsection{Evaluation with Real-World Point Clouds Under Environment Changes}
We evaluated robustness to realistic perception artifacts, planning latency, and environment changes during execution. We captured the scene geometry using a handheld device equipped with a Livox Mid-360 LiDAR and reconstructed registered point clouds using FAST-LIO \cite{xu_fast-lio2_2022}. As shown in Fig.~\ref{fig: real_world_pcl_test}(a)--(b), we first collected a point cloud of the initial scene, then rearranged the obstacles and reconstructed a second point cloud. During evaluation, these two point clouds were published sequentially to emulate an environment change, while mapping from the incoming point cloud was performed online. To improve responsiveness to newly observed obstacles, we set the constraint sampling density to $\alpha_K = 20$.

We conducted a Gazebo simulation using the above real-world point clouds. The robot first planned a trajectory using the map reconstructed from the initial point cloud (Fig.~\ref{fig: real_world_pcl_test}(c)) and started executing it. We then switched the input to the second point cloud, yielding an updated volumetric occupancy map in which the original passage was blocked (Fig.~\ref{fig: real_world_pcl_test}(d), $t=\SI{17}{\second}$). As shown in Fig.~\ref{fig: real_world_pcl_test}(e), replanning produced a new feasible trajectory at $t=\SI{19}{\second}$, and the robot switched to it and continued to the goal. While the replanning time introduced non-negligible latency, the system recovered without failure. Despite the sparsity and outliers of real LiDAR point clouds, the planner remained robust, demonstrating that the proposed pipeline tolerated realistic perception noise and update latency while recovering promptly from environment changes.

\section{Experiments}
\label{Experiments}

\subsection{Experiment Settings}
To evaluate the algorithm in broader real-world scenarios, we built a floating-base multi-link robot inspired by the designs in \cite{zhao_transformable_2018, zhao_singularity-free_2021}, as shown in Fig.~\ref{fig: robot_picture}. Each link employed a 1-DoF yaw-vectorable propeller with a fixed tilt angle, allowing thrust redirection to improve rotational controllability. However, if the deformation trajectory is not properly planned, certain configuration transitions may admit no continuous feasible vectoring trajectory, causing complete loss of controllability and a rapid crash. The robot state was tracked using a motion capture system. The state feedback pipeline follows the setup described in \cite{zhao_transformable_2018}. The planning algorithm ran on the same ground computer used in simulation and communicated with the robot via Wi-Fi within the \textit{ROS} framework. The planner published desired commands as stated in Section \ref{sec: Implementation and Reproducibility}, while a low-level controller executed them onboard using a \textit{Khadas VIM4}.

\begin{figure*}[!t]
    \centering

    \includegraphics[width=\linewidth]{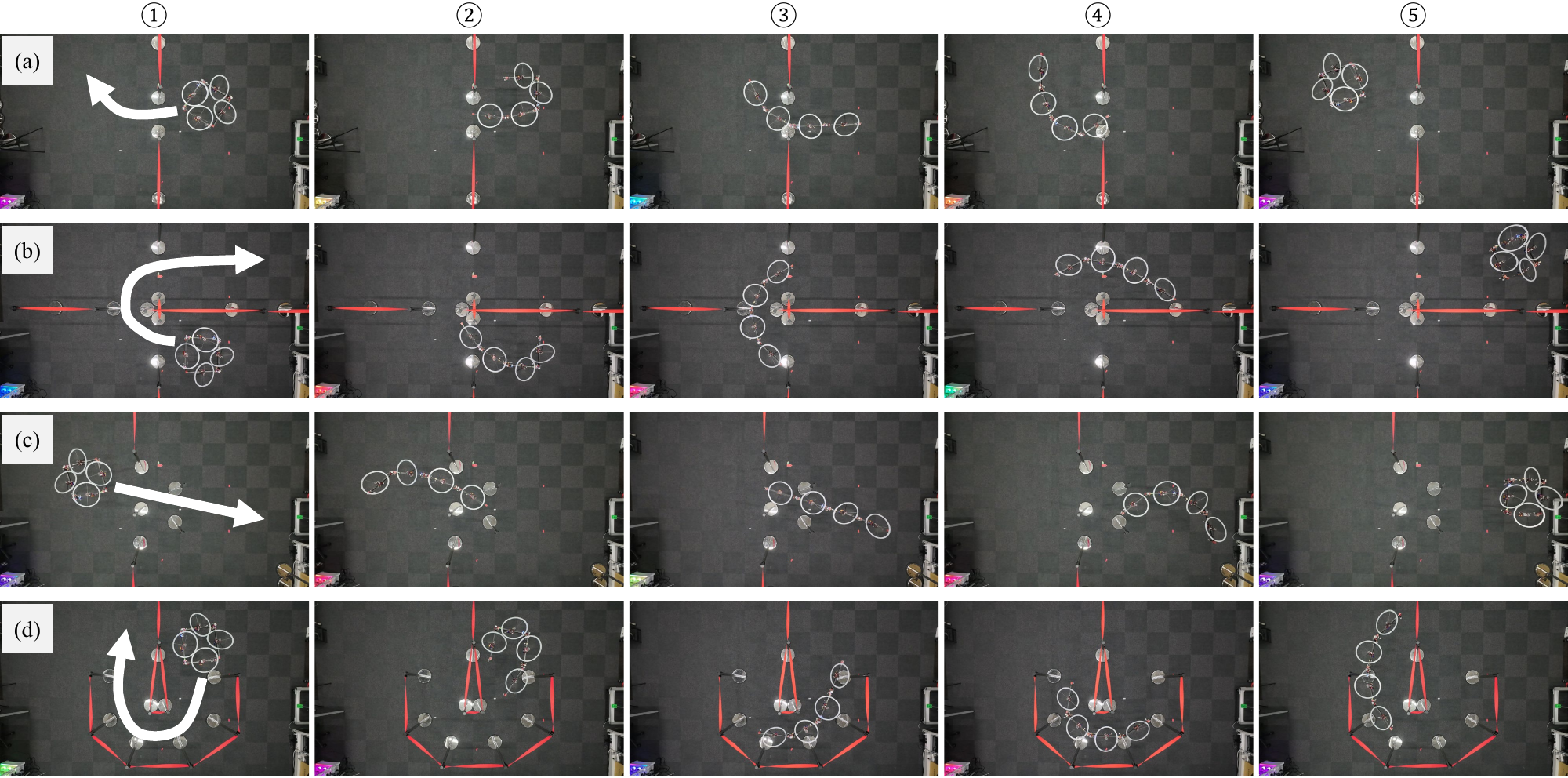}

    \setcounter{subfigure}{4}

    \subfloat[CoG motion in Scene (a)]{
        \includegraphics[width=0.23\linewidth]{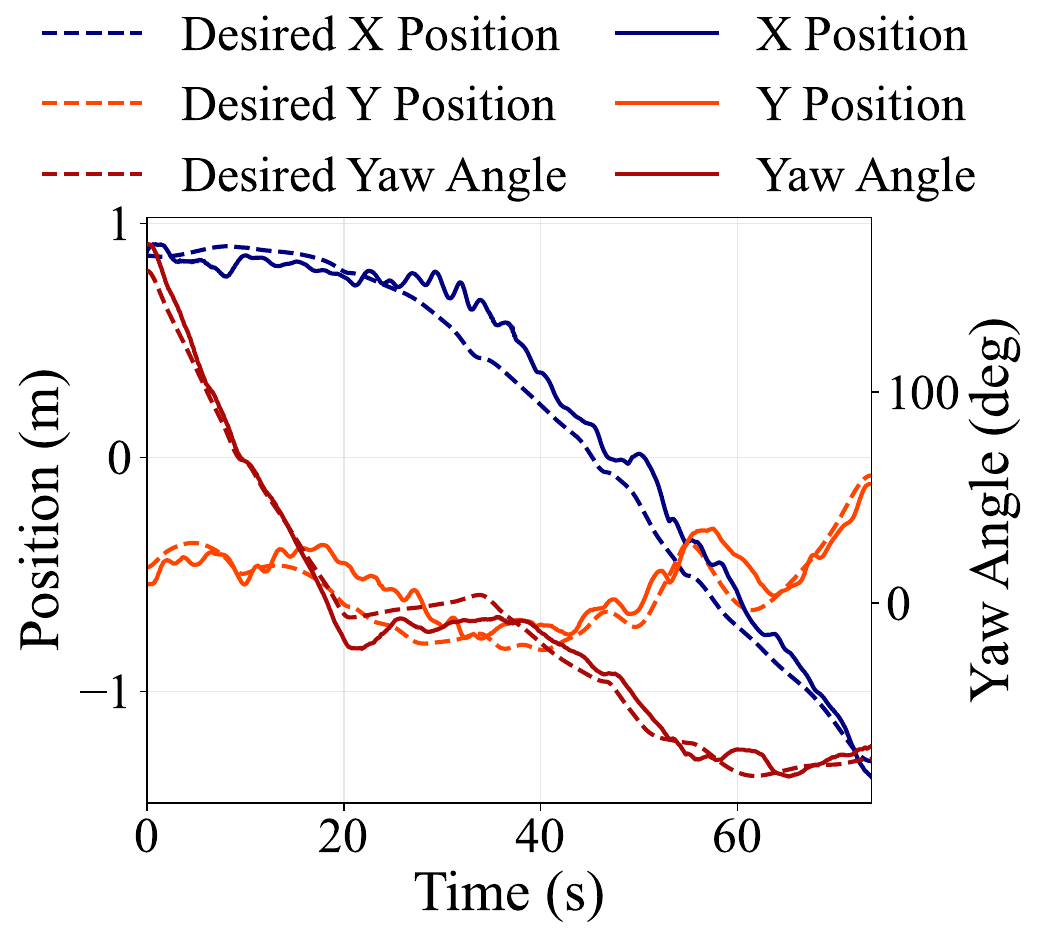}
        \label{fig: gap_cog}}
    \hfil
    \subfloat[CoG motion in Scene (b)]{
        \includegraphics[width=0.23\linewidth]{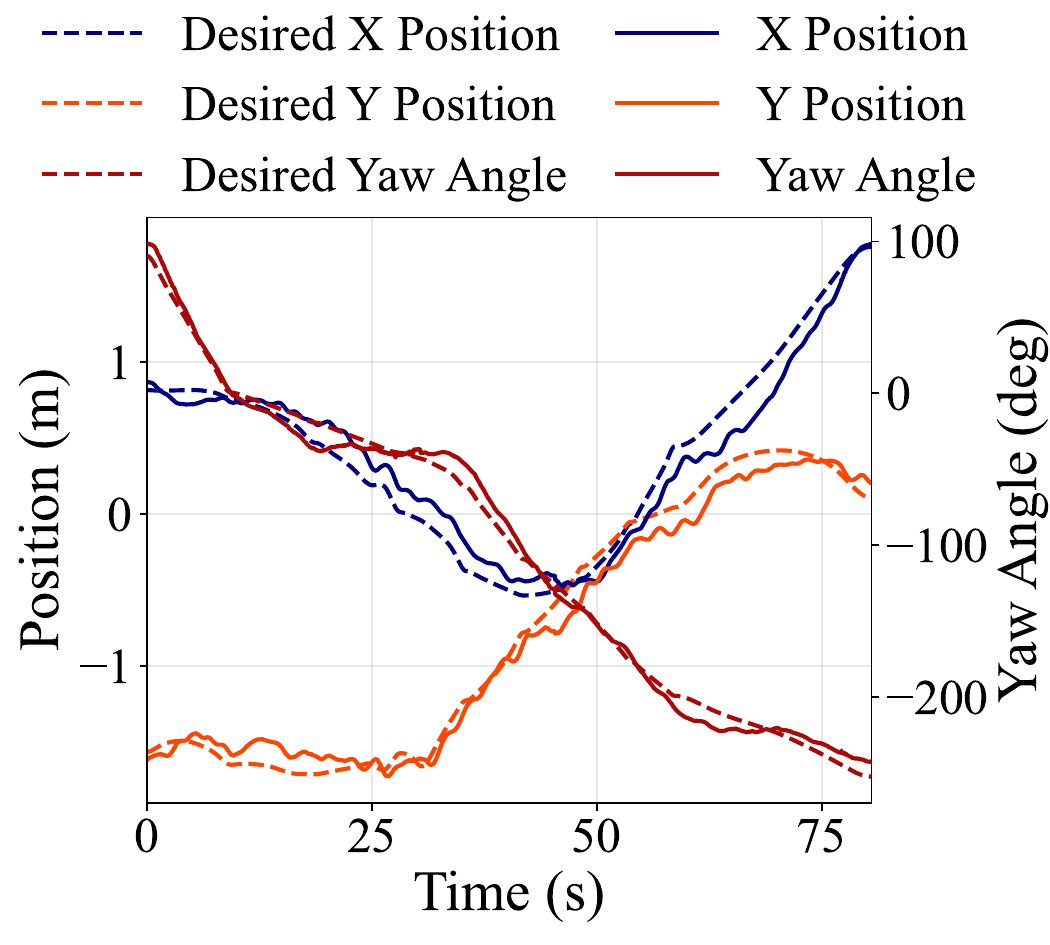}
        \label{fig: mix_gap_cog}}
    \hfil
    \subfloat[CoG motion in Scene (c)]{
        \includegraphics[width=0.23\linewidth]{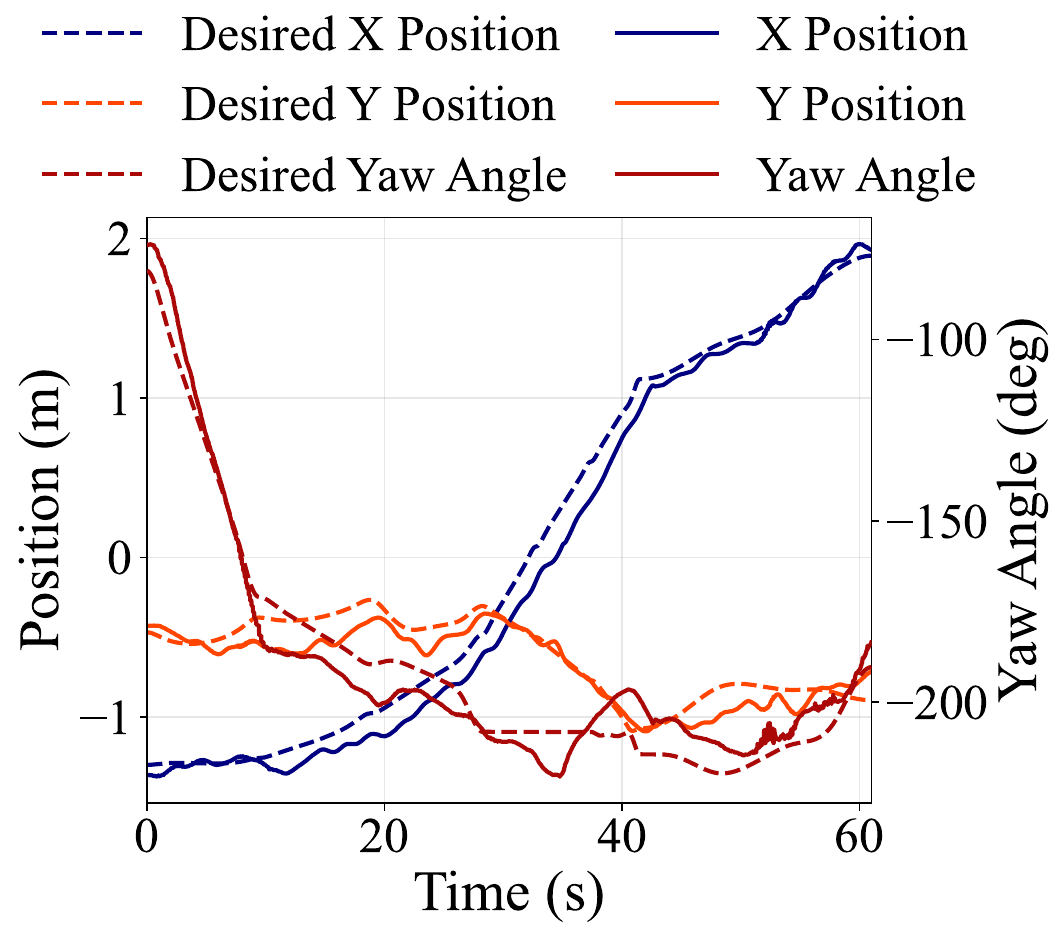}
        \label{fig: poles_cog}}
    \hfil
    \subfloat[CoG motion in Scene (d)]{
        \includegraphics[width=0.23\linewidth]{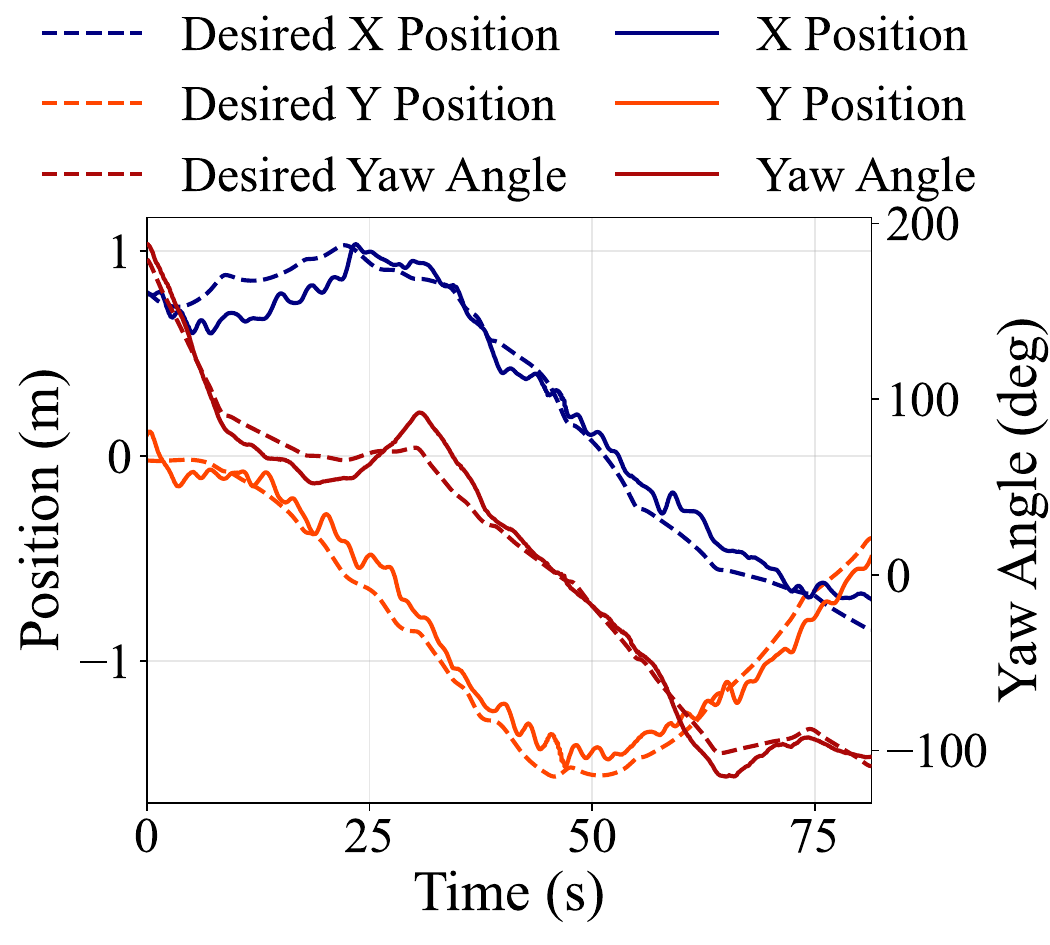}
        \label{fig: turn_cog}}
    \hfil
    \subfloat[Joint angles in Scene (a)]{
        \includegraphics[width=0.23\linewidth]{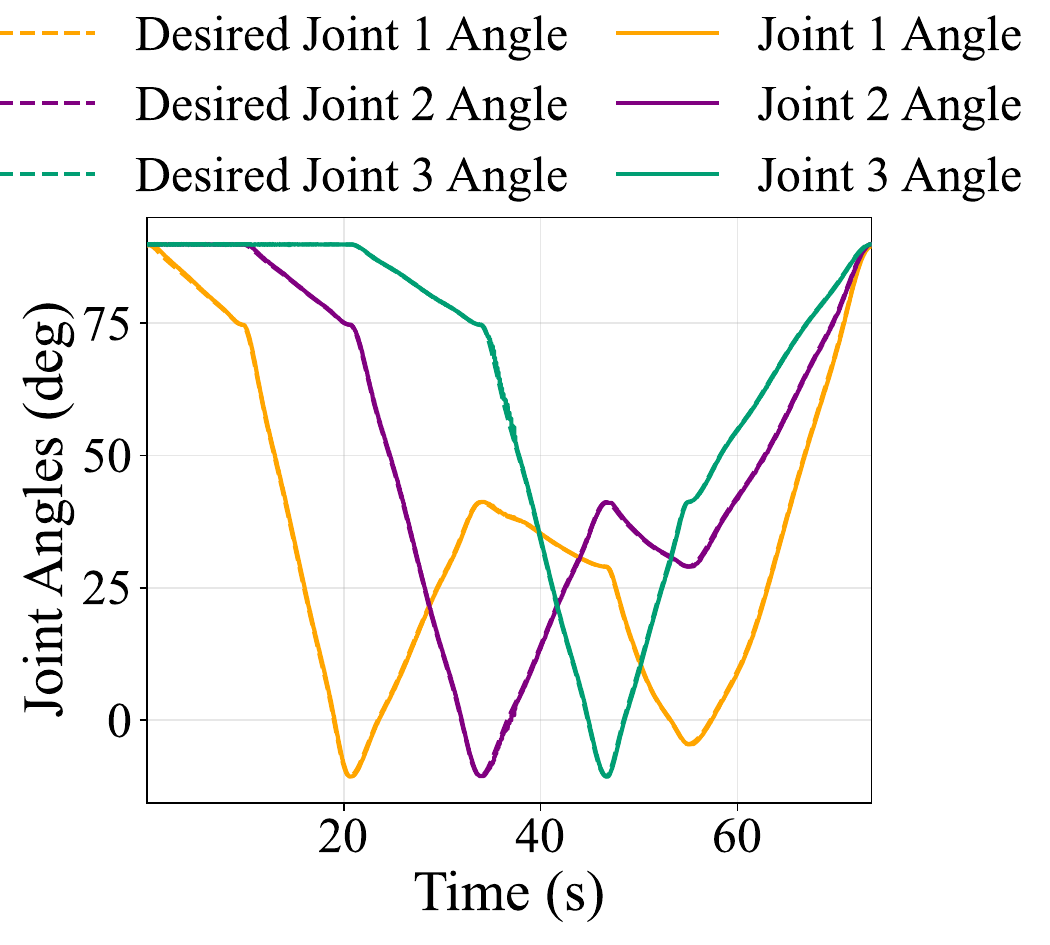}
        \label{fig: gap_joints}}
    \hfil
    \subfloat[Joint angles in Scene (b)]{
        \includegraphics[width=0.23\linewidth]{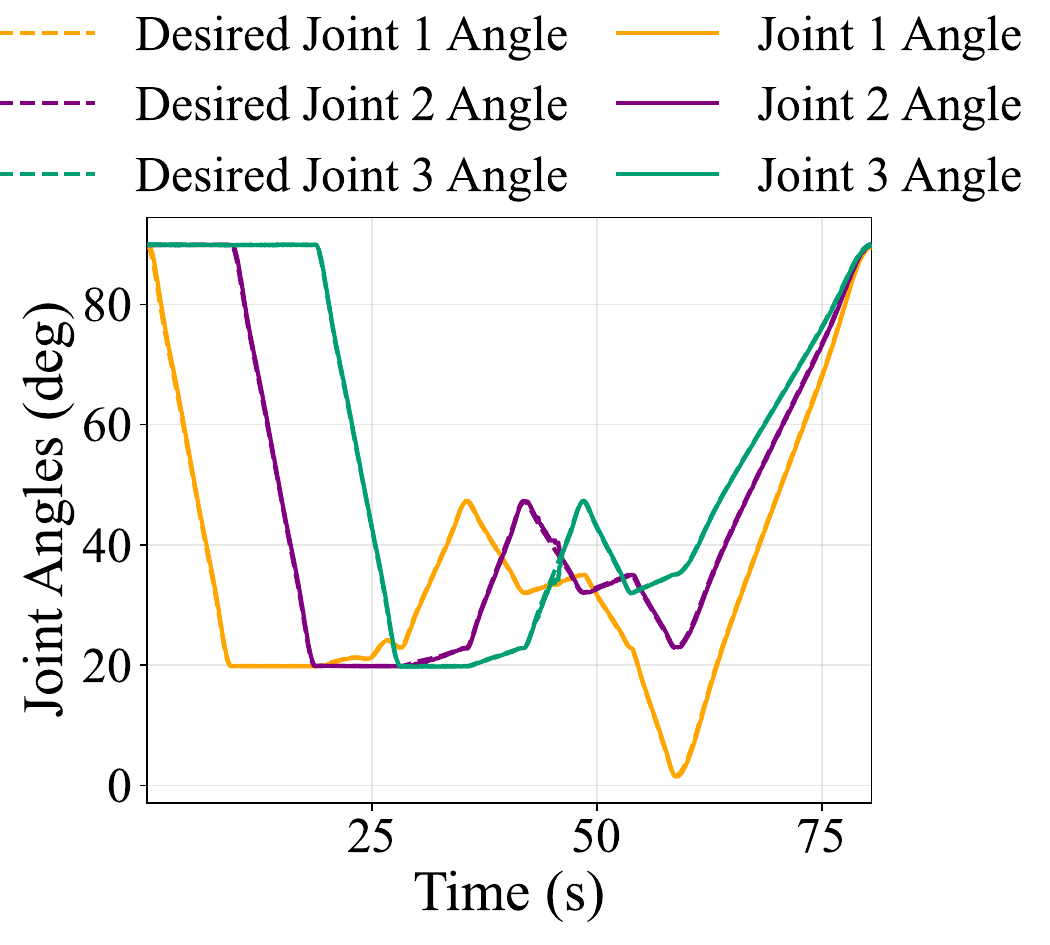}
        \label{fig: mix_gap_joints}}
    \hfil
    \subfloat[Joint angles in Scene (c)]{
        \includegraphics[width=0.23\linewidth]{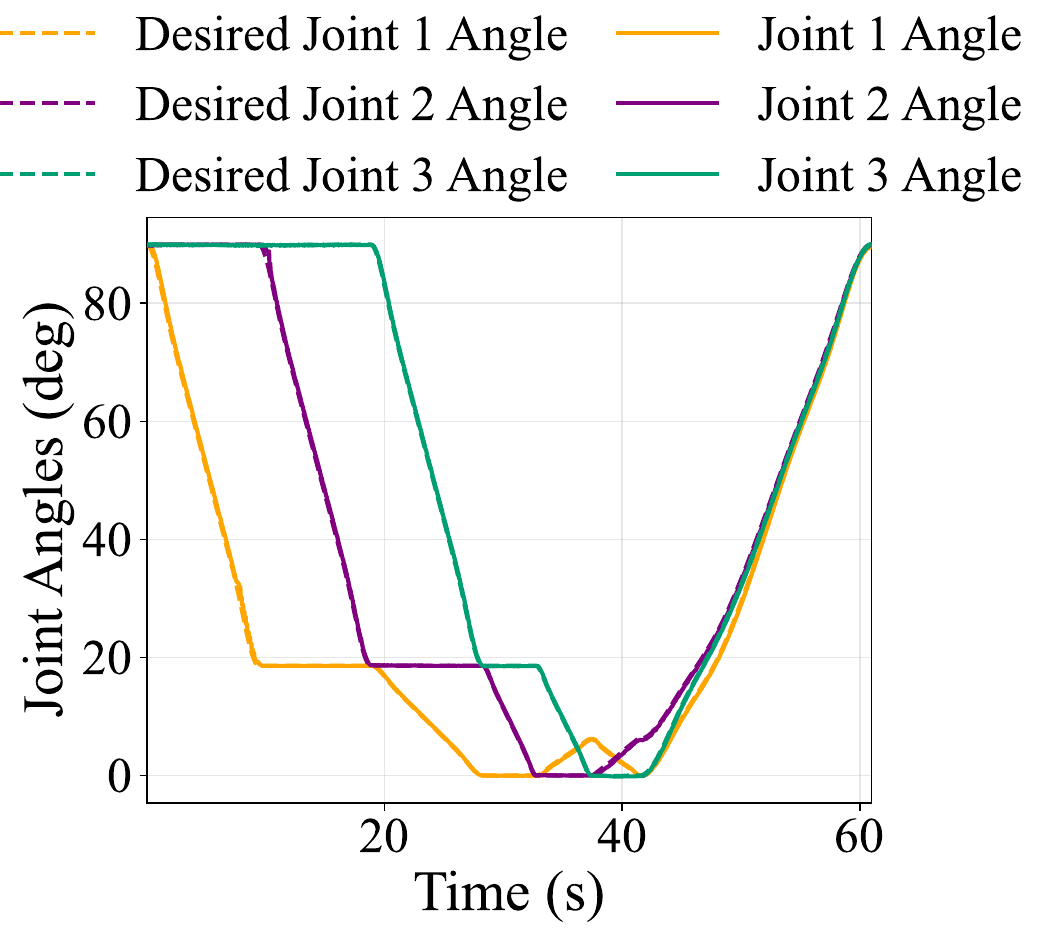}
        \label{fig: poles_joints}}
    \hfil
    \subfloat[Joint angles in Scene (d)]{
        \includegraphics[width=0.23\linewidth]{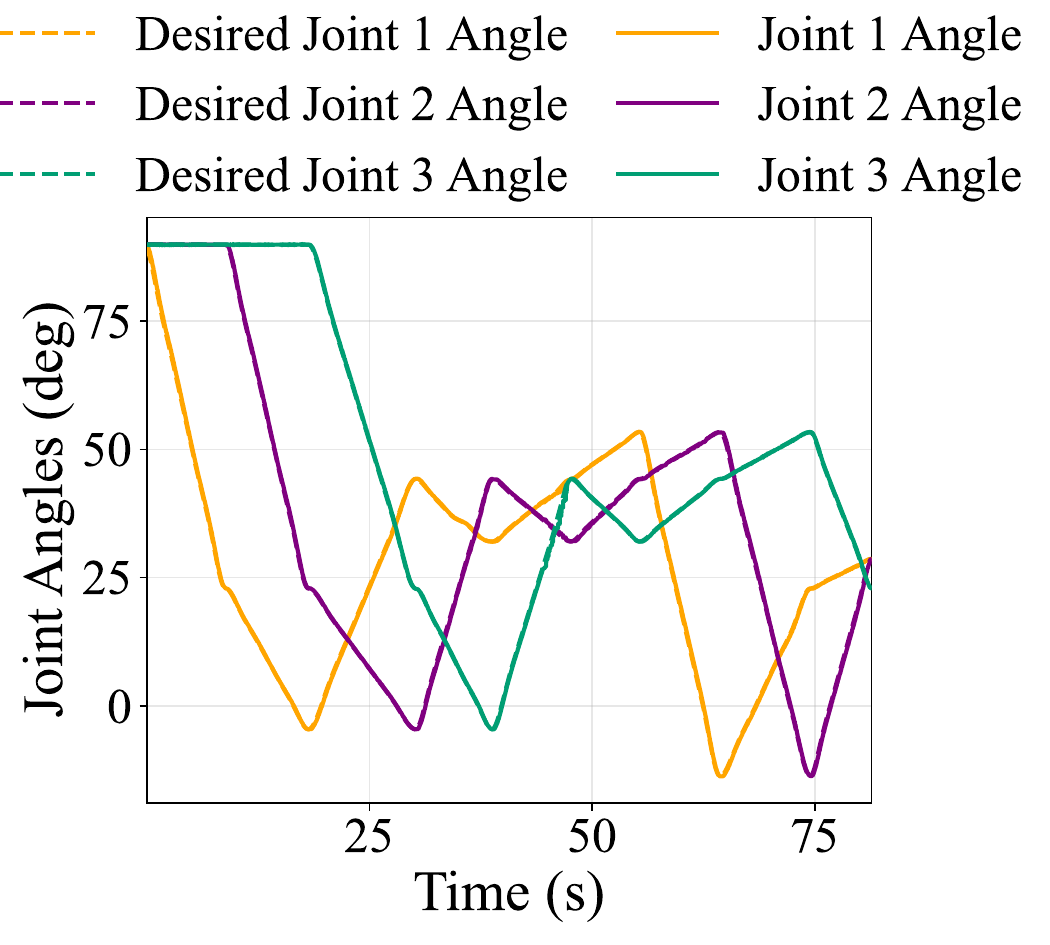}
        \label{fig: turn_joints}}

    \caption{Experimental demonstration of the floating-base multi-link robot performing complex maneuvers in confined environments. The first four rows show four representative scenarios: (a) single gap, (b) triple gaps, (c) pole obstacles, and (d) U-shaped passage. Columns 1–5 depict sequential snapshots where the robot adapted its articulated configuration to maneuver through narrow gaps and obstacles, with white arrows in the first column indicating motion direction. Subfigures (e)–(l) present the corresponding CoG and joint angle trajectories. Together, these results demonstrate that the hierarchical trajectory planning framework generated smooth, feasible motions and effectively exploited the robot's multi-link morphology to achieve maneuvering infeasible for conventional rigid robots.}
    \label{fig: exp_collection}
    \vspace{-10pt}
\end{figure*}

As shown in Fig. \ref{fig: exp_collection}, we tested the system in four environments: (a) a single gap, requiring passage through a narrow opening, (b) triple gaps, demanding consistent maneuvering across consecutive passages where the robot may span multiple gaps simultaneously, (c) pole obstacles, fragmenting the free space by multiple inadmissible voids, and (d) a U-shaped passage, forming a tunnel rather than restricting motion at specific points. Scenarios (b)--(d) presented structural challenges that are generally beyond the capability of existing planning algorithms for such robots. These environments represented typical confined structures and provided a progressive evaluation of the planner in realistic settings. Since our focus was motion planning rather than perception, the planner received environment information from ground-truth point clouds.

In all experiments, we set $\alpha_v = 0.15$ and $f_{\text{tol}} = 10^{-5}$. The translational velocity limit of the root link was set to $v_{\max}=\SI{0.5}{\meter\per\second}$ along each axis, and the rotational velocity limit was set to $\omega_{\max}=\SI{0.25}{\radian\per\second}$. The same $\omega_{\max}$ was applied to both the root link orientation rate and the joint angular velocities. All other parameters were consistent with Table~\ref{tab: params}.

\subsection{Results and Discussions}
The experiments in Fig.~\ref{fig: exp_collection} demonstrated that the proposed planner enabled the robot to accomplish challenging maneuvers in four representative environments. In each run, the optimization terminated after reaching the preset maximum computation time of \SI{10}{\second}. In all cases, the robot successfully reconfigured its articulated body to pass through narrow openings---as small as \SI{0.7}{\meter} wide---or maneuvered through the confined tunnel, demonstrating behaviors unattainable by rigid aerial robots.

The motion profiles from Fig.~\ref{fig: gap_cog} to Fig.~\ref{fig: turn_joints} further illustrated how the planner generated smooth and coordinated trajectories. The CoG paths followed continuous curves without abrupt changes, and the joint angles evolved in structured patterns tailored to each scenario. These results indicated that the framework adapted to different spatial constraints while systematically exploiting the robot's redundant morphology to generate feasible trajectories across diverse scenarios.

To substantiate feasibility with respect to the imposed velocity and controllability requirements, Table~\ref{tab: max_velocities} reports peak trajectory velocities (with fractions of $v_{\max}$ and $\omega_{\max}$) and statistics of the controllability metric $\tau_{\min}$ in \eqref{eq: control_check}.

\textbf{Velocity Constraint Compliance and Margins:}
Across all trials, the maximum translational velocity along any axis was \SI{0.149}{\meter\per\second} (30\% of $v_{\max}$), leaving a 70\% margin. The maximum root-link yaw rate was \SI{0.202}{\radian\per\second} (81\% of $\omega_{\max}$), which was the closest case to saturation but still retained a 19\% margin. The maximum joint angular velocity was \SI{0.171}{\radian\per\second} (68\% of $\omega_{\max}$), leaving a 32\% margin. Overall, all trajectories stayed strictly within the prescribed bounds, and no saturation of the enforced limits was observed.

\textbf{Sustained Controllability:}
We evaluated controllability using $\tau_{\min}$ defined in \eqref{eq: control_check} across all trials. This metric quantifies the configuration-dependent control-torque margin about the CoG: When $\tau_{\min} \leq \delta_{\tau}$, the platform loses attitude controllability, which inevitably results in a rapid crash. Table~\ref{tab: max_velocities} reports the minimum, average, and maximum $\tau_{\min}$ over each trial. In all four trials, the minimum observed $\tau_{\min}$ remained safely above $\delta_{\tau}$. Consistent with this, we did not observe any uncontrollable attitude divergence or drop-to-ground events during the experiments, indicating sustained control authority throughout the maneuvers.

Taken together, the experiments validated that the proposed hierarchical planner translated effectively from simulation to real hardware. The framework consistently generated smooth, feasible motions that enabled the robot to leverage its multi-link morphology for maneuvering that rigid aerial robots cannot achieve, supporting safe and reliable deployment of floating-base multi-link robots in confined environments.

\begin{table}[t]
\centering
\caption{Peak trajectory velocities and the controllability metric in four experimental trials.}
\renewcommand{\arraystretch}{1.15}
\label{tab: max_velocities}
\begin{tabular}{cccc}
\toprule
\multicolumn{4}{c}{Maximum root-link velocities on trajectory} \\
\cmidrule(lr){1-4}
Trial & X-Trans. (\si{\meter\per\second}) & Y-Trans. (\si{\meter\per\second}) & Yaw-Rot. (\si{\radian\per\second}) \\
\midrule
a & \SI{0.077}{} (15\%)  & \SI{0.100}{} (20\%) & \SI{0.169}{} (68\%) \\
b & \SI{0.146}{} (29\%)  & \SI{0.149}{} (30\%) & \SI{0.202}{} (81\%) \\
c & \SI{0.149}{} (30\%)  & \SI{0.133}{} (27\%) & \SI{0.139}{} (56\%) \\
d & \SI{0.0975}{} (20\%) & \SI{0.114}{} (23\%) & \SI{0.155}{} (62\%) \\
\midrule
\multicolumn{4}{c}{Maximum joint angular velocities on trajectory (\si{\radian\per\second})} \\
\cmidrule(lr){1-4}
Trial & Joint 1 & Joint 2 & Joint 3 \\
\midrule
a & \SI{0.171}{} (68\%) & \SI{0.140}{} (56\%) & \SI{0.145}{} (58\%) \\
b & \SI{0.164}{} (66\%) & \SI{0.164}{} (66\%) & \SI{0.164}{} (66\%) \\
c & \SI{0.169}{} (68\%) & \SI{0.169}{} (68\%) & \SI{0.169}{} (68\%) \\
d & \SI{0.166}{} (66\%) & \SI{0.157}{} (63\%) & \SI{0.124}{} (50\%) \\
\midrule
\multicolumn{4}{c}{Controllability metric $\tau_{\min}$ in \eqref{eq: control_check} during the flight (\si{\newton\meter})}  \\
\cmidrule(lr){1-4}
Trial & Minimum & Average & Maximum \\
\midrule
a & \SI{0.857}{} & \SI{0.940}{} & \SI{1.10}{} \\
b & \SI{0.858}{} & \SI{0.920}{} & \SI{1.12}{} \\
c & \SI{0.390}{} & \SI{0.887}{} & \SI{1.13}{} \\
d & \SI{0.856}{} & \SI{0.898}{} & \SI{1.11}{} \\
\bottomrule
\end{tabular}
\\[1.0ex] 
\footnotesize{Percentages indicate the fraction of the imposed limits, with $v_{\max}=\SI{0.5}{\meter\per\second}$ for each translational axis and $\omega_{\max}=\SI{0.25}{\radian\per\second}$ for rotation.}
\vspace{-10pt}
\end{table}

\section{Conclusions}
\label{Conclusions}

This work presented a trajectory-based planning framework for floating-base multi-link robots operating in confined environments. By combining global decomposition with tailored trajectory parameterization and optimization, the method achieved parallel computation while producing naturally continuous trajectories. The framework was validated through both large-scale simulations and real-world experiments, demonstrating how it systematically exploits the robot's articulated morphology to accomplish maneuvering infeasible for rigid aerial robots.

The results confirmed that each component of the framework---global anchor states, local trajectory planning, and parallel computation---played a vital role in ensuring both feasibility and efficiency. Real-world experiments further established the practicality of the approach, showing that it can process point-cloud data to generate feasible trajectories without relying on hand-crafted semantic obstacle information.

\textbf{Limitations and Future Directions:} First, as an initial step toward general motion planning for floating-base multi-link robots in confined environments, the evaluation relied on motion-capture feedback, and incorporating onboard perception and localization is a promising next step toward broader autonomy. Second, the current algorithm is not specifically designed for dynamic environments, and extending it to scenarios with dynamically moving obstacles would expand its applicability and robustness. Third, due to site constraints, experiments were limited to short-horizon tasks, and scaling to longer missions and more complex conditions remains an encouraging direction, including contact-rich maneuvering and interactions with deformable obstacles. Fourth, while this study was validated on an aerial robot, adapting the same core framework to other domains, such as underwater and terrestrial robots, could further demonstrate its generality.

Beyond these extensions, an important next step is to deploy the system in real-world applications. Floating-base multi-link robots are particularly promising for industrial inspection in confined mechanical structures (e.g., boilers, ducts, pipe networks), infrastructure maintenance in cluttered spaces such as bridges or culverts, and search-and-rescue operations where maneuvering through collapsed or partially obstructed environments is critical. We aim to deploy the framework in such field-relevant scenarios to assess performance under realistic situations. Advancing these directions may strengthen the framework's robustness and scalability and broaden its applicability to real-world domains.





\appendices

\section*{Appendix A \\ Calculation of the energy cost of b-spline}
\refstepcounter{section}
\label{appendix_a}
From the knot vector defined in \eqref{eq: knot}, we denote the interval $\left[u_i, u_{i+1}\right)$ for $i=0,1,\dots, N+p+3$ as the $i$-th knot span. This span may have zero length if consecutive knots coincide. A basis function $B_{j,p}(t)$, where $j=0,1,\dots, N+3$, is nonzero only within the $j$-th to $(j+p)$-th knot spans. For \eqref{eq: energy cost}, the matrix $\boldsymbol{M}$ is a symmetric band matrix, with entries given by
\begin{equation}
M_{ij} =
\begin{cases}
\displaystyle
\frac{1}{h}\sum_{u=j}^{i+p} \int_{0}^{1}\dot{B}_{i,p}^u(t) \dot{B}_{j,p}^u(t) \text{dt},
& \text{if }i\le j \le i+p,\\[1ex]
M_{ji}, &\text{if }j\le i \le j+p, \\ 
0, & \text{else},
\end{cases}
\end{equation}
where $\dot{B}_{i,p}^u(t)$ denotes the derivative of the $i$-th basis function on the $u$-th knot span.

\section*{Appendix B \\ Calculation of the gradient of the face-normal torque w.r.t. configuration}
\refstepcounter{section}
\label{appendix_b}
We here present explicit formulas for the gradient $\nabla_{\boldsymbol{q}}\tau_{ijk} \in \mathbb{R}^D$ used in \eqref{eq: grad_controllability2q}. Let $\boldsymbol{\tau}_{ij} = \boldsymbol{\tau}_{i} \times \boldsymbol{\tau}_{j} \in \mathbb{R}^3$. Then, \eqref{eq: fc_t_dis} can be rewritten as 
\begin{equation}
    \tau_{ijk} = \frac{ \boldsymbol{\tau}_{ij}^{\top} \boldsymbol{\tau}_k}{ \left\| \boldsymbol{\tau}_{ij} \right\| },
\end{equation}
its gradient w.r.t. the robot's configuration $\nabla_{\boldsymbol{q}}\tau_{ijk}$ can be derived using the chain rule:
\begin{equation}
    \nabla_{\boldsymbol{q}}\tau_{ijk} = \frac{\nabla_{\boldsymbol{q}} \left(\boldsymbol{\tau}_{ij}^{\top} \boldsymbol{\tau}_k \right) \left\| \boldsymbol{\tau}_{ij} \right\| - \boldsymbol{\tau}_{ij}^{\top} \boldsymbol{\tau}_k \nabla_{\boldsymbol{q}}\left\| \boldsymbol{\tau}_{ij} \right\|}{\left\| \boldsymbol{\tau}_{ij} \right\|^2},
\end{equation}
where
\begin{subequations}
\begin{align}
    \nabla_{\boldsymbol{q}} \left(\boldsymbol{\tau}_{ij}^{\top} \boldsymbol{\tau}_k \right) &= \left(\nabla_{\boldsymbol{q}}\boldsymbol{\tau}_{ij} \right)^\top \boldsymbol{\tau}_{k} + \left(\nabla_{\boldsymbol{q}}\boldsymbol{\tau}_{k}  \right)^\top \boldsymbol{\tau}_{ij} \in \mathbb{R}^{D}, \\
    \nabla_{\boldsymbol{q}}\left\| \boldsymbol{\tau}_{ij} \right\| &= \frac{\left(\nabla_{\boldsymbol{q}}\boldsymbol{\tau}_{ij}  \right)^\top \boldsymbol{\tau}_{ij}}{\left\| \boldsymbol{\tau}_{ij} \right\|} \in \mathbb{R}^D, \\
    \nabla_{\boldsymbol{q}}\boldsymbol{\tau}_{ij} &= - [\boldsymbol{\tau}_{j}]_\times \nabla_{\boldsymbol{q}}\boldsymbol{\tau}_{i} +  [\boldsymbol{\tau}_{i}]_\times \nabla_{\boldsymbol{q}}\boldsymbol{\tau}_{j} \in \mathbb{R}^{3\times D},
\end{align}
\end{subequations}
with $\nabla_{\boldsymbol{q}}\boldsymbol{\tau}_{i}$ given in \eqref{eq: grad_tau2q}.

\section*{Appendix C \\ Calculation of Jacobians in Arbitrary Frames}
\refstepcounter{section}
\label{appendix_c}
For a more general derivation of $\prescript{C}{}{\boldsymbol{J}}_{i}^{\text{rot}}$ and $\prescript{C}{}{\boldsymbol{J}}_{i}^{\boldsymbol{e}}$ in \eqref{eq: grad_tau2q}, we introduce a method to compute the Jacobian $\prescript{T}{}{\boldsymbol{J}}^{\boldsymbol{p}}$ of a geometric vector $\boldsymbol{p}$, expressed in an arbitrary operational-space frame ${T}$, with respect to the robot configuration $\boldsymbol{q}$ (the dependence on $\boldsymbol{q}$ is omitted below for brevity). The problem is defined as follows.

\textbf{Given: }
\begin{itemize}
\item The coordiate $\prescript{S}{}{\boldsymbol{p}} \in \mathbb{R}^3$ of a vector in a known source frame $\{S\}$.
\item The Jacobian $\prescript{S}{}{\boldsymbol{J}}^{\boldsymbol{p}} \in \mathbb{R}^{3\times D}$ of the vector expressed in the source frame $\{S\}$.
\item The positions $\prescript{W}{}{\boldsymbol{o}_S}, \prescript{W}{}{\boldsymbol{o}_T} \in \mathbb{R}^3$ of the source frame and target frame's origins in the world frame.
\item The Jacobians $\prescript{W}{}{\boldsymbol{J}}^{\boldsymbol{o}_S},\prescript{W}{}{\boldsymbol{J}}^{\boldsymbol{o}_T} \in \mathbb{R}^{6\times D}$ of the source and target frames (in $ \mathrm{SE}(3)$) expressed in the world frame $\{W\}$.
\item The rotation matrix $\prescript{W}{S}{\boldsymbol{R}}$ from the source frame $\{S\}$ to the world frame $\{W\}$ and the rotation matrix $\prescript{T}{W}{\boldsymbol{R}}$ from the world frame $\{W\}$ to the target frame $\{T\}$.
\end{itemize}

\textbf{To obtain: }
\begin{itemize}
\item The Jacobian $\prescript{T}{}{\boldsymbol{J}}^{\boldsymbol{p}}  \in \mathbb{R}^{3\times D}$ of the vector expressed in the target frame $\{T\}$. 
\end{itemize}

All input variables can be obtained from modern robotics libraries such as \textit{Pinocchio}. Using $\{W\}$ as an intermediate reference, the vector coordinates in $\{W\}$ and $\{T\}$ are given by
\begin{equation}
    \prescript{W}{}{\boldsymbol{p}} = \prescript{W}{S}{\boldsymbol{R}} \prescript{S}{}{\boldsymbol{p}} + \prescript{W}{}{\boldsymbol{o}_S},
\end{equation}
\begin{equation}
    \prescript{T}{}{\boldsymbol{p}} = \prescript{T}{W}{\boldsymbol{R}}
    \left(\prescript{W}{}{\boldsymbol{p}} - \prescript{W}{}{\boldsymbol{o}_T} \right).
\end{equation}
Thus, the Jacobian in $\{W\}$ is first obtained as
\begin{equation}
\label{eq: jac_world}
    \prescript{W}{}{\boldsymbol{J}}^{\boldsymbol{p}} = \prescript{W}{S}{\boldsymbol{R}} \prescript{S}{}{\boldsymbol{J}}^{\boldsymbol{p}} + \prescript{W}{}{\boldsymbol{J}}^{\boldsymbol{o}_S}_v - \left[ \prescript{W}{S}{\boldsymbol{R}} \prescript{S}{}{\boldsymbol{p}} \right]_\times \prescript{W}{}{\boldsymbol{J}}^{\boldsymbol{o}_S}_{\omega},
\end{equation}
and then transformed into the target frame $\{T\}$ as
\begin{equation}
\label{eq: jac_target}
    \prescript{T}{}{\boldsymbol{J}}^{\boldsymbol{p}} = \prescript{T}{W}{\boldsymbol{R}} \left(\prescript{W}{}{\boldsymbol{J}}^{\boldsymbol{p}} -  \prescript{W}{}{\boldsymbol{J}}^{\boldsymbol{o}_T}_v\right) + \left[\prescript{T}{}{\boldsymbol{p}} \right]_\times \prescript{T}{W}{\boldsymbol{R}} \prescript{W}{}{\boldsymbol{J}}^{\boldsymbol{o}_T}_{\omega}.
\end{equation}
Here, the subscripts $v$ and $\omega$ denote the translational and rotational components of the Jacobians, respectively.


\bibliographystyle{IEEEtran_ref_v1.14}

\bibliography{references}

@article{karaman_sampling-based_2011,
	title = {Sampling-{Based} {Algorithms} for {Optimal} {Motion} {Planning}},
	volume = {30},
	issn = {0278-3649, 1741-3176},
	url = {http://journals.sagepub.com/doi/10.1177/0278364911406761},
	doi = {10.1177/0278364911406761},
	language = {en},
	number = {7},
	urldate = {2023-05-10},
	journal = {The International Journal of Robotics Research},
	author = {Karaman, Sertac and Frazzoli, Emilio},
	month = jun,
	year = {2011},
	pages = {846--894},
}

@article{kavraki_probabilistic_1996,
	title = {Probabilistic {Roadmaps} for {Path} {Planning} in {High}-{Dimensional} {Configuration} {Spaces}},
	volume = {12},
	copyright = {https://ieeexplore.ieee.org/Xplorehelp/downloads/license-information/IEEE.html},
	issn = {1042296X},
	url = {http://ieeexplore.ieee.org/document/508439/},
	doi = {10.1109/70.508439},
	language = {en},
	number = {4},
	urldate = {2024-05-06},
	journal = {IEEE Transactions on Robotics and Automation},
	author = {Kavraki, L.E. and Svestka, P. and Latombe, J.-C. and Overmars, M.H.},
	month = aug,
	year = {1996},
	pages = {566--580},
}

@article{xu_fast-lio2_2022,
	title = {{FAST}-{LIO2}: {Fast} {Direct} {LiDAR}-{Inertial} {Odometry}},
	volume = {38},
	issn = {1941-0468},
	shorttitle = {{FAST}-{LIO2}},
	url = {https://ieeexplore.ieee.org/document/9697912/},
	doi = {10.1109/TRO.2022.3141876},
	number = {4},
	urldate = {2025-12-31},
	journal = {IEEE Transactions on Robotics},
	author = {Xu, Wei and Cai, Yixi and He, Dongjiao and Lin, Jiarong and Zhang, Fu},
	month = aug,
	year = {2022},
	pages = {2053--2073},
}

@article{richter_polynomial_2016,
	title = {Polynomial {Trajectory} {Planning} for {Aggressive} {Quadrotor} {Flight} in {Dense} {Indoor} {Environments}},
	url = {https://doi.org/10.1007/978-3-319-28872-7_37},
	language = {en-US},
	journal = {Robotics Research: The 16th International Symposium ISRR},
	author = {Richter, Charles and Bry, Adam and Roy, Nicholas},
	editor = {Inaba, Masayuki and Corke, Peter},
	year = {2016},
	doi = {10.1007/978-3-319-28872-7_37},
	pages = {649--666},
}

@article{jennings_computational_1990,
	title = {A {Computational} {Algorithm} for {Functional} {Inequality} {Constrained} {Optimization} {Problems}},
	volume = {26},
	issn = {0005-1098},
	url = {https://www.sciencedirect.com/science/article/pii/000510989090131Z},
	doi = {10.1016/0005-1098(90)90131-Z},
	number = {2},
	journal = {Automatica},
	author = {Jennings, L.S. and Teo, K.L.},
	month = mar,
	year = {1990},
	pages = {371--375},
}

@article{liu_planning_2017,
	title = {Planning {Dynamically} {Feasible} {Trajectories} for {Quadrotors} {Using} {Safe} {Flight} {Corridors} in 3-{D} {Complex} {Environments}},
	volume = {2},
	issn = {2377-3766},
	url = {https://ieeexplore.ieee.org/document/7839930/},
	doi = {10.1109/LRA.2017.2663526},
	number = {3},
	urldate = {2025-09-27},
	journal = {IEEE Robotics and Automation Letters},
	author = {Liu, Sikang and Watterson, Michael and Mohta, Kartik and Sun, Ke and Bhattacharya, Subhrajit and Taylor, Camillo J. and Kumar, Vijay},
	month = jul,
	year = {2017},
	pages = {1688--1695},
}

@article{pi_omepp_2024,
	title = {{OMEPP}: {Online} {Multi}-{Population} {Evolutionary} {Path} {Planning} for {Mobile} {Manipulators} in {Dynamic} {Environments}},
	volume = {22},
	issn = {1558-3783},
	shorttitle = {{OMEPP}},
	url = {https://ieeexplore.ieee.org/document/10636084/},
	doi = {10.1109/TASE.2024.3440252},
	language = {en-US},
	urldate = {2025-09-13},
	journal = {IEEE Transactions on Automation Science and Engineering},
	author = {Pi, Yangjun and Liu, Xin and Yang, Zuodong and Zhong, Yunlin and Huang, Tao and Pu, Huayan and Luo, Jun},
	month = aug,
	year = {2024},
	pages = {6234--6245},
}

@article{sun_orthogonal_2024,
	title = {An {Orthogonal} {Repetitive} {Motion} and {Obstacle} {Avoidance} {Scheme} for {Omnidirectional} {Mobile} {Robotic} {Arm}},
	volume = {72},
	issn = {1557-9948},
	url = {https://ieeexplore.ieee.org/document/10735396/},
	doi = {10.1109/TIE.2024.3451063},
	language = {en-US},
	number = {5},
	urldate = {2025-09-13},
	journal = {IEEE Transactions on Industrial Electronics},
	author = {Sun, Zhongbo and Tang, Shijun and Fei, Yuzhe and Xiao, Xingtian and Hu, Yunfeng and Yu, Junzhi},
	month = oct,
	year = {2024},
	pages = {4978--4989},
}

@article{pettinger_efficient_2024,
	title = {Efficient {Constrained} {Motion} {Planning} {Using} {Direct} {Sampling} of {Screw}-{Constraint} {Manifolds}},
	volume = {22},
	issn = {1558-3783},
	url = {https://ieeexplore.ieee.org/document/10794549/},
	doi = {10.1109/TASE.2024.3513160},
	language = {en-US},
	urldate = {2025-09-13},
	journal = {IEEE Transactions on Automation Science and Engineering},
	author = {Pettinger, Adam and Panthi, Janak and Alambeigi, Farshid and Pryor, Mitch},
	month = dec,
	year = {2024},
	pages = {9793--9809},
}

@article{covic_real-time_2025,
	title = {Real-{Time} {Sampling}-{Based} {Safe} {Motion} {Planning} for {Robotic} {Manipulators} in {Dynamic} {Environments}},
	volume = {41},
	issn = {1941-0468},
	url = {https://ieeexplore.ieee.org/document/11122893/},
	doi = {10.1109/TRO.2025.3598119},
	urldate = {2025-09-13},
	journal = {IEEE Transactions on Robotics},
	author = {Covic, Nermin and Lacevic, Bakir and Osmankovic, Dinko and Uzunovic, Tarik},
	month = aug,
	year = {2025},
	pages = {5287--5306},
}

@article{wensing_optimization-based_2023,
	title = {Optimization-{Based} {Control} for {Dynamic} {Legged} {Robots}},
	volume = {40},
	issn = {1941-0468},
	url = {https://ieeexplore.ieee.org/document/10286076/},
	doi = {10.1109/TRO.2023.3324580},
	urldate = {2025-09-17},
	journal = {IEEE Transactions on Robotics},
	author = {Wensing, Patrick M. and Posa, Michael and Hu, Yue and Escande, Adrien and Mansard, Nicolas and Prete, Andrea Del},
	month = oct,
	year = {2023},
	pages = {43--63},
}

@article{zhao_versatile_2021,
	title = {Versatile {Multilinked} {Aerial} {Robot} {With} {Tilted} {Propellers}: {Design}, {Modeling}, {Control}, and {State} {Estimation} for {Autonomous} {Flight} and {Manipulation}},
	volume = {38},
	issn = {1556-4967},
	shorttitle = {Versatile multilinked aerial robot with tilted propellers},
	url = {https://onlinelibrary.wiley.com/doi/abs/10.1002/rob.22019},
	doi = {10.1002/rob.22019},
	language = {en},
	number = {7},
	urldate = {2024-11-27},
	journal = {Journal of Field Robotics},
	author = {Zhao, Moju and Anzai, Tomoki and Shi, Fan and Maki, Toshiya and Nishio, Takuzumi and Ito, Keita and Kuromiya, Naoki and Okada, Kei and Inaba, Masayuki},
	month = may,
	year = {2021},
	note = {\_eprint: https://onlinelibrary.wiley.com/doi/pdf/10.1002/rob.22019},
	pages = {933--966},
}

@article{marshall_survey_2021,
	title = {A {Survey} of {Guidance}, {Navigation}, and {Control} {Systems} for {Autonomous} {Multi}-{Rotor} {Small} {Unmanned} {Aerial} {Systems}},
	volume = {52},
	issn = {13675788},
	url = {https://linkinghub.elsevier.com/retrieve/pii/S1367578821000882},
	doi = {10.1016/j.arcontrol.2021.10.013},
	language = {en},
	urldate = {2023-05-10},
	journal = {Annual Reviews in Control},
	author = {Marshall, Julius A. and Sun, Wei and L’Afflitto, Andrea},
	month = dec,
	year = {2021},
	pages = {390--427},
}

@article{lee_introspective_2024,
	title = {Introspective {Perception} for {Long}-{Term} {Aerial} {Telemanipulation} {With} {Virtual} {Reality}},
	volume = {1},
	copyright = {https://ieeexplore.ieee.org/Xplorehelp/downloads/license-information/IEEE.html},
	issn = {2997-1101},
	url = {https://ieeexplore.ieee.org/document/10755951/},
	doi = {10.1109/TFR.2024.3494719},
	language = {en},
	urldate = {2025-09-16},
	journal = {IEEE Transactions on Field Robotics},
	author = {Lee, Jongseok and Balachandran, Ribin and Kondak, Konstantin and Coelho, Andre and Stefano, Marco De and Humt, Matthias and Feng, Jianxiang and Asfour, Tamim and Triebel, Rudolph},
	month = nov,
	year = {2024},
	pages = {360--393},
}

@article{wang_unlocking_2025,
	title = {Unlocking {Aerobatic} {Potential} of {Quadcopters}: {Autonomous} {Freestyle} {Flight} {Generation} and {Execution}},
	volume = {10},
	issn = {2470-9476},
	shorttitle = {Unlocking aerobatic potential of quadcopters},
	url = {https://www.science.org/doi/10.1126/scirobotics.adp9905},
	doi = {10.1126/scirobotics.adp9905},
	language = {en},
	number = {101},
	urldate = {2025-09-22},
	journal = {Science Robotics},
	author = {Wang, Mingyang and Wang, Qianhao and Wang, Ze and Gao, Yuman and Wang, Jingping and Cui, Can and Li, Yuan and Ding, Ziming and Wang, Kaiwei and Xu, Chao and Gao, Fei},
	month = apr,
	year = {2025},
	pages = {eadp9905},
}

@article{xing_morphing_2024,
	title = {Morphing {Quadrotors}: {Enhancing} {Versatility} and {Adaptability} in {Drone} {Applications}—{A} {Review}},
	volume = {8},
	copyright = {https://creativecommons.org/licenses/by/4.0/},
	issn = {2504-446X},
	shorttitle = {Morphing {Quadrotors}},
	url = {https://www.mdpi.com/2504-446X/8/12/762},
	doi = {10.3390/drones8120762},
	language = {en},
	number = {12},
	urldate = {2025-09-16},
	journal = {Drones},
	author = {Xing, Siyuan and Zhang, Xuhui and Tian, Jiandong and Xie, Chunlei and Chen, Zhihong and Sun, Jianwei},
	month = dec,
	year = {2024},
	pages = {1--20},
}

@article{jin_whole-body_2024,
	title = {Whole-{Body} {Inverse} {Kinematics} and {Operation}-{Oriented} {Motion} {Planning} for {Robot} {Mobile} {Manipulation}},
	volume = {20},
	issn = {1941-0050},
	url = {https://ieeexplore.ieee.org/document/10644079/},
	doi = {10.1109/TII.2024.3441661},
	number = {12},
	urldate = {2025-09-13},
	journal = {IEEE Transactions on Industrial Informatics},
	author = {Jin, Tianlei and Zhu, Hongwei and Zhu, Jiakai and Zhu, Shiqiang and He, Zaixing and Zhang, Shuyou and Song, Wei and Gu, Jason},
	month = dec,
	year = {2024},
	pages = {14239--14248},
}

@inproceedings{hameed_dragonfly_2025,
	address = {Atlanta, GA, USA},
	title = {Dragonfly {Drone}: {A} {Novel} {Tilt}-{Rotor} {Aerial} {Platform} with {Body}-{Morphing} {Capability}},
	shorttitle = {Dragonfly {Drone}},
	url = {https://ieeexplore.ieee.org/document/11127712/},
	doi = {10.1109/ICRA55743.2025.11127712},
	urldate = {2025-09-16},
	booktitle = {2025 {IEEE} {International} {Conference} on {Robotics} and {Automation} ({ICRA})},
	author = {Hameed, Syed Waqar and Jie, Alex Liew Jun and Imanberdiyev, Nursultan and Camci, Efe and Yau, Wei-Yun and Feroskhan, Mir},
	month = may,
	year = {2025},
	pages = {8642--8648},
}

@inproceedings{thakar_accelerating_2020,
	address = {Las Vegas, NV, USA},
	title = {Accelerating {Bi}-{Directional} {Sampling}-{Based} {Search} for {Motion} {Planning} of {Non}-{Holonomic} {Mobile} {Manipulators}},
	url = {https://ieeexplore.ieee.org/document/9340782/},
	doi = {10.1109/IROS45743.2020.9340782},
	urldate = {2025-09-06},
	booktitle = {2020 {IEEE}/{RSJ} {International} {Conference} on {Intelligent} {Robots} and {Systems} ({IROS})},
	author = {Thakar, Shantanu and Rajendran, Pradeep and Kim, Hyojeong and Kabir, Ariyan M. and Gupta, Satyandra K.},
	month = oct,
	year = {2020},
	note = {ISSN: 2153-0866},
	pages = {6711--6717},
}

@inproceedings{zhao_flight_2018,
	address = {Madrid, Spain},
	title = {Flight {Motion} of {Passing} {Through} {Small} {Opening} by {DRAGON}: {Transformable} {Multilinked} {Aerial} {Robot}},
	isbn = {978-1-5386-8094-0},
	shorttitle = {Flight {Motion} of {Passing} {Through} {Small} {Opening} by {DRAGON}},
	url = {https://ieeexplore.ieee.org/document/8593368/},
	doi = {10.1109/IROS.2018.8593368},
	language = {en},
	urldate = {2024-01-03},
	booktitle = {2018 {IEEE}/{RSJ} {International} {Conference} on {Intelligent} {Robots} and {Systems} ({IROS})},
	publisher = {IEEE},
	author = {Zhao, Moju and Shi, Fan and Anzai, Tomoki and Chaudhary, Krishneel and Chen, Xiangyu and Okada, Kei and Inaba, Masayuki},
	month = oct,
	year = {2018},
	pages = {4735--4742},
}

@article{marcucci_motion_2023,
	title = {Motion {Planning} {Around} {Obstacles} with {Convex} {Optimization}},
	volume = {8},
	issn = {2470-9476},
	url = {https://www.science.org/doi/10.1126/scirobotics.adf7843},
	doi = {10.1126/scirobotics.adf7843},
	language = {en},
	number = {84},
	urldate = {2024-12-02},
	journal = {Science Robotics},
	author = {Marcucci, Tobia and Petersen, Mark and Von Wrangel, David and Tedrake, Russ},
	month = nov,
	year = {2023},
	pages = {eadf7843},
}

@article{schulman_motion_2014,
	title = {Motion {Planning} with {Sequential} {Convex} {Optimization} and {Convex} {Collision} {Checking}},
	volume = {33},
	issn = {0278-3649, 1741-3176},
	url = {https://journals.sagepub.com/doi/10.1177/0278364914528132},
	doi = {10.1177/0278364914528132},
	language = {en},
	number = {9},
	urldate = {2024-11-01},
	journal = {The International Journal of Robotics Research},
	author = {Schulman, John and Duan, Yan and Ho, Jonathan and Lee, Alex and Awwal, Ibrahim and Bradlow, Henry and Pan, Jia and Patil, Sachin and Goldberg, Ken and Abbeel, Pieter},
	month = aug,
	year = {2014},
	pages = {1251--1270},
}

@article{zucker_chomp_2013,
	title = {{CHOMP}: {Covariant} {Hamiltonian} {Optimization} for {Motion} {Planning}},
	volume = {32},
	issn = {0278-3649},
	shorttitle = {{CHOMP}},
	url = {https://doi.org/10.1177/0278364913488805},
	doi = {10.1177/0278364913488805},
	language = {en},
	number = {9-10},
	urldate = {2024-10-28},
	journal = {The International Journal of Robotics Research},
	author = {Zucker, Matt and Ratliff, Nathan and Dragan, Anca D. and Pivtoraiko, Mihail and Klingensmith, Matthew and Dellin, Christopher M. and Bagnell, J. Andrew and Srinivasa, Siddhartha S.},
	month = aug,
	year = {2013},
	note = {Publisher: SAGE Publications Ltd STM},
	pages = {1164--1193},
}

@article{kuindersma_optimization-based_2016,
	title = {Optimization-{Based} {Locomotion} {Planning}, {Estimation}, and {Control} {Design} for the {Atlas} {Humanoid} {Robot}},
	volume = {40},
	issn = {0929-5593, 1573-7527},
	url = {http://link.springer.com/10.1007/s10514-015-9479-3},
	doi = {10.1007/s10514-015-9479-3},
	language = {en},
	number = {3},
	urldate = {2025-09-16},
	journal = {Autonomous Robots},
	author = {Kuindersma, Scott and Deits, Robin and Fallon, Maurice and Valenzuela, Andrés and Dai, Hongkai and Permenter, Frank and Koolen, Twan and Marion, Pat and Tedrake, Russ},
	month = mar,
	year = {2016},
	pages = {429--455},
}

@article{elango_continuous-time_2025,
	title = {Continuous-{Time} {Successive} {Convexification} for {Constrained} {Trajectory} {Optimization}},
	volume = {180},
	issn = {00051098},
	url = {https://linkinghub.elsevier.com/retrieve/pii/S0005109825003589},
	doi = {10.1016/j.automatica.2025.112464},
	language = {en},
	urldate = {2025-09-16},
	journal = {Automatica},
	author = {Elango, Purnanand and Luo, Dayou and Kamath, Abhinav G. and Uzun, Samet and Kim, Taewan and Açıkmeşe, Behçet},
	month = oct,
	year = {2025},
	pages = {112464},
}

@article{liu_coordinated_2024,
	title = {A {Coordinated} {Framework} of {Aerial} {Manipulator} for {Safe} and {Compliant} {Physical} {Interaction}},
	volume = {146},
	issn = {09670661},
	url = {https://linkinghub.elsevier.com/retrieve/pii/S0967066124000583},
	doi = {10.1016/j.conengprac.2024.105898},
	language = {en},
	urldate = {2025-09-16},
	journal = {Control Engineering Practice},
	author = {Liu, Qianyuan and Lyu, Shangke and Guo, Kexin and Wang, Jianliang and Yu, Xiang and Guo, Lei},
	month = may,
	year = {2024},
	pages = {105898},
}

@article{al_ali_path_2024,
	title = {Path {Planning} of 6-{DOF} {Free}-{Floating} {Space} {Robotic} {Manipulators} {Using} {Reinforcement} {Learning}},
	volume = {224},
	issn = {00945765},
	url = {https://linkinghub.elsevier.com/retrieve/pii/S009457652400451X},
	doi = {10.1016/j.actaastro.2024.08.015},
	language = {en},
	urldate = {2025-09-16},
	journal = {Acta Astronautica},
	author = {Al Ali, Ahmad and Shi, Jian-Feng and Zhu, Zheng H.},
	month = nov,
	year = {2024},
	pages = {367--378},
}

@article{zhao_transformable_2016,
	title = {Transformable {Multirotor} {With} {Two}-{Dimensional} {Multilinks}: {Modeling}, {Control}, and {Motion} {Planning} for {Aerial} {Transformation}},
	volume = {30},
	issn = {0169-1864, 1568-5535},
	shorttitle = {Transformable multirotor with two-dimensional multilinks},
	url = {http://www.tandfonline.com/doi/full/10.1080/01691864.2016.1181006},
	doi = {10.1080/01691864.2016.1181006},
	language = {en},
	number = {13},
	urldate = {2024-11-17},
	journal = {Advanced Robotics},
	author = {Zhao, Moju and Kawasaki, Koji and Okada, Kei and Inaba, Masayuki},
	month = jul,
	year = {2016},
	pages = {825--845},
}

@article{vaquero_eels_2024,
	title = {{EELS}: {Autonomous} {Snake}-{Like} {Robot} {With} {Task} and {Motion} {Planning} {Capabilities} for {Ice} {World} {Exploration}},
	volume = {9},
	issn = {2470-9476},
	shorttitle = {{EELS}},
	url = {https://www.science.org/doi/10.1126/scirobotics.adh8332},
	doi = {10.1126/scirobotics.adh8332},
	language = {en},
	number = {88},
	urldate = {2025-04-23},
	journal = {Science Robotics},
	author = {Vaquero, T. S. and Daddi, G. and Thakker, R. and Paton, M. and Jasour, A. and Strub, M. P. and Swan, R. M. and Royce, R. and Gildner, M. and Tosi, P. and Veismann, M. and Gavrilov, P. and Marteau, E. and Bowkett, J. and De Mola Lemus, D. Loret and Nakka, Y. and Hockman, B. and Orekhov, A. and Hasseler, T. D. and Leake, C. and Nuernberger, B. and Proença, P. and Reid, W. and Talbot, W. and Georgiev, N. and Pailevanian, T. and Archanian, A. and Ambrose, E. and Jasper, J. and Etheredge, R. and Roman, C. and Levine, D. and Otsu, K. and Yearicks, S. and Melikyan, H. and Rieber, R. R. and Carpenter, K. and Nash, J. and Jain, A. and Shiraishi, L. and Robinson, M. and Travers, M. and Choset, H. and Burdick, J. and Gardner, A. and Cable, M. and Ingham, M. and Ono, M.},
	month = mar,
	year = {2024},
	pages = {eadh8332},
}

@article{liu_review_2021,
	title = {Review of {Snake} {Robots} in {Constrained} {Environments}},
	volume = {141},
	issn = {09218890},
	url = {https://linkinghub.elsevier.com/retrieve/pii/S0921889021000701},
	doi = {10.1016/j.robot.2021.103785},
	language = {en},
	urldate = {2025-09-06},
	journal = {Robotics and Autonomous Systems},
	author = {Liu, Jindong and Tong, Yuchuang and Liu, Jinguo},
	month = jul,
	year = {2021},
	pages = {103785},
}

@article{zhao_versatile_2023,
	title = {Versatile {Articulated} {Aerial} {Robot} {DRAGON}: {Aerial} {Manipulation} and {Grasping} by {Vectorable} {Thrust} {Control}},
	volume = {42},
	issn = {0278-3649, 1741-3176},
	shorttitle = {Versatile articulated aerial robot {DRAGON}},
	url = {http://journals.sagepub.com/doi/10.1177/02783649221112446},
	doi = {10.1177/02783649221112446},
	language = {en},
	number = {4-5},
	urldate = {2023-12-28},
	journal = {The International Journal of Robotics Research},
	author = {Zhao, Moju and Okada, Kei and Inaba, Masayuki},
	month = apr,
	year = {2023},
	pages = {214--248},
}

@article{zhao_transformable_2018,
	title = {Transformable {Multirotor} {With} {Two}-{Dimensional} {Multilinks}: {Modeling}, {Control}, and {Whole}-{Body} {Aerial} {Manipulation}},
	volume = {37},
	issn = {0278-3649, 1741-3176},
	shorttitle = {Transformable multirotor with two-dimensional multilinks},
	url = {http://journals.sagepub.com/doi/10.1177/0278364918801639},
	doi = {10.1177/0278364918801639},
	language = {en},
	number = {9},
	urldate = {2023-12-31},
	journal = {The International Journal of Robotics Research},
	author = {Zhao, Moju and Kawasaki, Koji and Anzai, Tomoki and Chen, Xiangyu and Noda, Shintaro and Shi, Fan and Okada, Kei and Inaba, Masayuki},
	month = aug,
	year = {2018},
	pages = {1085--1112},
}

@article{liu_planning_2025,
	title = {A {Planning} {Framework} for {Complex} {Flipping} {Manipulation} of {Multiple} {Mobile} {Manipulators}},
	volume = {10},
	issn = {2377-3766},
	url = {https://ieeexplore.ieee.org/document/10948354/},
	doi = {10.1109/LRA.2025.3557749},
	number = {5},
	urldate = {2025-09-17},
	journal = {IEEE Robotics and Automation Letters},
	author = {Liu, Wenhang and Ren, Meng and Song, Kun and Wang, Michael Yu and Xiong, Zhenhua},
	month = may,
	year = {2025},
	pages = {5162--5169},
}

@article{kulkarni_reconfigurable_2020,
	title = {The {Reconfigurable} {Aerial} {Robotic} {Chain}: {Shape} and {Motion} {Planning}},
	volume = {53},
	issn = {24058963},
	shorttitle = {The {Reconfigurable} {Aerial} {Robotic} {Chain}},
	url = {https://linkinghub.elsevier.com/retrieve/pii/S2405896320330548},
	doi = {10.1016/j.ifacol.2020.12.2383},
	language = {en},
	number = {2},
	urldate = {2025-09-16},
	journal = {IFAC-PapersOnLine},
	author = {Kulkarni, Mihir and Nguyen, Huan and Alexis, Kostas},
	year = {2020},
	pages = {9295--9302},
}

@article{meng_online_2024,
	title = {Online {Adaptive} {Motion} {Generation} for {Humanoid} {Locomotion} on {Non}-{Flat} {Terrain} via {Template} {Behavior} {Extension}},
	volume = {21},
	issn = {1558-3783},
	url = {https://ieeexplore.ieee.org/document/10305536/},
	doi = {10.1109/TASE.2023.3327819},
	number = {4},
	urldate = {2025-09-16},
	journal = {IEEE Transactions on Automation Science and Engineering},
	author = {Meng, Xiang and Yu, Zhangguo and Chen, Xuechao and Huang, Zelin and Meng, Fei and Huang, Qiang},
	month = oct,
	year = {2024},
	pages = {6563--6574},
}

@article{gammell_asymptotically_2021,
	title = {Asymptotically {Optimal} {Sampling}-{Based} {Motion} {Planning} {Methods}},
	volume = {4},
	issn = {2573-5144, 2573-5144},
	url = {https://www.annualreviews.org/doi/10.1146/annurev-control-061920-093753},
	doi = {10.1146/annurev-control-061920-093753},
	language = {en},
	number = {1},
	urldate = {2025-09-06},
	journal = {Annual Review of Control, Robotics, and Autonomous Systems},
	author = {Gammell, Jonathan D. and Strub, Marlin P.},
	month = may,
	year = {2021},
	pages = {295--318},
}

@article{richter_arcsnake_2022,
	title = {{ARCSnake}: {Reconfigurable} {Snakelike} {Robot} {With} {Archimedean} {Screw} {Propulsion} for {Multidomain} {Mobility}},
	volume = {38},
	issn = {1941-0468},
	shorttitle = {{ARCSnake}},
	url = {https://ieeexplore.ieee.org/document/9530203/},
	doi = {10.1109/TRO.2021.3104968},
	number = {2},
	urldate = {2025-09-06},
	journal = {IEEE Transactions on Robotics},
	author = {Richter, Florian and Gavrilov, Peter V. and Lam, Hoi Man and Degani, Amir and Yip, Michael C.},
	month = apr,
	year = {2022},
	pages = {797--809},
}

@article{zhao_online_2020,
	title = {Online {Motion} {Planning} for {Deforming} {Maneuvering} and {Manipulation} by {Multilinked} {Aerial} {Robot} {Based} on {Differential} {Kinematics}},
	volume = {5},
	issn = {2377-3766, 2377-3774},
	url = {https://ieeexplore.ieee.org/document/8962166/},
	doi = {10.1109/LRA.2020.2967285},
	language = {en},
	number = {2},
	urldate = {2023-12-28},
	journal = {IEEE Robotics and Automation Letters},
	author = {Zhao, Moju and Shi, Fan and Anzai, Tomoki and Okada, Kei and Inaba, Masayuki},
	month = apr,
	year = {2020},
	pages = {1602--1609},
}

@article{zhao_singularity-free_2021,
	title = {Singularity-{Free} {Aerial} {Deformation} by {Two}-{Dimensional} {Multilinked} {Aerial} {Robot} {With} 1-{DoF} {Vectorable} {Propeller}},
	volume = {6},
	issn = {2377-3766},
	url = {https://ieeexplore.ieee.org/document/9343757/},
	doi = {10.1109/LRA.2021.3056027},
	number = {2},
	urldate = {2025-07-15},
	journal = {IEEE Robotics and Automation Letters},
	author = {Zhao, Moju and Anzai, Tomoki and Okada, Kei and Kawasaki, Koji and Inaba, Masayuki},
	month = apr,
	year = {2021},
	pages = {1367--1374},
}

@inproceedings{thakker_eels_2023,
	address = {Detroit, MI, USA},
	title = {{EELS}: {Towards} {Autonomous} {Mobility} in {Extreme} {Terrain} with a {Versatile} {Snake} {Robot} with {Resilience} to {Exteroception} {Failures}},
	copyright = {https://doi.org/10.15223/policy-029},
	isbn = {978-1-6654-9190-7},
	shorttitle = {{EELS}},
	url = {https://ieeexplore.ieee.org/document/10341448/},
	doi = {10.1109/IROS55552.2023.10341448},
	language = {en},
	urldate = {2025-04-23},
	booktitle = {2023 {IEEE}/{RSJ} {International} {Conference} on {Intelligent} {Robots} and {Systems} ({IROS})},
	publisher = {IEEE},
	author = {Thakker, Rohan and Paton, Michael and Strub, Marlin P. and Swan, Michael and Daddi, Guglielmo and Royce, Rob and Tosi, Phillipe and Gildner, Matthew and Vaquero, Tiago and Veismann, Marcel and Gavrilov, Peter and Marteau, Eloise and Bowkett, Joseph and Loret, Daniel and Nakka, Yashwanth and Hockman, Benjamin and Orekhov, Andrew and Hasseler, Tristan and Leake, Carl and Nuernberger, Benjamin and Proença, Pedro and Reid, William and Talbot, William and Georgiev, Nikola and Pailevanian, Torkom and Archanian, Avak and Ambrose, Eric and Jasper, Jay and Etheredge, Rachel and Roman, Christiahn and Levine, Dan and Otsu, Kyohei and Melikyan, Hovhannes and Nash, Jeremy and Rieber, Richard and Carpenter, Kalind and Jain, Abhinandan and Shiraishi, Lori and Pastor, Daniel and Yearicks, Sarah and Ingham, Michel and Robinson, Matthew and Agha, Ali and Travers, Matthew and Choset, Howie and Burdick, Joel and Ono, Masahiro},
	month = oct,
	year = {2023},
	pages = {9886--9893},
}

@article{wang_implicit_2024,
	title = {Implicit {Swept} {Volume} {SDF}: {Enabling} {Continuous} {Collision}-{Free} {Trajectory} {Generation} for {Arbitrary} {Shapes}},
	volume = {43},
	issn = {0730-0301, 1557-7368},
	shorttitle = {Implicit {Swept} {Volume} {SDF}},
	url = {https://dl.acm.org/doi/10.1145/3658181},
	doi = {10.1145/3658181},
	language = {en},
	number = {4},
	urldate = {2025-04-16},
	journal = {ACM Transactions on Graphics},
	author = {Wang, Jingping and Zhang, Tingrui and Zhang, Qixuan and Zeng, Chuxiao and Yu, Jingyi and Xu, Chao and Xu, Lan and Gao, Fei},
	month = jul,
	year = {2024},
	pages = {1--14},
}

@article{ollero_past_2022,
	title = {Past, {Present}, and {Future} of {Aerial} {Robotic} {Manipulators}},
	volume = {38},
	issn = {1941-0468},
	url = {https://ieeexplore.ieee.org/document/9462539/?arnumber=9462539},
	doi = {10.1109/TRO.2021.3084395},
	number = {1},
	urldate = {2024-11-08},
	journal = {IEEE Transactions on Robotics},
	author = {Ollero, Anibal and Tognon, Marco and Suarez, Alejandro and Lee, Dongjun and Franchi, Antonio},
	month = feb,
	year = {2022},
	note = {Conference Name: IEEE Transactions on Robotics},
	pages = {626--645},
}

@article{yang_rampage_2024,
	title = {{RAMPAGE}: {Toward} {Whole}-{Body}, {Real}-{Time}, and {Agile} {Motion} {Planning} in {Unknown} {Cluttered} {Environments} for {Mobile} {Manipulators}},
	volume = {71},
	issn = {1557-9948},
	shorttitle = {{RAMPAGE}},
	url = {https://ieeexplore.ieee.org/document/10472786/?arnumber=10472786},
	doi = {10.1109/TIE.2024.3370969},
	number = {11},
	urldate = {2024-11-08},
	journal = {IEEE Transactions on Industrial Electronics},
	author = {Yang, Yuqiang and Meng, Fei and Meng, Zehui and Yang, Chenguang},
	month = nov,
	year = {2024},
	note = {Conference Name: IEEE Transactions on Industrial Electronics},
	pages = {14492--14502},
}

@article{salzman_motion_2016,
	title = {Motion {Planning} for {Multilink} {Robots} by {Implicit} {Configuration}-{Space} {Tiling}},
	volume = {1},
	issn = {2377-3766},
	url = {https://ieeexplore.ieee.org/document/7397927/?arnumber=7397927},
	doi = {10.1109/LRA.2016.2524066},
	language = {en-US},
	number = {2},
	urldate = {2024-11-08},
	journal = {IEEE Robotics and Automation Letters},
	author = {Salzman, Oren and Solovey, Kiril and Halperin, Dan},
	month = jul,
	year = {2016},
	note = {Conference Name: IEEE Robotics and Automation Letters},
	pages = {760--767},
}

@inproceedings{mellinger_minimum_2011,
	address = {Shanghai, China},
	title = {Minimum {Snap} {Trajectory} {Generation} and {Control} for {Quadrotors}},
	isbn = {978-1-61284-386-5},
	url = {http://ieeexplore.ieee.org/document/5980409/},
	doi = {10.1109/ICRA.2011.5980409},
	language = {en},
	urldate = {2023-05-10},
	booktitle = {2011 {IEEE} {International} {Conference} on {Robotics} and {Automation} ({ICRA})},
	publisher = {IEEE},
	author = {Mellinger, Daniel and Kumar, Vijay},
	month = may,
	year = {2011},
	pages = {2520--2525},
}

@article{wu_deep_2024,
	title = {Deep {Learning} for {Optimization} of {Trajectories} for {Quadrotors}},
	volume = {9},
	issn = {2377-3766, 2377-3774},
	url = {https://ieeexplore.ieee.org/document/10412114/},
	doi = {10.1109/LRA.2024.3357399},
	language = {en},
	number = {3},
	urldate = {2024-03-12},
	journal = {IEEE Robotics and Automation Letters},
	author = {Wu, Yuwei and Sun, Xiatao and Spasojevic, Igor and Kumar, Vijay},
	month = mar,
	year = {2024},
	pages = {2479--2486},
}

@article{ding_efficient_2019,
	title = {An {Efficient} {B}-{Spline}-{Based} {Kinodynamic} {Replanning} {Framework} for {Quadrotors}},
	volume = {35},
	copyright = {https://ieeexplore.ieee.org/Xplorehelp/downloads/license-information/IEEE.html},
	issn = {1552-3098, 1941-0468},
	url = {https://ieeexplore.ieee.org/document/8811597/},
	doi = {10.1109/TRO.2019.2926390},
	language = {en},
	number = {6},
	urldate = {2024-04-10},
	journal = {IEEE Transactions on Robotics},
	author = {Ding, Wenchao and Gao, Wenliang and Wang, Kaixuan and Shen, Shaojie},
	month = dec,
	year = {2019},
	pages = {1287--1306},
}

@article{liu_integrated_2023,
	title = {Integrated {Planning} and {Control} for {Quadrotor} {Navigation} in {Presence} of {Suddenly} {Appearing} {Objects} and {Disturbances}},
	issn = {2377-3766, 2377-3774},
	url = {https://ieeexplore.ieee.org/document/10238764/},
	doi = {10.1109/LRA.2023.3311358},
	language = {en},
	urldate = {2023-10-19},
	journal = {IEEE Robotics and Automation Letters},
	author = {Liu, Wenyi and Ren, Yunfan and Zhang, Fu},
	year = {2023},
	pages = {1--8},
}

@inproceedings{ren_online_2023,
	address = {London, United Kingdom},
	title = {Online {Whole}-{Body} {Motion} {Planning} for {Quadrotor} using {Multi}-{Resolution} {Search}},
	isbn = {9798350323658},
	url = {https://ieeexplore.ieee.org/document/10160767/},
	doi = {10.1109/ICRA48891.2023.10160767},
	language = {en},
	urldate = {2023-10-19},
	booktitle = {2023 {IEEE} {International} {Conference} on {Robotics} and {Automation} ({ICRA})},
	publisher = {IEEE},
	author = {Ren, Yunfan and Liang, Siqi and Zhu, Fangcheng and Lu, Guozheng and Zhang, Fu},
	month = may,
	year = {2023},
	pages = {1594--1600},
}

@article{hart_formal_1968,
	title = {A {Formal} {Basis} for the {Heuristic} {Determination} of {Minimum} {Cost} {Paths}},
	volume = {4},
	issn = {0536-1567},
	number = {2},
	journal = {IEEE Transactions on Systems Science and Cybernetics},
	author = {Hart, Peter E and Nilsson, Nils J and Raphael, Bertram},
	month = jul,
	year = {1968},
	note = {Publisher: IEEE},
	pages = {100--107},
}

@article{zhou_raptor_2021,
	title = {{RAPTOR}: {Robust} and {Perception}-{Aware} {Trajectory} {Replanning} for {Quadrotor} {Fast} {Flight}},
	volume = {37},
	issn = {1552-3098, 1941-0468},
	shorttitle = {{RAPTOR}},
	url = {https://ieeexplore.ieee.org/document/9422918/},
	doi = {10.1109/TRO.2021.3071527},
	language = {en},
	number = {6},
	urldate = {2023-05-10},
	journal = {IEEE Transactions on Robotics},
	author = {Zhou, Boyu and Pan, Jie and Gao, Fei and Shen, Shaojie},
	month = dec,
	year = {2021},
	pages = {1992--2009},
}


 


\begin{IEEEbiography}[{\includegraphics[width=1in,height=1.25in,clip,keepaspectratio]{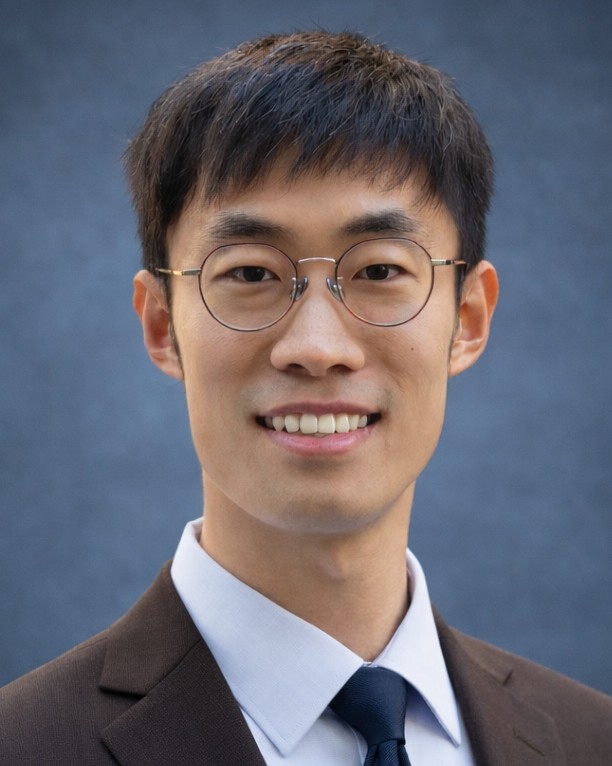}}]{Yicheng Chen} received the B.E. degree from Beihang University, Beijing, China, in 2021, and the M.Sc. degree from Beihang University in 2024. He is currently pursuing the Ph.D. degree with the Department of Mechanical Engineering, The University of Tokyo, Tokyo, Japan. His research interests include motion planning in aerial robotics and the intersection of optimization and learning-based methodologies.
\end{IEEEbiography}

\vspace{-20pt}

\begin{IEEEbiography}[{\includegraphics[width=1in,height=1.25in,clip,keepaspectratio]{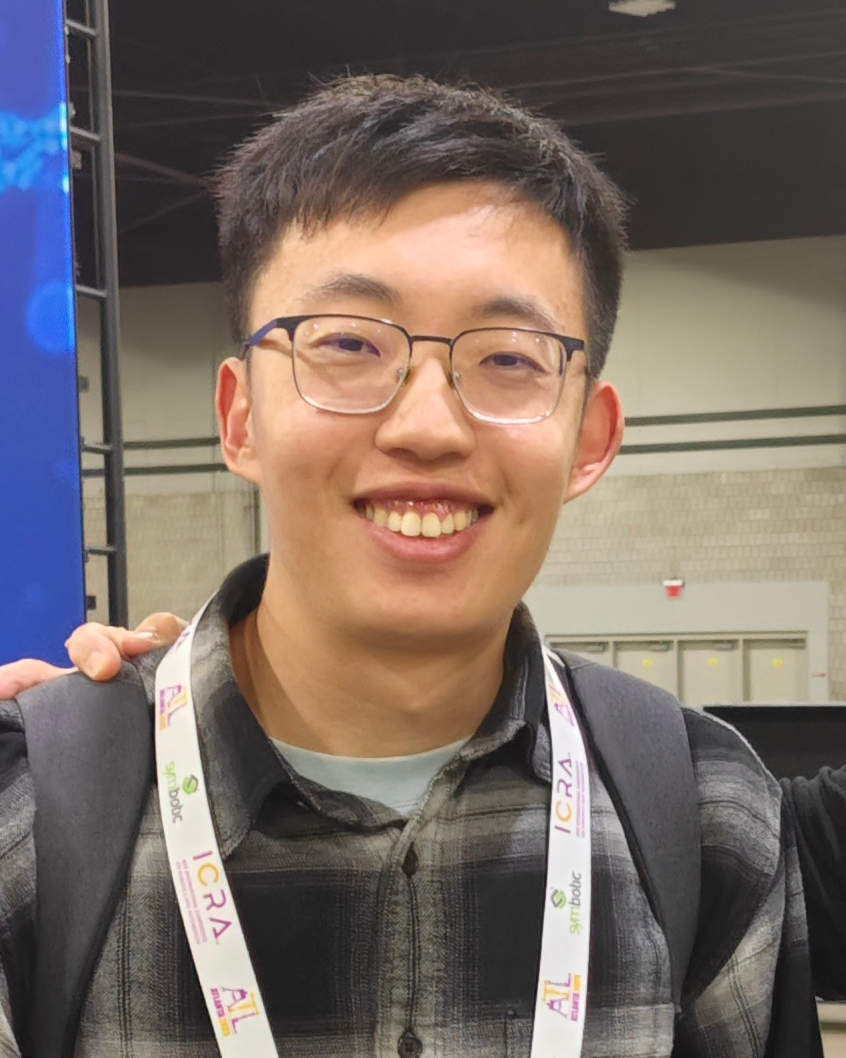}}]{Jinjie Li} received the B.Eng. degree in Automation and the M.Sc. degree in Control Science and Engineering from Beihang University, Beijing, China, in 2020 and 2023, respectively. He is currently pursuing the Ph.D. degree in the Department of Mechanical Engineering, The University of Tokyo, Tokyo, Japan. His research interests include optimization-based control for aerial manipulation, aiming to make aerial robots function as flying hands rather than just eyes.
\end{IEEEbiography}

\vspace{-20pt}

\begin{IEEEbiography}[{\includegraphics[width=1in,height=1.25in,clip,keepaspectratio]{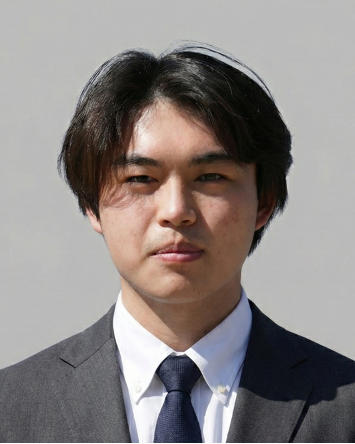}}]{Haokun Liu} is currently a Ph.D. student with the Department of Mechanical Engineering, The University of Tokyo, Tokyo, Japan. His research interests include vision–language and large language model-based robot control, heterogeneous multi-robot collaboration, and learning-enabled motion planning and state estimation.
\end{IEEEbiography}

\vspace{-20pt}

\begin{IEEEbiography}[{\includegraphics[width=1in,height=1.25in,clip,keepaspectratio]{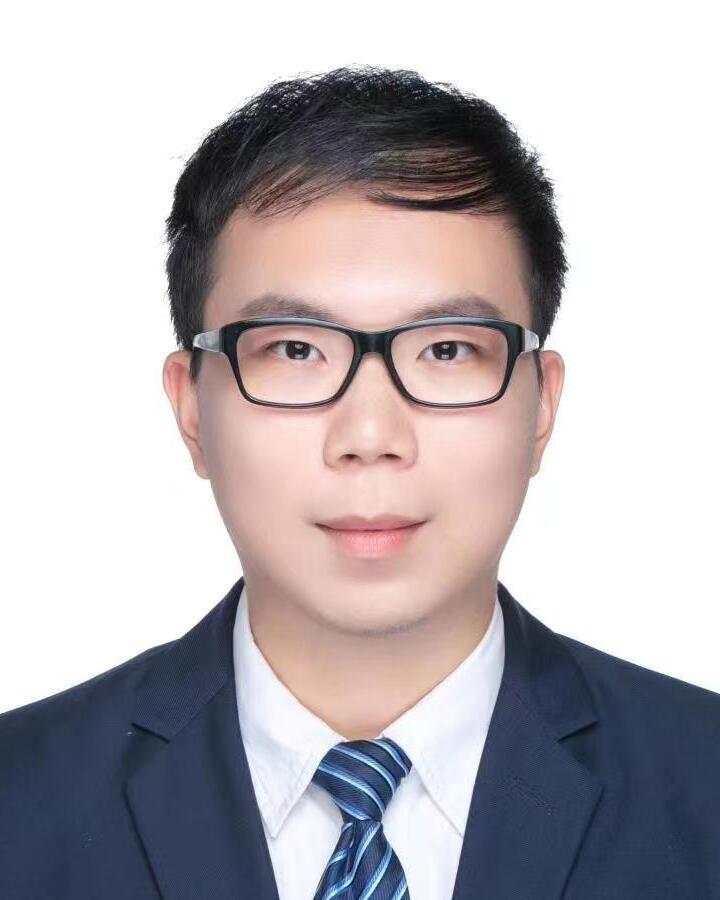}}]{Zicheng Luo} received the B.E. degree from the School of Mechanical Engineering, Tongji University, Shanghai, China, in 2024. He is currently working toward the M.S. degree with the Department of Mechanical Engineering, The University of Tokyo, Tokyo, Japan. His research interests include modeling and control, motion planning, and manipulation of aerial robots.
\end{IEEEbiography}

\vspace{-20pt}

\begin{IEEEbiography}[{\includegraphics[width=1in,height=1.25in,clip,keepaspectratio]{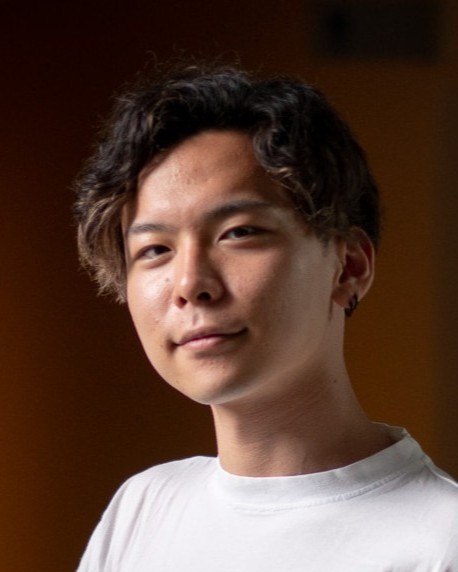}}]{Kotaro kaneko} received the B.E. degree in 2024 from the Department of Mechanical Engineering, University of Tokyo, Tokyo, Japan, where he is currently working toward the Master's degree. His research interests include teleoperation systems, particularly the design, modeling, and control of haptic devices, as well as teleoperation system architecture.
\end{IEEEbiography}

\vspace{-20pt}

\begin{IEEEbiography}[{\includegraphics[width=1in,height=1.25in,clip,keepaspectratio]{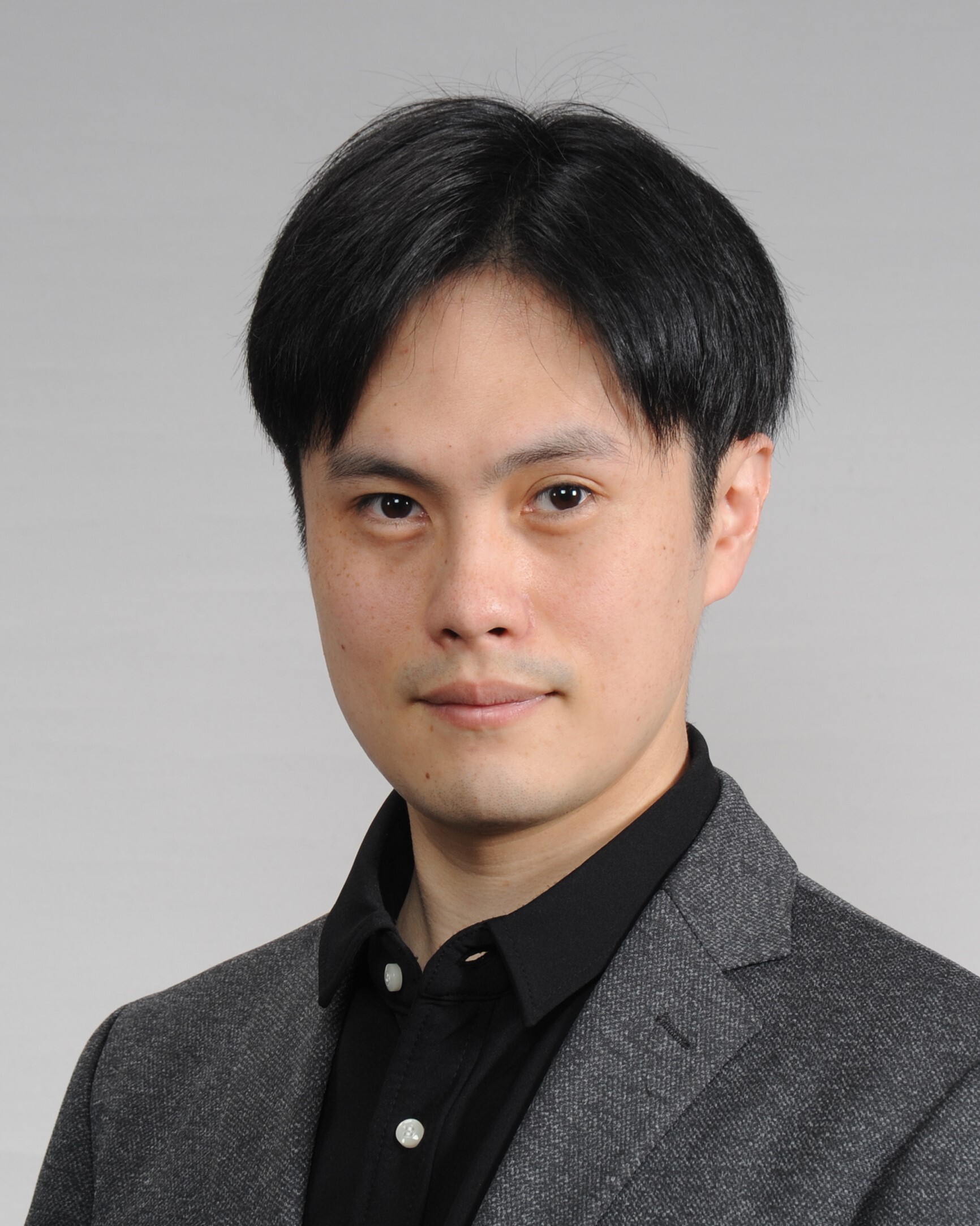}}]{Moju Zhao} is currently a Lecturer (Junior Associate Professor) at The University of Tokyo. He received his doctoral degree from the Department of Mechano-Informatics, The University of Tokyo, in 2018. His research interests include mechanical design, modeling and control, motion planning, and vision-based recognition in aerial robotics. His main achievements include articulated aerial robots such as DRAGON and SPIDAR, and he has received several awards at conferences and in journals, including the Best Paper Award at IEEE ICRA 2018. He has also been an Associate Editor of IEEE T-RO since 2026.
\end{IEEEbiography}


\end{document}